\documentclass[10pt,twocolumn,letterpaper]{article}

\usepackage{cvpr}
\usepackage{times}
\usepackage{epsfig}
\usepackage{graphicx}
\usepackage{amsmath}
\usepackage{amssymb}
\usepackage{comment}
\usepackage{subcaption}

\newcommand{\arxivcond}[2]{#1}  %arXiv

\newcommand{\supplement}{\arxivcond{Appendix}{Supplemental Material}}

\arxivcond{\cvprfinalcopy}{}

\newcommand{\bx}{\mathbf{x}}
\newcommand{\bw}{\mathbf{w}}

\newcommand{\tn}[1]{\textsf{#1}}
\newcommand{\synset}[1]{\textit{#1}}

\newcommand{\sigmoid}{\sigma}
\usepackage[pagebackref=true,breaklinks=true,letterpaper=true,colorlinks,bookmarks=false]{hyperref}

\graphicspath{{figures/}}

\def\cvprPaperID{337} % *** Enter the CVPR Paper ID here

\ifcvprfinal\fi
\begin{document}

%%%%%%%%% TITLE
\title{Image Classification and Retrieval from User-Supplied Tags}

\author{
Hamid Izadinia \\
Univ.~of Washington
\and
Ali Farhadi \\
Univ.~of Washington 
\and
Aaron Hertzmann \\
Adobe Research
\and
Matthew D.~Hoffman \\
Adobe Research
}

\maketitle
%\thispagestyle{empty}

%%%%%%%%%%%%%%%%%%%%%%%%%%%%%%%%%%%% ABSTRACT %%%%%%%%%%%%%%%%%%%%%%%%%%%%%%%%%%%%%
\begin{abstract}
This paper proposes direct learning of image classification from user-supplied tags, without filtering.
Each tag is supplied by the user who shared the image online. Enormous numbers of these tags are freely available online, and they give insight about the image categories important to users and to image classification. Our approach is complementary to the conventional approach of manual annotation, which is extremely costly.  We analyze of the Flickr 100 Million Image dataset, making several useful observations about the statistics of these tags. We introduce a large-scale robust classification algorithm, in order to handle the inherent noise in these tags, and a calibration procedure to better predict objective annotations. We show that freely available, user-supplied tags can obtain similar or superior results to large databases of costly manual annotations.
\arxivcond{}{Interactive annotation and retrieval demos can be found online at \url{http://54.244.94.90}.
\\
\textbf{Note to reviewers: extensive qualitative results can be found in the Supplemental Material.}
}
\end{abstract}

%%%%%%%%%%%%%%%%%%%%%%%%%%%%%%%%%%%% BODY TEXT %%%%%%%%%%%%%%%%%%%%%%%%%%%%%%%%%%%%
\section{Introduction}

%Research in object recognition has made dramatic strides in the past few years, with substantial improvements in scores in the ImageNet and PASCAL competitions.This work relies on large scale, hand-labeled datasets. Gathering these datasets entails considerable time and expense, because of the need to provide labels of all images, and to ensure accuracy of these labels. Despite this progress, these datasets comprise only a fraction of recognizable visual objects and concepts.  The labels missing from these datasets include many important image concepts that are important to human users. 

Object recognition has made dramatic strides in the past few years. This progress is partly due to the creation of large-scale hand-labeled datasets. Collecting these datasets involves listing object categories, searching the web for images of each category, pruning irrelevant images and providing detailed labels for each image. There are several major issues with this approach. First, gathering high-quality annotations for large datasets requires substantial effort and expense.
Second, it remains unclear how best to determine the list of categories.
Existing datasets comprise only a fraction of recognizable visual concepts, and often miss concepts that are important to end-users.
These datasets draw rigid distinctions between different types of concepts (e.g., scenes, attributes, objects) that exclude many important concepts.
%The space of important visual concepts remains unknown.

%However, little attention has been made into the list of categories. 
%Adding new concepts to these datasets involves repeating all the expensive steps above.

This paper introduces an approach to learning about visual concepts by employing user-supplied tags. That is, we directly use the tags provided by the users that uploaded the images to photo-sharing services, without any subsequent manual filtering or curation. Tags in the photosharing services reflect the image categories that are important to users and include scenes (\tn{beach}), objects (\tn{car}), attributes (\tn{rustic}), activities (\tn{wedding}),
and visual styles (\tn{portrait}), as well as concepts that are harder to categorize (\tn{family}). Online sharing is growing and many services host content other than photographs (e.g., Behance, Imgur, Shapeways). The tags in these services are abundant, and learning about them could benefit a broad range of consumer applications such as tag suggestion and search-by-tag.

%\paragraph{Challenges.} 
User-supplied tags are freeform and using them presents significant challenges. 
These tags are entirely uncurated, so users provide tags for their images in different ways. Different users provide different numbers of tags per image, and, conversely, choose different subsets of tags. One tag may have multiple meanings, and, conversely, multiple terms may be used for the same concept. Most sharing sites provide no quality control whatsoever for their tags. Hence, it is important to design learning algorithms robust to these factors. %In this paper, we introduce a loss function that is robust to the inherent noise in the user provided tags.

\paragraph{Contributions.}
In addition to introducing the direct use of user-supplied tags, this paper presents several contributions.  First, we analyze statistics of tags in a large Flickr dataset, making useful observations about how tags are used and when they are reliable. Second, we introduce a robust logistic
regression method for classification with user-supplied tags, which is robust to randomly omitted positive labels. Since tag noise is different for different tags, the tag outlier probabilities are learned simultaneously with the classifier weights.  Third, we describe calibration of the trained model probabilities from a small annotation set.

We demonstrate results for several tags: predicting the tags that a user would give to an image,
predicting objective annotations for an image,
and retrieving images for a tag query. For the latter two tasks, which require objective anotations, we calibrate and test on the manually-annotated NUS-WIDE \cite{NUS} dataset.
%Note that predicting user tags is different from predicting objective annotations because there are many true annotations that users rarely provide, e.g., ``clouds" or ``buildings."

We show that training on a large collection of freely available, user-supplied tags alone obtains comparable performance to using a smaller, manually-annotated training set.
That is, we can learn to predict thousands of tags \textit{without any curated annotations at all.}
Moreover, if we calibrate the model with a small annotated dataset, we can obtain superior performance to conventional annotations at a tiny fraction (1/200) of the labeling cost.
%We also show that using robust classification yields better class probabilities, as evidence by substantial improvement in multi-tag-based image retrieval.
Our methods could support several annotation applications, such as auto-suggesting tags to users, clustering user photos by activity or event, and photo database search.
We also demonstrate that using robust classification substantially improves image retrieval performance with multi-tag queries.

\section{Related Work}

The amazing progress of the recent years of vision has been driven in part by datasets.
These datasets are built through a combination of webscraping and crowd-sourcing, with the aim of labeling the data as cleanly as possible.  ImageNet \cite{ImageNet} is the most prominent whole-image classification dataset, but other recent examples include NUS-WIDE \cite{NUS}, the SUN scene attribute database \cite{SUNattributes,SUN}, and PLACES \cite{zhou2014places}. %, and  Microsoft COCO \cite{coco}.
The curation process has a number of drawbacks, 
such as the cost of gathering clean labels and the difficulty in determining a useful space of labels. It is unclear that this procedure alone will scale to the space of all important concepts for vision \cite{ImageNet}. We take a complementary approach of using a massive database of freely available images with noisy, unfiltered tags.

Merging noisy labels is a classic problem in item-response theory, and has been applied in the crowdsourcing literature \cite{raykar,welinder}. We extend robust logistic regression \cite{raykar} to large-scale learning with Stochastic EM.  In image recognition, a related problem occurs when harvesting noisy data from the web \cite{Chatfield12,neil,visualCategoryFilter,optimol,Schroff11}; these methods take complementary approaches to ours, and focus on object and scene categories.

To our knowledge, no previous work directly learns image classifiers from raw Flickr tags without curation.
Most similar to our own work, 
Zhu et al.~\cite{zhu_refinement} use matrix factorization to clean up a collection of tags.  In principle, this method could be used as a first step toward learning classifiers, though it has not been tested as such. This method requires batch computation and is unlikely to be practical for large numbers of tags and images. 
Gong et al.~\cite{gongEmbedding} use raw Flickr tags as side-information for associating images with descriptive text.

%Many of the tags that we learn are related to specific individal datasets, such as scene tags \cite{}, object identity \cite{ImageNet}, and image style \cite{AVA?,Karayev}. 
Most previous work has focused on names and attributes for objects and scenes, including previous work on image tagging (e.g., \cite{attr09,google_rank_loss,grauman_tags,SUNattributes,ImageNet, SUN,zhou2014places}). Unfortunately these datasets  are disjoint and little attention has been paid to the list of objects, scenes, and attributes. Our solution is to learn what users care about, using a robust loss function that takes into account the noise in the labels. 
We learn many other kinds of tags, such as tags for events, activities, and image style. There have been a few efforts aimed at modeling a few kinds of image style and aesthetics \cite{karayev,Murray-CVPR-2012}.

\section{Analysis of User-Supplied Tags}
\label{sec:analysis}

When can user-supplied tags be useful, and when can they be trusted?  In this section, we analyze the tags provided on Flickr, and compare them to two datasets with ground truth labels.
Some of these observations motivate our algorithm in Section \ref{sec:algorithm}, and others provide fodder for future research.

%In this paper, we focus on the large Flickr dataset, but also evaluate using other datsets gathered from Flickr.
\paragraph{Flickr 100 Million (F100M).}  Our main dataset is the Yahoo/Flickr Creative Commons 100M dataset\footnote{http://yahoolabs.tumblr.com/post/89783581601}.  This dataset comprises 99.3 million images, each of which includes a list of the tags supplied by the user that uploaded the image.  

\subsection{Types of tags}

\begin{figure}
\centering
\includegraphics[width=3in]{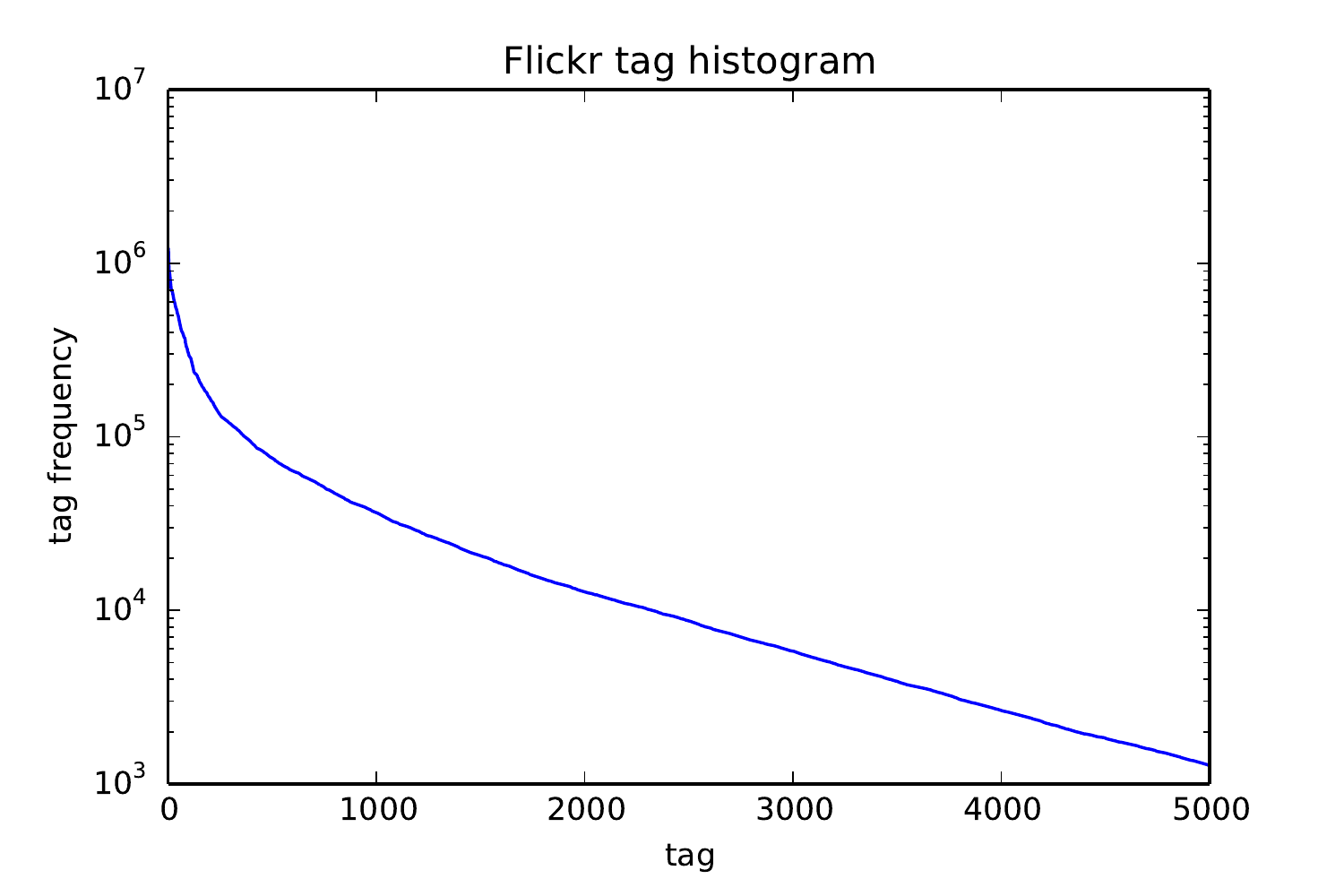}
\caption{
   \label{fig:taghist}
Tag histogram for the most popular tags, excluding non-image tags.
The distribution is heavy-tailed, and there are 5400 tags with more than 1000 images each. 
}
\vspace{-0.1in}
\end{figure}

The F100M dataset provides an enormous number of images and tags (Figure \ref{fig:taghist}) that could be used for learning.  Some of the most frequent tags are shown in Table \ref{fig:frequenttags}. There are 5400  tags that occur in at least 1000 images. 
The set of tags provides a window into the image concepts that are important to users.
Many of these represent types of image label that are not represented in previous datasets.

\begin{table}
\centering
\begin{tabular}{l@{~~~}c@{~~~}l@{~~~}c@{~~~}c}
\hline
Flickr tag & \# Flickr & synset & \# node & \# subtree \\
\hline
\tn{travel} & 1221148 & \synset{travel.n.01} & 0 & 0 \\
\tn{wedding} & 734438 & \synset{wedding.n.03} & 1257 & 1257 \\ 
\tn{flower} & 907773 & \synset{flower.n.01} & 1924 & 339376 \\
\tn{art} & 902043 & \synset{art.n.01} & 0 & 11353 \\
\tn{music} & 826692&\synset{music.n.01} & 0 & 0 \\
\tn{party} & 669065&\synset{party.n.01} & 0${}^*$ & 0 \\
\tn{nature} & 872029&\synset{nature.n.01} & 0 & 0 \\
\tn{beach} & 768752&\synset{beach.n.01} & 1713 & 1773 \\
\tn{city} & 701823 & \synset{city.n.01} & 1224 & 1224 \\
\tn{tree} & 697009 & \synset{tree.n.01} & 1181 & 563038 \\
\tn{vacation} & 694523&\synset{vacation.n.01} & 0 & 0 \\
\tn{park} & 686458 & \synset{park.n.01} & 0 & 0 \\
\tn{people} & 641571&\synset{people.n.01} & 1431 & 1431 \\
\tn{water} & 640259 & \synset{water.n.06} & 759 & 7585 \\
{\tiny\tn{architecture}} & 616299 & {\tiny\synset{architecture.n.01}} & 1298 & 1298 \\
\tn{car} & 610114 & \synset{car.n.01} & 1307 & 40970 \\
\tn{festival} & 609638 & \synset{festival.n.01} & 0 & 0 \\
\tn{concert} & 605163 & \synset{concert.n.01} & 1322 & 1322 \\
\tn{summer} & 601816&\synset{summer.n.01} & 0 & 0 \\
\tn{sport} & 564703 & \synset{sport.n.01} & 1888 & 200402 \\
%\tn{food} & 500102 & \synset{food.n.01} & 1272 & 1000971 \\
%\tn{family} &5040 &\synset{kin.n.01} & 1326 & 1326 \\
\hline
\end{tabular}
\caption{The 20 most frequent tags in F100M, after merging plurals and omitting non-image/location tags. Corresponding ImageNet synsets are given, along with synset node and subtree counts.  These statistics are typical: we estimate that nearly half of popular Flickr tags are absent from ImageNet. Moreover, even when there is correspondence, some ImageNet tags do not capture all meanings of a term (Section \ref{sec:imageNet}). 
Some of these tags are covered by scene attribute databases \cite{SUN,SUNattributes,zhou2014places}.  (${}^*$There are 66 \tn{party} images in ImageNet, in the wrong synset \synset{party.n.04}.)
\label{fig:frequenttags}
\vspace{-0.1in}
}
\end{table}

%\begin{itemize}
Some of the most important tag types are as follows:
\textbf{events and activities}
 such as \tn{travel}, \tn{music}, \tn{party}, \tn{festival}, \tn{football}, \tn{school};
\textbf{specific locations}
 such as \tn{california} and \tn{italy};
\textbf{scene types}
 such as \tn{nature}, \tn{part}, \tn{urban}, \tn{sunset}, etc.;
\textbf{the seasons}
(\tn{fall}, \tn{winter}, \tn{summer}, \tn{spring});
\textbf{image style}
such as \tn{portrait}, \tn{macro}, \tn{vintage}, \tn{hdr};
and \textbf{art and culture}
such as \tn{painting}, \tn{drawing}, \tn{graffiti}, \tn{fashion}, \tn{punk}.
Many frequent tags also represent categories that do not seem learnable from image data alone, 
which we call \textbf{non-image tags}, 
including years (\tn{2011}, \tn{2012}, ...), and specific camera and imaging platforms (\tn{nikon}, \tn{iphone}, \tn{slr}).

\subsection{Correspondence with ImageNet}
\label{sec:imageNet}

A main motivation for using F100M is that it contains information missing from existing, curated datasets. Does it?
We compare F100M to the ImageNet image classification dataset \cite{ImageNet}, %one of the pre-eminent classification datasets in current computer vision.  
which comprises 14 million images gathered from Flickr, 
labeled according to the WordNet hierarchy \cite{wordnet} through a carefully-designed crowdsourcing procedure.

In order to quantify the dataset gap, we studied the 100 most frequent tags in F100M (after omitting the non-image and location tags described above). For each tag, we manually determined a correspondence to WordNet, as follows. In WordNet, each concept is represented by a synonym set, or \textit{synset}. WordNet synsets are ordered, and most tags (78\%) correspond to the first WordNet noun synset for that word. For example, the tag \tn{beach} corresponds to the synset \synset{beach.n.01}.
In other cases, we corrected the match manually.
The most-frequent examples are shown in Table \ref{fig:frequenttags}, and more are shown in the \supplement.
\textit{Based on this analysis and some simple calculations, we estimate that about 
half of the common Flickr non-image tags are absent from ImageNet.}
Details of how this estimate was formed are given in the \supplement.%
Some of these missing tags are covered by scene \cite{SUNattributes,SUN,zhou2014places} and style databases \cite{karayev,Murray-CVPR-2012}. 

Even when there is a corresponding tag in ImageNet, the tag may be poorly represented.  There are 11k images in the ImageNet \synset{art.n.01} hierarchy, but there are only 8 subtrees of \synset{art.n.01} with at least 1000 images; the  biggest ones are ``olympian zeus,'' ``cinquefoil,'' and ``finger-painting;''  and  there are no subtrees for ``painting,'' ``drawing,'' or ``illustration.'' The ImageNet synset for ``band'' includes only images for ``marching bands'' and not, say, ``rock bands.''

Many image categories that are significant to users---for example, in analyzing personal photo collections---are not well represented in the ImageNet categories.
Examples include \tn{family}, \tn{travel}, \tn{festival}, and \tn{summer}.

Some common tags in Flickr do not even exist in the WordNet hierarchy, such as \tn{cosplay} (a popular form of costume play), \tn{macro} (as in macro photography), and \tn{vintage} (in the sense of ``retro'' or ``old-style").  We also observed problems in the full ImageNet database, where large sets of images are assigned to the wrong synset, such as 
``party," ``landscape," and ``tree/tree diagram."

%Although the ImageNet challenge images are very carefully curated, the full database is not. There are a few cases where images are assigned to the wrong WordNet sense, including ``party'' ``landscape'' and ``tree.'' For example, the images in the synset node \synset{party.n.01} (A person involved in legal proceedings) are images of partygoers, whereas the subtrees are legal parties.
%  There are two senses of ``friend'' that seem to be used in the same way. It could also be argued that the selection of images assigned to \synset{call\_girl.n.01} is sexist. %``street'' has two very similar senses.
%\aaron{todo: fill in details if we want to keep this, maybe move to appendix.}

This is not in any way meant to disparage the substantial, important efforts of the ImageNet team, but to emphasize the enormous difficulty in trying to precisely curate a dataset including all important visual concepts.

\subsection{Label noise and ambiguities}

A fundamental challenge in dealing with user-supplied tags is that the mapping from observed tags to underlying concepts is ambiguous.  Here we discuss many types of these ambiguities that we have observed.

%Many words have multiple senses \cite{polysemousWords}, e.g., \tn{rock} can mean both ``rock-and-roll music,'' and ``rocky landscapes,'' among others.
Many terms have multiple or overlapping meanings.
The simplest case is for plurals, e.g., \tn{car} and \tn{cars}, which have different meanings but which seem to be more or less interchangeable tags on Flickr. 
Some tags have multiple distinct meanings \cite{polysemousWords}, e.g., \tn{rock} can mean both ``rock-and-roll music,'' and ``rocky landscapes.''
Trickier cases include terms like \tn{music}, \tn{concert}, and \tn{performance}, which often overlap, but often do not.  
Some words are used nearly interchangeably, such as \tn{cat} and \tn{kitten}, even though their meanings are not the same.  It seems that nearly all common tags exhibit some multiple meanings, though often one sense dominates the others.
%There are numerous other examples: \tn{bike} is both bicycle and motorbike; \tn{football} is American football and soccer, and so on.  %"me" is both "selfie" and "Maine." In many cases, one sense seems to dominate, but often not. Our algorithm seems to be picking out dominant senses. "Bike" is mostly "bicycle" but there are also some motorbikes. "Hockey" is two kinds of hockey (mostly ice) and "football" has both American and soccer.
Synonyms are also common, e.g., \tn{cat} and \tn{gato}, as well as misspellings.

Multi-word tags often occur split up, e.g., images in New York are frequently tagged as \tn{New} and \tn{York} rather than \tn{New York}. For this reason, tags like \tn{New} and \tn{San} are largely meaningless on their own. Merging these split tags (especially using cues from the other image metadata) is a natural problem for future research.

\subsection{Analysis with Ground Truth}

%\paragraph{NUS-WIDE.} 
In this section, we perform analysis using the annotated subset of the \textbf{NUS-WIDE dataset} \cite{NUS}.
This is a set of 269,642 Flickr images with annotations with both user-supplied tags, and ``ground truth'' annotations by undergraduate and high school students according to 81 concepts.  
There are a number of potential sources of noise with this dataset. Since the dataset was constructed by keyword searches, it is not an unbiased sample of Flickr, e.g., only one image in the dataset has zero keywords. Annotators were not asked to judge every image for every concept; a query expansion strategy was used to reduce annotator effort. Annotators were also asked to judge whether concepts were present in images in ways that may differ from how the images were originally tagged.

\paragraph{Tagging likelihoods.}
\label{sec:tagLikelihoods}
We now quantify the accuracy of Flickr tags.
We consider the Flickr images in NUS-WIDE that contain manual annotations, and we treat these 81 labels as ground truth, thus expanding 
%greatly
on the discussion in \cite{NUS}.  We assume an identity mapping between tags and annotations, i.e., the Flickr tag \tn{cat} corresponds to the NUS-WIDE annotation \tn{cat}.

Overall, given that a tag correctly applies to an image, 
there is empirically a 38\% chance that the uploader will actually supply it. This probability varies considerably for different tags, ranging from 2\% for \tn{person} to 94\% for \tn{cat}.  Frequently-omitted tags are often non-entry-level categories \cite{entryLevel} (e.g., \tn{person}) or
they are not an important subject in the scene \cite{importance} (e.g., \tn{clouds}, \tn{buildings}).
Given that a tag does not apply, there is a 1\% chance that the uploader supplies it anyway. 
Across the NUS-WIDE tags, this probability ranges from 2\% (for \tn{street}) to 0.04\% (for \tn{toy}).
%% This probability does not vary much across the NUS-WIDE tags; it ranges from 2\% for \tn{street} to 0.04\% for \tn{toy}.

Despite these percentages, false tags and true tags are almost equally likely, since only a few of the 81 tags correctly apply to each image. Each image has an average of 1.3 tags (of the 81), and \textit{an observed tag has only a 62\% chance of being true}. This percentage varies across different tags.

None of these numbers should be taken as exact, because the NUS annotations are far from perfect (see \supplement). Additionally, many ``false" tags are due to differences in word senses between Flickr and NUS-WIDE. For example, many \tn{earthquake} images are clearly the result of earthquakes, but are labeled as negatives in NUS-WIDE. Many \tn{cat} images that are annotated as non-\tn{cat} are images of tigers, lions, and cat costumes. Many \tn{nighttime} images were probably taken at night but indoors. 

%Overall, given that a tag is present (i.e., was supplied by the image’s uploader), there is a 62\% chance that the tag actually applies to the image. \aaron{this doesn't match the number reported in Section 5 of the NUS paper.}  Conversely, given that tag is absent, there is a 1.5\% chance that is does apply.

%These probabilities vary substantially for individual tags. Some tags are usually correct when applied: \tn{animal} (94\%), \tn{horses} (90\%), \tn{railroad} (80\%), whereas others are rarely correct: \tn{book} (12\%), \tn{castle} (20\%), \tn{earthquake} (8\%).

%Most tags are have less than 1\% probability of applying when they are absent. But there are a few exceptions: \tn{clouds} (17\%), \tn{person} (19\%), \tn{sky} (22\%). In other words, many images contain clouds or sky but are not tagged as such.

\paragraph{Tag index effects on accuracy.}
\label{sec:index}
Flickr tags are provided in an ordered list. We observed that tags earlier in the list are often more accurate than later tags, and we again treat the NUS-WIDE annotations as ground truth in order to quantify this.  

\begin{figure}
\centering
\includegraphics[width=2.7in]{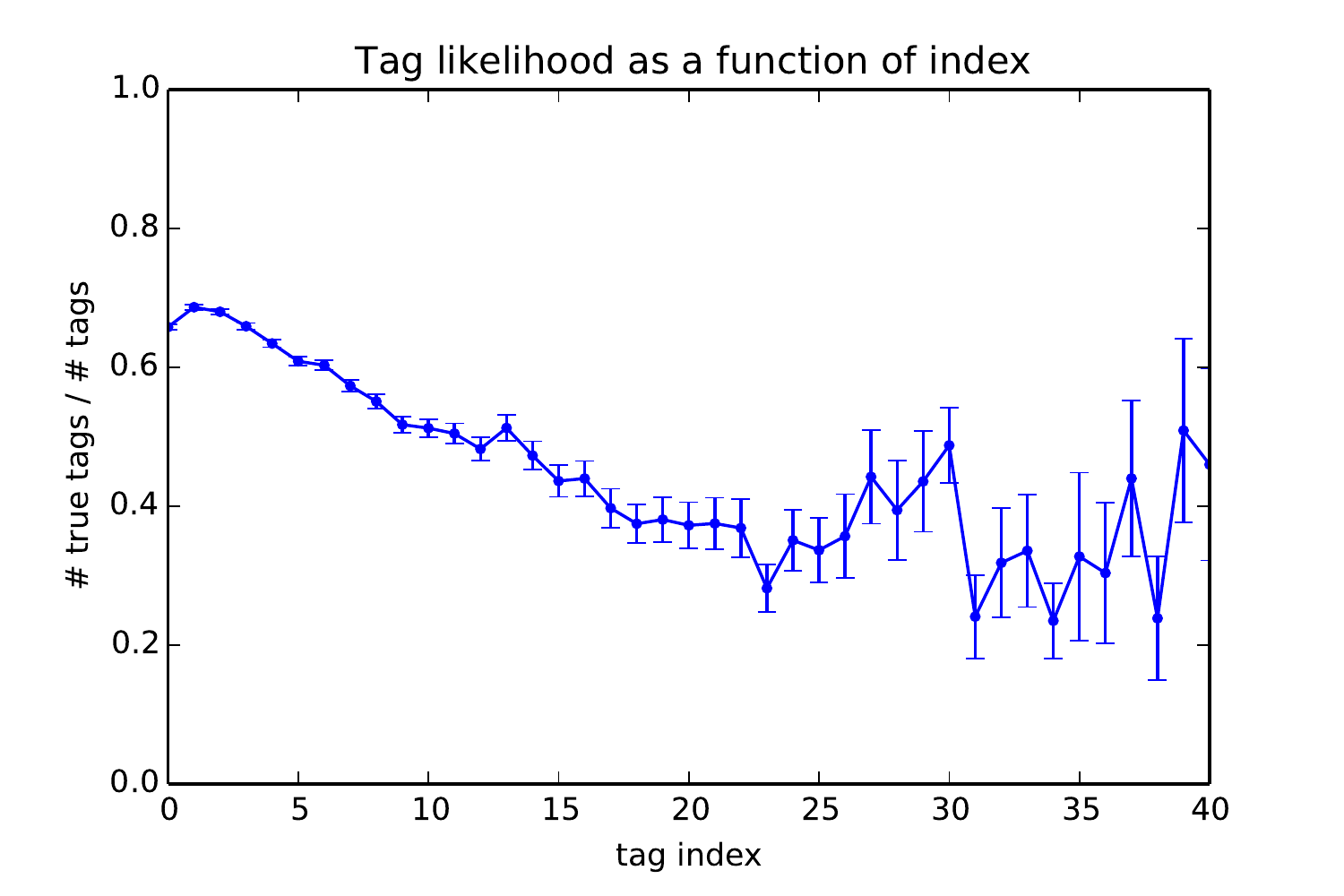}
\caption{\label{fig:tagIndex}
Empirically,  tags that occur earlier in the list of an image's tags are more likely to be accurate. This plot is computed from the NUS-WIDE dataset. (Error bars show standard error.)
\vspace{-0.1in}
}
\end{figure}

We find that the effect is substantial, as shown in Figure \ref{fig:tagIndex}. 
A tag that appears first or second in the list of tags has about 65\% chance of being accurate. A tag that occurs in position 20 or later has about a 35\% chance of being accurate.
The scales and shape of these plots also vary considerably across different tags. % (Figure \ref{fig:tagIndex}).

\paragraph{Effect of total number of tags.}
We also hypothesized that tag reliability could depend on the total number of tags provided for an image. 
This was motivated by our observation of commercially-oriented sharing sites, where uploaders are incentivized to include extraneous tags in order to boost search results. 
However, we did not find any significant effects in the Flickr data. % (Figure \ref{fig:numTags}).

%Conversely, suppose a tag is absent. Is the absence more ``accurate" for images that have more tags?  Across all tags, it does not seem to matter much how many tags. % (Figure \ref{fig:numTags}). 
%However, there are a few tags for which the effect is slightly higher, such as \tn{plants} and \tn{sky}, but the effect is small.

%In retrospect, neither result is not surprising, given the low propensity for inserting ``false'' tags (Section \ref{sec:tagLikelihoods}).

%\begin{figure}
%TBD
%\caption{\label{fig:numTags}\aaron{effect of total tag number on accuracy.}}
%\end{figure}

\section{Robust Tag Classification}
\label{sec:algorithm}

We now describe a robust classification algorithm, designed to address the following observations from the previous section: user-supplied tags often omit relevant tags, and these probabilities are different for each tag.  A conventional robust loss (e.g., Huber, Geman-McClure) would not be appropriate because of the need to set the loss function's parameters individually for each tag. The method is based on previous robust logistic regression methods \cite{raykar}. Previous approaches used batch computation, which cannot realistically be applied to millions of images; we adapt these methods to the large-scale setting using Stochastic EM \cite{Cappe:2009}.

The classifier takes as input image features $\bx$, and predicts class labels $y\in\{0,1\}$. We perform prediction for each possible tag independently, and so we consider simple binary classification in this paper.  As image features $\bx$, we use the output of the last fully-connected layer of Krizhevsky's ImageNet Convolutional Neural Network \cite{krizhevsky2012imagenet}; fc7 in the Caffe implementation \cite{jia2014caffe}.  We do not fine-tune the network parameters in this paper.

\subsection{Logistic Regression}
As our approach is based on logistic regression, we  begin by briefly
reviewing a conventional binary logistic regression classifier.  The
logistic regression model assumes that the probability of a positive
tag (i.e., the probability that $y=1$) given input features $\bx$ is a
linear function $\bw^T\bx$ passed through a sigmoid:
\begin{align}
\sigmoid(s) \equiv 1/(1+e^{-s});\quad
P(y=1 | \bx, \bw) = \sigmoid(\bw^T\bx)
%% \sigmoid(s) &\equiv 1/(1+e^{-s}) \\
%% P(y=1 | \bx, \bw) &= \sigmoid(\bw^T\bx)
\end{align}
The loss function $L(\bw)$ for a label training set $\{
(\bx_i, y_i)\}$ is the negative log-likelihood of the data:
\begin{align}
L(\bw) &= - \ln P(y_{1:N} | \bx_{1:N}, \bw) \\
&\textstyle= \sum_i (-y_i \ln \sigmoid(\bw^T\bx_i)
 -  (1-y_i) \ln(1-\sigmoid(\bw^T\bx_i)))\nonumber
\end{align}
%Using the identity $1-g(s)=e^{-s}g(s)$, this can also be rewritten as:
%\begin{align}
%L&=\sum_i \left(\ln(1+e^{-\bw^T\bx_i}) + (1-y_i) \bw^T\bx_i\right)
%\end{align}
Training entails optimizing $L$ with respect to $\bw$, using stochastic gradient descent.
%, using the gradient
%\begin{align}
%\frac{dL}{d\bw} &= \left( g(\bw^T\bx_i) - y_i \right) \bx_i
%\end{align}
Prediction entails computing the label probability $P(y|\bx,\bw)$ for a new image.

\subsection{Robust model}
As discussed in Section \ref{sec:analysis}, user-supplied tags are
often noisy. However, the logistic regression model assumes that the
observed labels $\{y_i\}$ are mostly reliable---that is, it assumes
that $y_i=1$ almost always when $\bw^T \bx_i$ is large.

To cope with this issue, we relax the assumption that the observed
training label $y$ is the true class label. We introduce a hidden
variable $z\in\{0,1\}$ representing the true (hidden) class label. We
also add a variable $\pi$ to represent the probability that a true label is added as a tag.
The model parameters are then
$\theta = \{\bw, \pi\}$, and the model is:
\begin{gather}
\label{eq:rlrpz}
P(z=1 | \bx, \bw) = \sigmoid(\bw^T\bx) \\
P(y=1 | z=1, \pi ) = \pi; \quad % &  p(y=0 | z=1 ) &= 1-\pi \\
P(y=0 | z=0 ) = 1 %\gamma  % &  p(y=1 | z=0 ) &= 1-\gamma
\end{gather}
\vspace{-0.1in}
and thus:
\begin{align}
\label{eq:rlrprob}
P(y=1|\bx,\pi, \bw) &= \pi \sigmoid(\bw^T\bx) % +(1-\gamma)(1-\sigmoid(\bw^T\bx))
\end{align}
The loss function for training is again the negative log-likelihood of the data:
\begin{align}
L(\bw,\pi) &=\textstyle \sum_i \left(-y_i \ln \pi \sigmoid(\bw^T\bx_i) \right. \\
&\hspace{0.42in}\left.- (1-y_i) \ln(1-\pi \sigmoid(\bw^T\bx_i))\right) \nonumber
\end{align}
We also experimented with a model in which false tags are occasionally added: $P(y=0|z=0) = \gamma$, where $\gamma$ is another learned parameter. We found that this model did not improve performance, and so, for clarity, we omit $\gamma$ from the rest of the paper. The $\gamma$ parameter may be useful for other datasets where users produce more spurious tags. 
Detailed derivations of the model and gradients are straightforward, and are given in the \supplement~(with $\gamma$).

%The details of these equations and their gradients (and numerically stable ways to compute them) are given in the Supplemental Material. \aaron{how much should we move to the body of the paper?}
%\aaron{remove $\gamma$ from the rest of the paper?}

\begin{figure}
\centering
\includegraphics[width=2.7in]{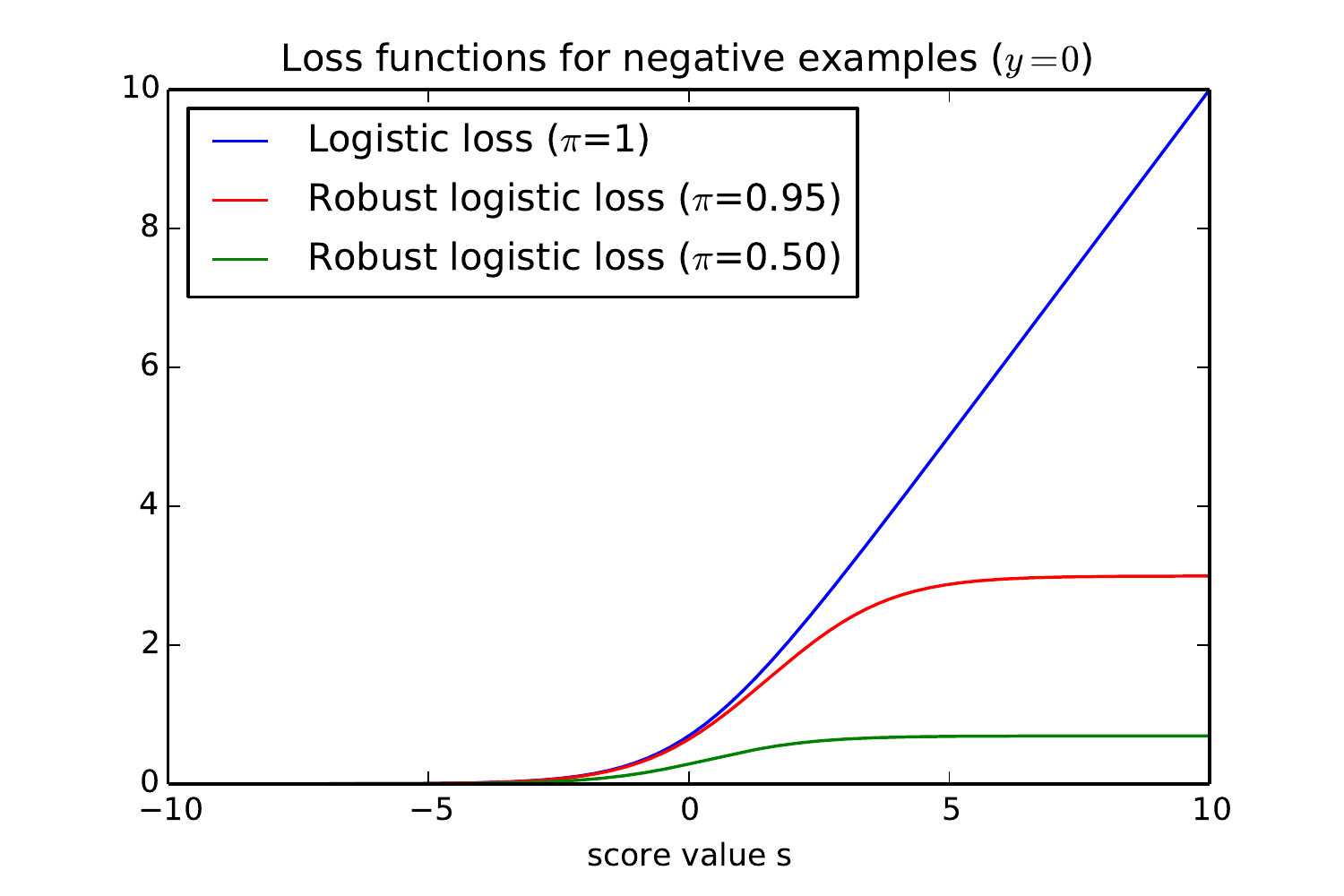}
\caption{\label{fig:lossPlot} 
Loss functions for negative examples ($y=0$). Many Flickr users omit relevant tags, which is steeply penalized by the conventional logistic loss ($\ln(1-\sigmoid(s))$).  The robust logistic loss ($\ln(1-\pi\sigmoid(s))$), is tolerant to missing labels.
}
\vspace{-0.14in}
\end{figure}

Although the loss function is unchanged for positive labels ($y=1$), the model is robust to outliers for negative examples ($y=0$); see Figure \ref{fig:lossPlot} for loss function plots.  
The classical logistic loss is unbounded,
meaning that an overly confident prediction may be heavily penalized.
With a true positive rate of $\pi=0.95$, the loss is bounded above by
$-\ln(1-\pi)\approx 3$, since no matter what the image there is
always at least a probability of $1-\pi$ of a negative label. The
impact of $\pi$ becomes smaller as the score $s=\bw^T\bx$ becomes
small, since if $s\ll 0$ then $P(z=0)\approx 1$ and $\pi$ is only
relevant when $z=1$. When the true positive rate is lower (e.g.,
$\pi=0.5$ as in Figure \ref{fig:lossPlot}), the dynamic range of the
loss function is further compressed. 

\subsection{Stochastic EM algorithm}
Learning the model for a given tag entails minimization of the loss with respect to $\bw$ and $\pi$. Stochastic gradient descent could be used for all parameters, and we provide gradients in the \supplement. However, we use Stochastic Expectation-Maximization (EM) \cite{Cappe:2009}, since
the steps are simpler to interpret and implement, and the updates to $\pi$ are numerically stable by design. All derivations and detailed versions of these equations are given in the \supplement.

Our stochastic EM algorithm applies the following steps to each minibatch:
\begin{enumerate}
\item
For each image $i$ in the minibatch, the conditional probability of the true label $z_i$ is computed as:
\begin{align}
\alpha_i &\leftarrow P(z=1|y_i,\bx_i,\bw,\pi ) \\
& = \left \{ \begin{array}{ll}
1 & y_i = 1 \\
\frac{ (1-\pi) \sigmoid(\bw^T\bx_i)} {1-\pi\sigmoid(\bw^T\bx_i)}
% (1-\pi) \sigmoid(\bw^T\bx_i) /  (1-\pi\sigmoid(\bw^T\bx_i))
& y_i =0 \end{array} \right.
%\frac{p(y_i|z=1,\pi)\ p(z_i=1|\bx_i,\bw)}{p(y_i|\bw,\bx_i,\pi)} \\
%&= \frac{ \pi^{y_i} (1-\pi)^{1-y_i} \sigmoid(\bw^T \bx_i)}
%{ \pi^{y_i} (1-\pi)^{1-y_i} \sigmoid(\bw^T \bx_i) + 
%}
\end{align}
%where $\theta = \{\bw, \pi, \gamma\}$ denotes the current model parameters.
\item 
We define the sufficient statistics for the minibatch as
\begin{equation}
\begin{split}
\textstyle S^{\mathrm{mb}}_\alpha\equiv \sum_i \alpha_i/N
;\quad S^{\mathrm{mb}}_{y\alpha}\equiv \sum_i y_i\alpha_i / N,
\end{split}
\end{equation}
where $N$ is the number of datapoints in the summation.  Estimates of
the average sufficient statistics for the full dataset are updated
with a step size $\eta$:
\begin{align}
S^{\mathrm{ds}} &\textstyle\leftarrow 
(1-\eta)S^{\mathrm{ds}} + \eta S^{\mathrm{mb}} %\eta_t \sum_i \alpha_i / N,
%% S_\alpha &\textstyle\equiv \sum_i \alpha_i / N,
%S_y \equiv \sum_i y_i /N \\
%S^{\mathrm{ds}}_{y\alpha} &\textstyle\leftarrow (1-\eta_t)S^{\mathrm{ds}}_{y\alpha} + \eta_t S^{\mathrm{mb}}
\end{align}
In our experiments, we initialized $S^{\mathrm{ds}}_{\alpha}$ and
$S^{\mathrm{ds}}_{y\alpha}$ to 1 and used a fixed step size of
$\eta=0.01$.

%% S^{\mathrm{ds}}_\alpha &\textstyle\leftarrow 
%% (1-\eta_t)S^{\mathrm{ds}}_\alpha + \eta_t S^{\mathrm{mb}} %\eta_t \sum_i \alpha_i / N,
%% %% S_\alpha &\textstyle\equiv \sum_i \alpha_i / N,
%% \\
%% %S_y \equiv \sum_i y_i /N \\
%% S_{y\alpha} &\textstyle\leftarrow (1-\eta_t)S_{y\alpha} + \eta_t \sum_i y_i\alpha_i /N
%% \end{align}

%\begin{align}
%S_\alpha &\equiv \sum_i \alpha_i & S_y &\equiv \sum_i y_i \\
%S_{y\alpha} &\equiv \sum_i y_i\alpha_i & S_1 &\equiv \sum_i 1 
%\end{align}

\item $\pi$ is computed from the current estimate of the sufficient statistics,
so that $\pi$ is an estimate of the percentage of true labels that were actually supplied as
tags:
\begin{align}
\pi &\leftarrow S^{\mathrm{ds}}_{y\alpha} / S^{\mathrm{ds}}_\alpha
%% \pi &\leftarrow \frac{\sum_i y_i \alpha_i}{\sum_i \alpha_i}  = \frac{S_{y\alpha}}{S_\alpha} 
%\gamma &\leftarrow 
%\frac{\sum_i (1-y_i)(1-\alpha_i)}{\sum_i (1-\alpha_i)} 
%= \frac{1 - S_y - S_{\alpha} + S_{y\alpha}}{1 - S_\alpha}
\end{align}

\item 
The weights $\bw$ are updated using stochastic gradient on $L$.
It is straightforward to verify that the gradient w.r.t. $\bw$ is 
\vspace{-0.1in}
\begin{align}
\textstyle\frac{dL}{d\bw} = \sum_i (\sigmoid(\bw^T\bx_i) - \alpha_i)\ \bx_i.
\end{align}
\end{enumerate}

\subsection{Calibration}
\label{sec:calibration}

In many cases, we would like to predict the true class probabilities
$P(z|\bx)$.  
Well-calibrated estimates of these probabilities are particularly
useful in applications where it is important to weight the importance
of multiple tags for an image, such as when trying to retrieve images
characterized by multiple tags or when choosing a small number of tags
to apply to an image \cite{Scheirer:2012}.

In theory, the robust model above could learn a well-calibrated estimate of $P(z|\bx)$.
However, the model still makes strong simplifying assumptions---for example, it assumes linear decision boundaries, and that label noise is independent of image content.
To the extent that these assumptions are unrealistic, the model may benefit from an additional calibration step.

We tried to apply the calibration method from \cite{Scheirer:2012},
but found that it degraded the logistic regression model's
performance. This may be because it is designed to address
miscalibration due to using non-probabilistic classifiers such as
SVMs, rather than due to label noise.

Instead, we propose the following strategy.  First, a model is learned
with the large dataset.  The weight vector $\bw$ is then held fixed
for each tag.  However, an intercept $\beta$ is added to the model, so
that the new class probability is
\begin{align}
P(z=1 | \bx, \bw, \beta) &= \sigmoid(\bw^T\bx + \beta)  \label{eq:calibzpred}
\end{align}
The intercept allows the model to adjust the prior probability of each class in the new dataset.
Then, we continue training the model on a small curated dataset (treating it as ground truth $z$), but only update the $\beta$ parameters. % and leave the weight vector $\bw$ unchanged. %\aaron{what do we do with $\pi$?}
Very little curated data is necessary for this process, since only one new parameter is being estimated per tag.

In our experiments, we simulate this procedure by training on the F100M data and
calibrating on a subset of NUS-WIDE annotations. More general domain
adaptation methods (e.g., \cite{hoffmanDA}) could also be used.

%Ideally, one would simply retrain the entire model on the new dataset.  However, the advantage of this procedure over simply retraining the entire model on the new dataset is that it can potentially require very little curated data, as there is only one free parameter $\beta$ per tag.

\begin{figure}
\centering
\includegraphics[width=3in]{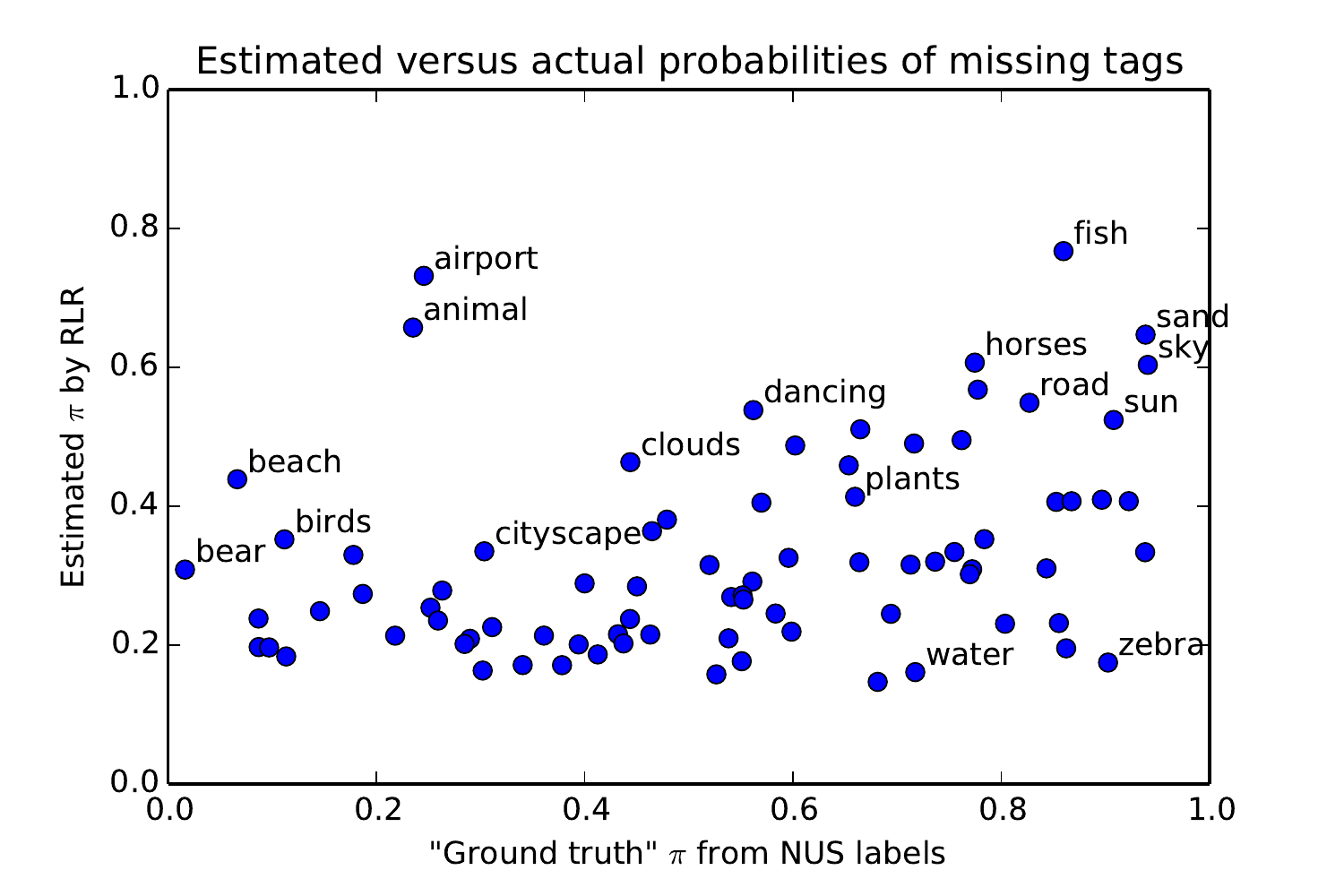}
\caption{\label{fig:pi_scatterplot}
Tagging likelihoods $\pi$ estimated from Flickr images with RLR, versus estimation from the ``ground truth" NUS annotations. The likelihoods are correlated ($r=0.34$), though the tagging likelihood is mostly underestimated, probably due to inaccuracies in both the predictor and the annotations.
}
\vspace{-0.05in}
\end{figure}

\begin{table}
\centering
\begin{tabular}{l@{~~~}c@{~~~}c@{~~~}c}
\hline
& Recall & Precision & F-score \\
\hline
LR & 9.7	&7.9&	8.7 \\
RLR & 11.7 &	8.0	&9.5 \\
\hline
\end{tabular}
\caption{Flickr tag prediction results. Robust logistic regression improves over logistic regression's ability to predict which tags a user is likely to apply to an image.
\label{table:tagPrediction}
\vspace{-0.05in}
}
\end{table}

% \vspace{-0.15in}

\section{Experiments}

\begin{figure*}
\centering
\includegraphics[width=5.5in]{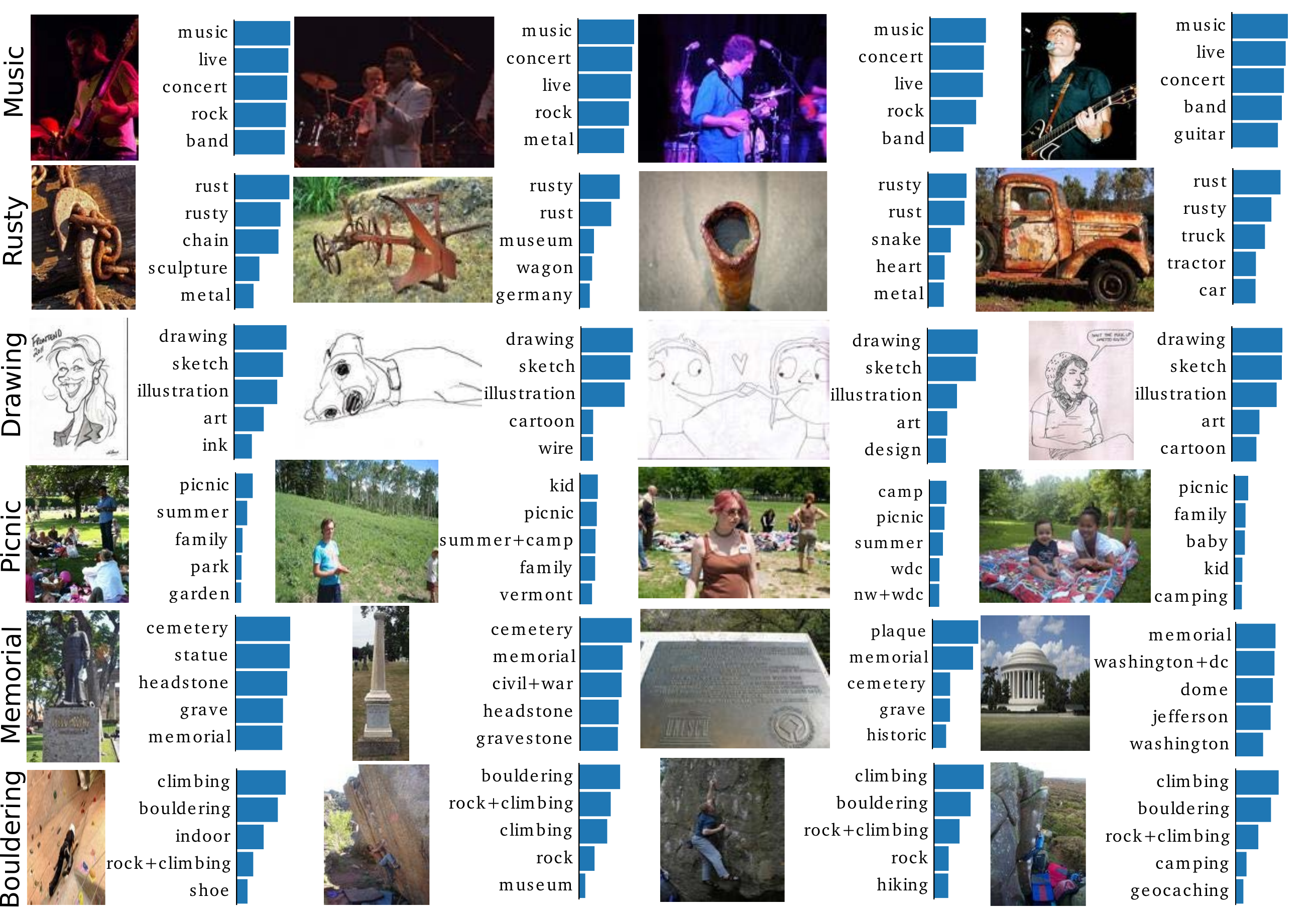}
\caption{Single-tag retrieval results, and automatically-generated
  annotations. None of the query tags are in NUS-WIDE, and most
  (\tn{music}, \tn{rusty}, \tn{drawing}, \tn{bouldering}) are also
  absent from ImageNet.  Many of the annotations are also absent from the other datasets as well.
\label{fig:flickrretrieval}
}
\vspace{-0.15in}
\end{figure*}

\begin{table}
\centering
\begin{tabular}{l@{~~~}c@{~~~}c@{~~~}c@{~~~}c}
\hline
& Recall & Precision & F-score \\
\hline
CNN+WARP \cite{google_rank_loss}  & 52.0 & 22.3 & 31.2 \\
NUS, LR  & 58.2 & 26.1 & 36.0 \\
\hline
F100M, LR  & 58.4 & 21.7 & 31.6 \\
F100M, RLR  & 58.0 & 22.3 & 32.3 \\
\hline
F100M, LR, Calib  & 42.5 & 32.2 & 36.6 \\
F100M, RLR, Calib  & 44.2 & 31.3 & \textbf{36.7} \\
\hline
\end{tabular}
\caption{Image annotation results, illustrating how the freely-available user-supplied tags can augment or supplant costly manual annotations.
Testing is performed on the NUS-WIDE test set.
The first two rows show training only on the NUS-WIDE training set with logistic regression, and the previously-reported state-of-the-art \cite{google_rank_loss}.
Each of the remaining rows is trained on F100M with either LR or Robust LR. The third and fourth rows are also calibrated on the NUS test set. All scores are predictions-at-5.
\vspace{-0.14in}
%\matt{Does simple logistic regression really dominate CNN+WARP? Should we say something about that?}
%\aaron{Hamid: mAPs after calibration?  How are mAPs computed?  Should we omit them? Maybe discuss with Matt.}
 %These results illustrate that good
%results can be obtained from user-supplied labels alone, and these models can be improved with just a few supervised labels. These results are competitive with acquiring high-quality annotations, which can be very expensive.
\label{fig:NUSscores}
}
\end{table}

%CNN+WARP \cite{google_rank_loss} & N/A & 52.03 & 22.31 & 31.23 \\
%NUS, LR & 44.54 & 58.21 & 26.1 & 36.04 \\
%\hline
%F100M, LR & 35.86 & 58.42 & 21.66 & 31.6 \\
%F100M, RLR & 37.11 & 58.04 & 22.36 & 32.28 \\
%\hline
%F100M, LR, Calib & --- & 42.52 & 32.18 & 36.63 \\
%F100M, RLR, Calib & --- & 44.24 & 31.28 & 36.65 \\

We now describe experiments to test models learned from F100M on several  tasks, including tag prediction, image annotation, and image retrieval with one or more tags. 

All training is performed using Caffe \cite{jia2014caffe}, running for 20,000 minibatches, with minibatch size of 500 images.  Training is performed on a GeForce GTX780 GPU. Each minibatch takes 2 seconds, and a complete run takes 11 hours.
Based on the observations in Section \ref{sec:index}, we only keep the first 20 tags in all Flickr images in our experiments.
We use a subset of 4,768,700 images from F100M as training set and hold out another 200,000 for testing.  
The sets are split by user ID in order to ensure that images from the same user do not occur in both sets.
Plural and singular tags are combined using WordNet's lemmatization.  

\arxivcond{}{Interactive annotation and retrieval demos can be found online at \url{http://54.244.94.90}.}

\subsection{Tag prediction}
\label{sec:tagprediction}
We first test the following prediction task: given a new image, what tags would a user be likely to apply to this image?  This task could be useful for consumer applications, for example, auto-suggesting tags for users when sharing their images. Note that this task is different from ground-truth prediction;
we want to suggest tags that are both objectively accurate and likely to be applied by a user.

We trained a logistic regression baseline and a robust logistic
regression model on our 4.7M-image F100M training set, 
and evaluated
the models' ability to annotate images in the 200K-image F100M test
set. 

For each test image, the model predicts the probability of each tag occurring: $P(y=1|\bx,\bw,\pi)$. 
%Tags are then ranked by these probabilities.
(Note that for robust logistic regression, this is Equation \ref{eq:rlrprob},
%not $p(z=1|\bx,\bw)$ from Equation \ref{eq:rlrpz}, 
since we want to predict tagging behavior $y$, not ground truth
$z$.) The final annotations were produced by selecting the top 5
most likely tags for each image.

\begin{figure*}
\centering
\includegraphics[width=5.7in]{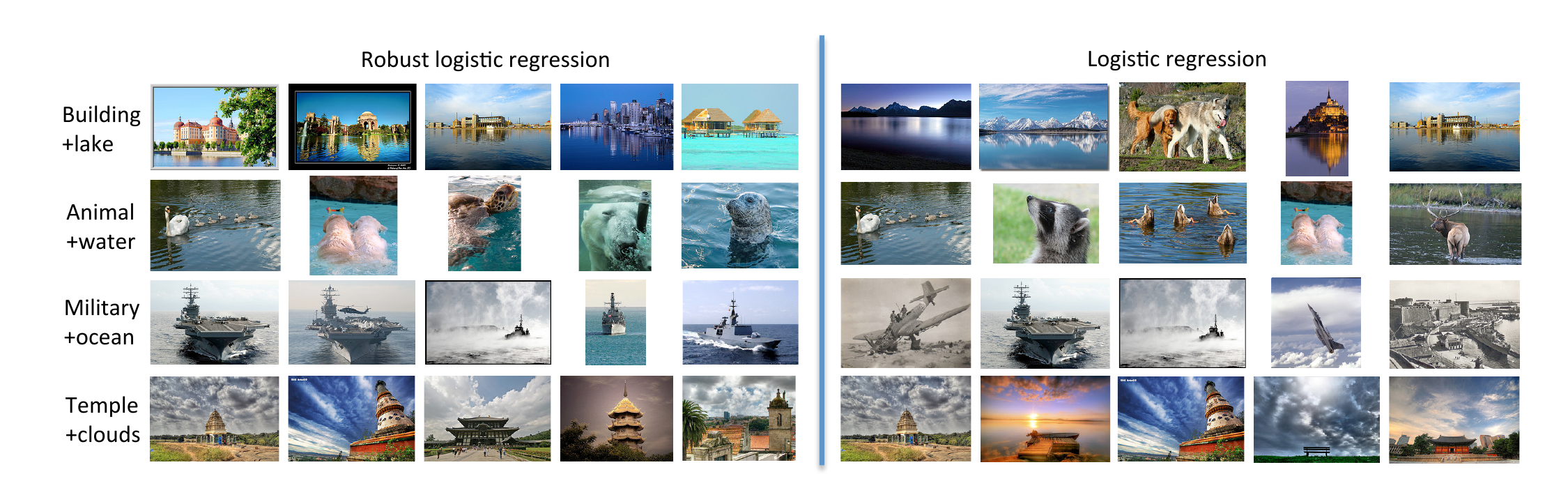}
\caption{Multi-tag retrieval queries where Robust LR gives notably superior results to LR. Retrieval results are sorted from left-to-right. % in each case.
%\aaron{are there similar cases where RLR is just as bad?}
\label{fig:multiretrieval}
\vspace{-0.15in}
}
\end{figure*}

We evaluate overall precision and recall at 5 for each image, averaged
over all images. We also compute the F-score, which is the harmonic
mean of the average precision and recall.  Table
\ref{table:tagPrediction} summarizes the results. RLR achieves higher
recall without sacrificing precision.
%% , indicating the
%% benefit of using a model that is robust to label noise.
Figure
\ref{fig:flickrretrieval} shows some qualitative results of calibrated
RLR's ability to predict tags for images from the test
set. 
Figure \ref{fig:pi_scatterplot} compares RLR's estimated values of $\pi$ for each tag, versus the NUS annotations estimated in Section \ref{sec:tagLikelihoods}. 
RLR's estimates are correlated with the NUS ground truth, but discrepancies are common.
\subsection{Image annotation}

\begin{figure}
\centering
\includegraphics[width=2in]{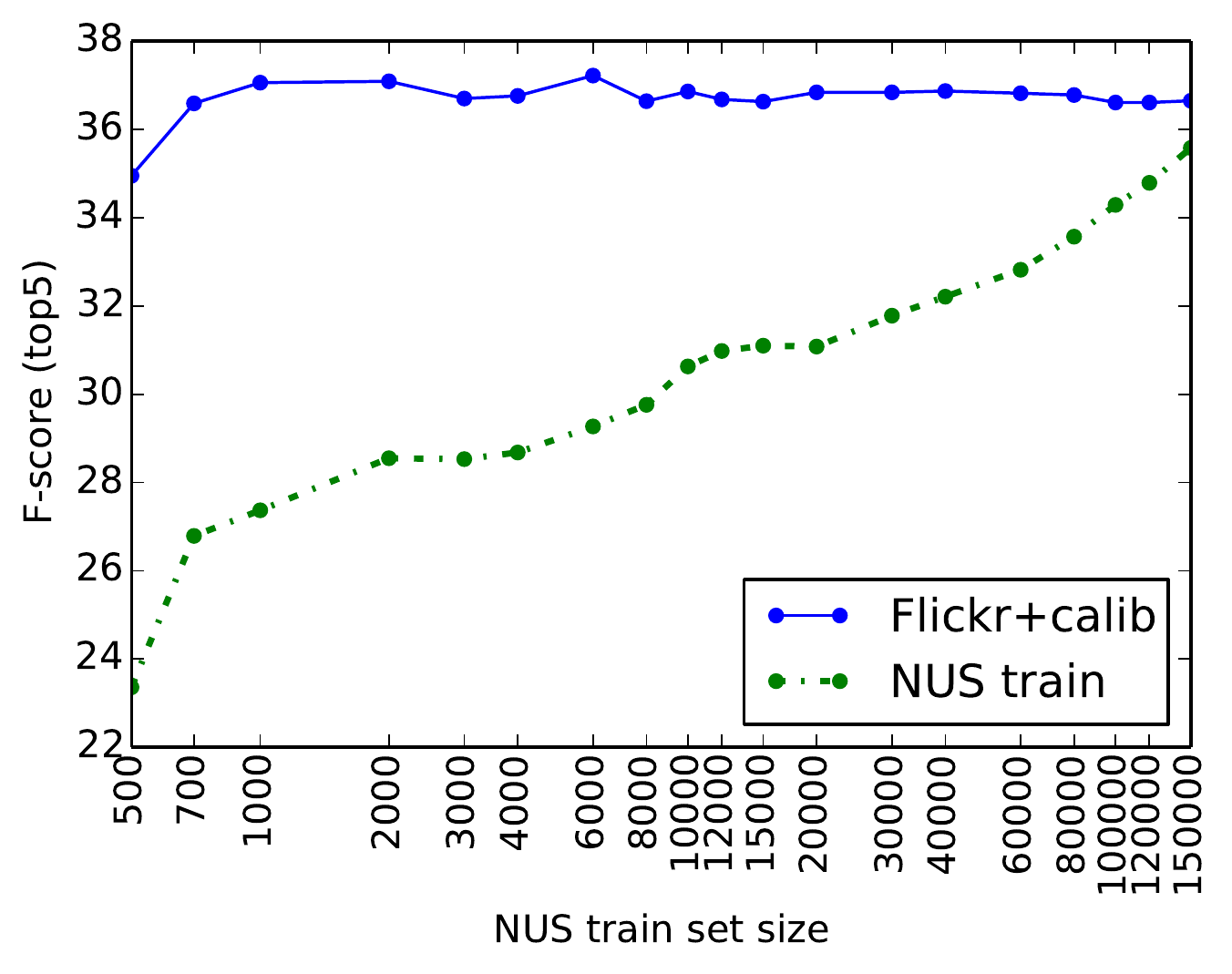}
\caption{\label{fig:calibSetSize}
Effect of calibration set size on image annotation score. 
Training on user-supplied tags and then calibrating on a small subset of manual annotations can outperform
the costly process of obtaining many manual annotations: \textbf{the annotation cost can be reduced by a factor of 200, while obtaining the same results.}
%The blue curve shows LR training on F100M followed by calibration on a subset of the NUS training, whereas the green curve shows LR training on the NUS subset only.
%\aaron{x-axis should read: "NUS training set size". make text bigger so we can shrink the plot.}
%\aaron{maybe make this plot friendly for colorblind readers}
\vspace{-0.1in}
}
\end{figure}

We next test the task: given an image, which labels objectively apply to it?
%We next evaluate our models' ability to predict which labels apply objectively to an image.
%the ability of models trained from user-supplied
%labels to estimate what tags objectively apply to an image. 
We use the same F100M training set as above, but evaluate on the 81 labels in the 
manually annotated NUS-WIDE dataset, treating the NUS-WIDE annotations as ground truth.
We also compare to models trained on NUS-WIDE.
%\aaron{discuss how we map from NUS to Flickr tags.}

We evaluate per-tag precision and recall averaged over tags. 
For a tag $j$, per-tag precision is defined as $N^c_j / N^p_j$
and per-tag recall is defined as $N^c_j / N^g_j$,
%% These
%% metrics are defined for a tag $j$ as
%% \begin{equation}
%% \begin{split}
%% \textrm{Per-tag precision}=N^c_j / N^p_j
%% ;\quad
%% \textrm{Per-tag recall}=N^c_j / N^g_j,
%% \nonumber
%% \end{split}
%% \end{equation}
where $N^p_j$ is the number of images that the system annotated with
tag $j$, $N^c_j$ is the number of images that a user annotated with
tag $j$ that the system also annotated with tag $j$, and $N^g_j$ is
the number of images in the test set that a user annotated with tag
$j$. Per-tag precision is undefined if a tag is never used by the
system; when this happens we define the precision to be 0.  We also
computed the per-tag F-score.
%% We use the same %average per-tag precision, recall, and F-score
%% metrics as for tag prediction. % in Section \ref{sec:tagprediction}. 
To predict annotations with RLR, we predict $z$, not $y$ (Equation \ref{eq:rlrpz} or \ref{eq:calibzpred}).
Scores are reported in Table \ref{fig:NUSscores}.

Testing and training LR on NUS data produces somewhat better scores than training on F100M alone; it also produces better scores than the reported state-of-the-art on this dataset \cite{google_rank_loss}.
We get the best scores by training on F100M and then calibrating on the NUS training set (Section \ref{sec:calibration}).

It is important to consider the cost of annotated labels. The user-supplied tags in F100M are basically free, whereas obtaining manual annotations is a very costly process.
We compare training on a subset of NUS training annotations, versus F100M training plus calibration with the same NUS subset.
As shown in Figure \ref{fig:calibSetSize},
\textit{the calibration process can yield scores superior to training on the full annotation set, but with a 200x reduction in annotation cost.}

\begin{table}
\centering
\begin{tabular}{l@{~~~}c@{~~~}c@{~~~}c@{~~~}c}
\hline
%&\multicolumn{3}{c}{Precision at 5 for}\\
& 1 Tag & 2 Tags & 3 Tags \\
\hline
NUS, LR  & \emph{81} & \emph{17.9} & \emph{9.1} \\
\hline
F100M, LR  & 70.1 & 8.5 &2.3 \\
F100M, RLR  & \textbf{71.9} & 9.2 & 2.7 \\
\hline
F100M, LR, Calib  & 70.1 & 10.3 & 3.6 \\
F100M, RLR, Calib  & \textbf{71.9} & \textbf{11} & \textbf{3.9} \\
\hline
\end{tabular}
\caption{Image retrieval results, showing precision at 5 for multi-tag retrieval.
% illustrating the value of Robust LR and calibration for multi-tag queries.
Testing is performed on the NUS-WIDE test set.
Columns show performance for each method for the number of tags that need to be matched.
%Rows show the performance of different methods, and columns show performance for each method as a function
%of how many tags need to be matched).
See the caption to Table \ref{fig:NUSscores} for an explanation of the rows.
%The first row shows the results of training only on the NUS-WIDE training set with 
%logistic regression.
%Each of the remaining rows is trained on F100M with either LR or Robust LR. The third and four rows are also calibrated on the NUS test set.
Robust LR consistently outperforms LR, and calibration consistently improves results.
These trends are clearer for longer (and therefore more difficult) queries.
\label{fig:retrievalscores}
\vspace{-0.15in}
}
\end{table}

%We evaluated logistic
%regression and robust logistic regression models trained in several
%ways. First, we trained the models on various fractions of the
%NUS-WIDE training set to test how performance varies as a function of
%training data. Next, we trained the models on the F100M training set.
%Finally, we took the Flickr-trained models and calibrated them by
%fitting an intercept term as in \ref{sec:calibration}. This
%calibration was also done using varying-size subsets of the NUS-WIDE
%training set to evaluate how much ground-truth data we need to
%effectively calibrate the Flickr-trained models. We also evaluated the
%extreme-value-theory-based calibration method of Scheirer et al.
%\cite{Scheirer:2012}. \matt{details on EVT method.}

%\aaron{report overall recall for comparison with Google paper?}

\subsection{Image retrieval}

Finally, we consider the  tag-based image retrieval task:
given a set of query tags, find images that match all the tags.
We measure performance using normalized precision at 5; each system
returns a set of 5 images, and its score for a given query is the
number of those images that are characterized
by all tags divided by the smaller of 5 and the total number of
relevant images in the test set. We use the NUS-WIDE annotations
as ground truth.
We tested the same models from the previous section.
%We trained five models: LR on the NUS training set, LR and RLR on
%the F100M training set discussed above, and LR and RLR calibrated
%to the NUS training set.
% We also tested Weibull-calibrated LR and RLR \cite{evt}.
We tested each method with queries consisting of every combination of
one, two, and three tags that had at least one relevant image in the
test set. Scores are shown in 
Table \ref{fig:retrievalscores}.% shows normalized
%precision-at-5 scores for each method, averaged over each query.

All models perform well on single-tag queries, but the differences in
precision grow rapidly as the number of tags that the retrieved images
must match increases. RLR consistently
outperforms LR, and calibration significantly improves the models
trained on Flickr.
Figure \ref{fig:multiretrieval} shows some queries for which RLR
outperforms LR.

The model trained on NUS-WIDE achieves the best score. %, which confirms that for this task there is significant
%value to making full use of manually annotated data where it is
%available. 
However, there are many thousands of tags for which
no annotations are available, and these results show that good results can be obtained on these tags as well.
%for the thousands of tags for which no such data
%is available so that we must fall back on user-supplied training data,
%these results show that we still can obtain good results with little or no annotated data.
% that the RLR model will outperform the LR model.

%% We find that calibration is particularly important for the multi-tag
%% queries. An image's score for a multi-tag query is the model's
%% estimate of the log-probability that all of the query tags apply to
%% the image, which is a sum of nonlinear log-logistic functions. This
%% nonlinearity makes calibration necessary to avoiding over- or
%% under-emphasizing the importance of some tags in the query.

%% \aaron{cherrypicked results in Figure \ref{fig:multiretrieval}}

\section{Discussion and Future Work}

Online user-supplied tags represent a great, untapped natural resource. We show that, despite their noise, these tags can be useful, either on their own or in concert with a small amount of calibration data. %We also demonstrate an improved loss function for noisy tags.
Though we have tested the Flickr dataset, there are numerous other online datasets with different kinds of user-supplied tags that can also be leveraged and explored for different applications.
As noted in Section \ref{sec:analysis}, there is a great deal of structure in these tags that could be exploited in future work. Our work could be combined with methods that model the relationships between tags, as well as improved CNN models and fine-tuning.  These tags could also provide mid-level features for other classification tasks and consumer applications, such as tag suggestion and organizing personal photo collections.

{\small
\bibliographystyle{ieee}
\bibliography{paper}

\begin{thebibliography}{10}\itemsep=-1pt

\bibitem{importance}
A.~C. Berg, T.~L. Berg, H.~{Daum\'{e} III}, J.~Dodge, A.~Goyal, X.~Han,
  A.~Mensch, M.~Mitchell, A.~Sood, K.~Stratos, and K.~Yamaguchi.
\newblock Understanding and predicting importance in images.
\newblock In {\em Proc.~CVPR}, 2012.

\bibitem{Cappe:2009}
O.~Capp{\'e} and E.~Moulines.
\newblock On-line expectation-maximization algorithm for latent data models.
\newblock {\em Journal of the Royal Statistical Society: Series B (Statistical
  Methodology)}, 71(3):593--613, 2009.

\bibitem{Chatfield12}
K.~Chatfield and A.~Zisserman.
\newblock Visor: Towards on-the-fly large-scale object category retrieval.
\newblock In {\em Asian Conference on Computer Vision}, Lecture Notes in
  Computer Science. Springer, 2012.

\bibitem{neil}
X.~Chen, A.~Shrivastava, and A.~Gupta.
\newblock Enriching visual knowledge bases via object discovery and
  segmentation.
\newblock In {\em Proc.~CVPR}, 2014.

\bibitem{NUS}
T.-S. Chua, J.~Tang, R.~Hong, H.~Li, Z.~Luo, and Y.~Zheng.
\newblock {NUS-WIDE}: a real-world web image database from {N}ational
  {U}niversity of {S}ingapore.
\newblock In {\em Proc.~CIVR}, 2009.

\bibitem{attr09}
A.~Farhadi, I.~Endres, D.~Hoiem, and D.~Forsyth.
\newblock Describing objects by their attributes.
\newblock In {\em Proc.~CVPR}, 2009.

\bibitem{visualCategoryFilter}
R.~Fergus, P.~Perona, and A.~Zisserman.
\newblock A visual category filter for google images.
\newblock In {\em Proc.~ECCV}, 2004.

\bibitem{google_rank_loss}
Y.~Gong, Y.~Jia, T.~K. Leung, A.~Toshev, and S.~Ioffe.
\newblock Deep convolutional ranking for multilabel image annotation.
\newblock In {\em Proc.~ICLR}, 2014.

\bibitem{gongEmbedding}
Y.~Gong, L.~Wang, M.~Hodosh, J.~Hockenmaier, and S.~Lazebnik.
\newblock Improving image-sentence embeddings using large weakly annotated
  photo collections.
\newblock In {\em Proc.~ECCV}, 2014.

\bibitem{hoffmanDA}
J.~Hoffman, E.~Rodner, J.~Donahue, K.~Saenko, and T.~Darrell.
\newblock Efficient learning of domain-invariant image representations.
\newblock In {\em Proc.~ICLR}, 2013.

\bibitem{grauman_tags}
S.~J. Hwang and K.~Grauman.
\newblock Reading between the lines: Object localization using implicit cues
  from image tags.
\newblock {\em PAMI}, June 2012.

\bibitem{jia2014caffe}
Y.~Jia, E.~Shelhamer, J.~Donahue, S.~Karayev, J.~Long, R.~Girshick,
  S.~Guadarrama, and T.~Darrell.
\newblock Caffe: Convolutional architecture for fast feature embedding, 2014.
\newblock http://arxiv.org/abs/1408.5093.

\bibitem{karayev}
S.~Karayev, M.~Trentacoste, H.~Han, A.~Agarwala, T.~Darrell, A.~Hertzmann, and
  H.~Winnemoeller.
\newblock Recognizing image style.
\newblock In {\em Proc.~BMVC}, 2014.

\bibitem{krizhevsky2012imagenet}
A.~Krizhevsky, I.~Sutskever, and G.~E. Hinton.
\newblock {ImageNet Classification with Deep Convolutional Neural Networks}.
\newblock In {\em NIPS}, 2012.

\bibitem{optimol}
L.-J. Li and L.~Fei-Fei.
\newblock {OPTIMOL}: Automatic object picture collection via incremental model
  learning.
\newblock {\em IJCV}, 88:147--168, 2010.

\bibitem{wordnet}
G.~A. Miller.
\newblock Wordnet: A lexical database for english.
\newblock {\em Commun.~ACM}, 38(11), 1995.

\bibitem{Murray-CVPR-2012}
N.~Murray, D.~Barcelona, L.~Marchesotti, and F.~Perronnin.
\newblock {AVA: A Large-Scale Database for Aesthetic Visual Analysis}.
\newblock In {\em CVPR}, 2012.

\bibitem{entryLevel}
V.~Ordonez, J.~Deng, Y.~Choi, A.~C. Berg, and T.~L. Berg.
\newblock From large scale image categorization to entry-level categories.
\newblock In {\em Proc.~ICCV}, 2013.

\bibitem{SUNattributes}
G.~Patterson, C.~Xu, H.~Su, and J.~Hays.
\newblock The sun attribute database: Beyond categories for deeper scene
  understanding.
\newblock {\em IJCV}, 108:59--81, 2014.

\bibitem{raykar}
V.~C. Raykar, S.~Y, L.~H. Zhao, G.~H. Valadez, C.~Florin, L.~Bogoni, and
  L.~Moy.
\newblock Learning from crowds.
\newblock {\em JMLR}, 11:1297--1322, 2010.

\bibitem{ImageNet}
O.~Russakovsky, J.~Deng, H.~Su, J.~Krause, S.~Satheesh, S.~Ma, Z.~Huang,
  A.~Karpathy, A.~Khosla, M.~Bernstein, A.~C. Berg, and L.~Fei-Fei.
\newblock Imagenet large scale visual recognition challenge, 2014.
\newblock http://arxiv.org/abs/1409.0575.

\bibitem{polysemousWords}
K.~Saenko and T.~Darrell.
\newblock Unsupervised learning of visual sense models for polysemous words.
\newblock In {\em Proc.~NIPS}, 2008.

\bibitem{Scheirer:2012}
W.~J. Scheirer, N.~Kumar, P.~N. Belhumeur, and T.~E. Boult.
\newblock Multi-attribute spaces: Calibration for attribute fusion and
  similarity search.
\newblock In {\em The IEEE Conference on Computer Vision and Pattern
  Recognition (CVPR)}, June 2012.

\bibitem{Schroff11}
F.~Schroff, A.~Criminisi, and A.~Zisserman.
\newblock {H}arvesting {I}mage {D}atabases from the {W}eb.
\newblock {\em IEEE Transactions on Pattern Analysis and Machine Intelligence},
  33(4):754--766, apr 2011.

\bibitem{welinder}
P.~Welinder, S.~Branson, S.~Belongie, and P.~Perona.
\newblock The {M}ultidimensional {W}isdom of {C}rowds.
\newblock In {\em Proc. NIPS}, 2010.

\bibitem{SUN}
J.~Xiao, J.~Hays, K.~Ehinger, A.~Oliva, and A.~Torralba.
\newblock Sun database: Large-scale scene recognition from abbey to zoo.
\newblock In {\em Proc.~CVPR}, 2010.

\bibitem{zhou2014places}
B.~Zhou, A.~Lapedriza, J.~Xiao, A.~Torralba, and A.~Oliva.
\newblock {Learning Deep Features for Scene Recognition using Places Database.}
\newblock In {\em Proc.~NIPS}, 2014.

\bibitem{zhu_refinement}
G.~Zhu, S.~Yan, and Y.~Ma.
\newblock Image tag refinement towards low-rank, content-tag prior and error
  sparsity.
\newblock In {\em Proc.~MM}, 2010.

\end{thebibliography}
}

%%%%%%%%% TITLE
\title{Appendix 1:\\
Image Classification and Retrieval from User-Supplied Tags}

\author{} 

\maketitle
\end{comment}

\graphicspath{{figures_supp/}}

\appendix

%\thispagestyle{empty}

% %%%%%%%%%%%%%%%%%%%%%%%%%%%%%%%%%%%% ABSTRACT %%%%%%%%%%%%%%%%%%%%%%%%%%%%%%%%%%%%%
% \begin{abstract}
% 
% \end{abstract}

%%%%%%%%%%%%%%%%%%%%%%%%%%%%%%%%%%%% BODY TEXT %%%%%%%%%%%%%%%%%%%%%%%%%%%%%%%%%%%%

\arxivcond{}{  % ARXIV
\section{Online supplemental material}

You may browse retrieval results for any of the Flickr tags we studied at the project site:
\url{http://54.244.94.90}. It is quite fun and enlightening to explore the range of tags that can be learned from raw Flickr tags.
%We will not make any attempt to check web visitor logs. You may wish to view submission websites with an anonymous browser. We have tested our demo page successfully with hideme.be on Chrome. (Anonymouse seemed to be incompatible with our ``infinite-scrolling" results pages.)
}

\section{Details of correspondence calculation}
\label{sec:appCalc}

Here we explain how we estimated the percentage of Flickr tags absent from ImageNet concepts (Section 3.2 of the submission).
We collected the top 1000 Flickr tags, and manually filtered out non-image and location tags, with 612 tags remaining. 
We determined an automatic mapping from Flickr tags to WordNet synsets, by mapping each tag to its top WordNet noun synset, and manually corrected mismatches in the top 100 Flickr tags. We call an ImageNet synset \textit{large} if it has 1000 of more node in the subtree.
Of the top 100 Flickr tags, we found that 54 of them had large ImageNet subtree before correcting mismatches, and 62 had large subtrees after manual corrections.
Of the remaining 512 uncorrected tags, 189 (37\%) have large subtrees. Linear extrapolation suggests that 
$1- ((189.0 * (62.0/54.0)) + 62)/612=54\%$ of tags are missing ImageNet subtrees. Of course, there are a number of questionable assumptions in this model, e.g., 1000 images may not be enough images for many classes, such as \tn{art}.

\section{Issues with NUS-WIDE annotation}
\label{sec:NUSAnnotIssue}
\begin{figure}[th]

\begin{subfigure}{.5\textwidth}
	\includegraphics[width=\textwidth]{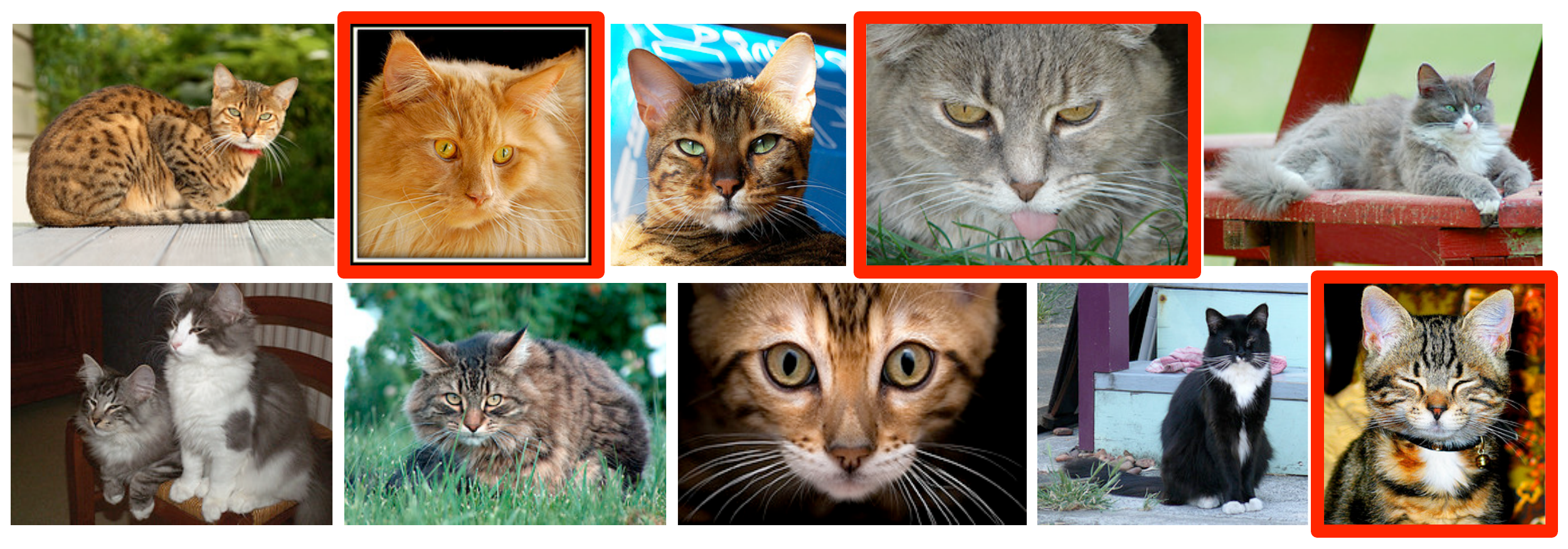}
	\vspace{-0.25in}
	\caption{Cat}
\end{subfigure} \\
\vspace{-.1in}
\begin{subfigure}{.5\textwidth}
	\includegraphics[width=\textwidth]{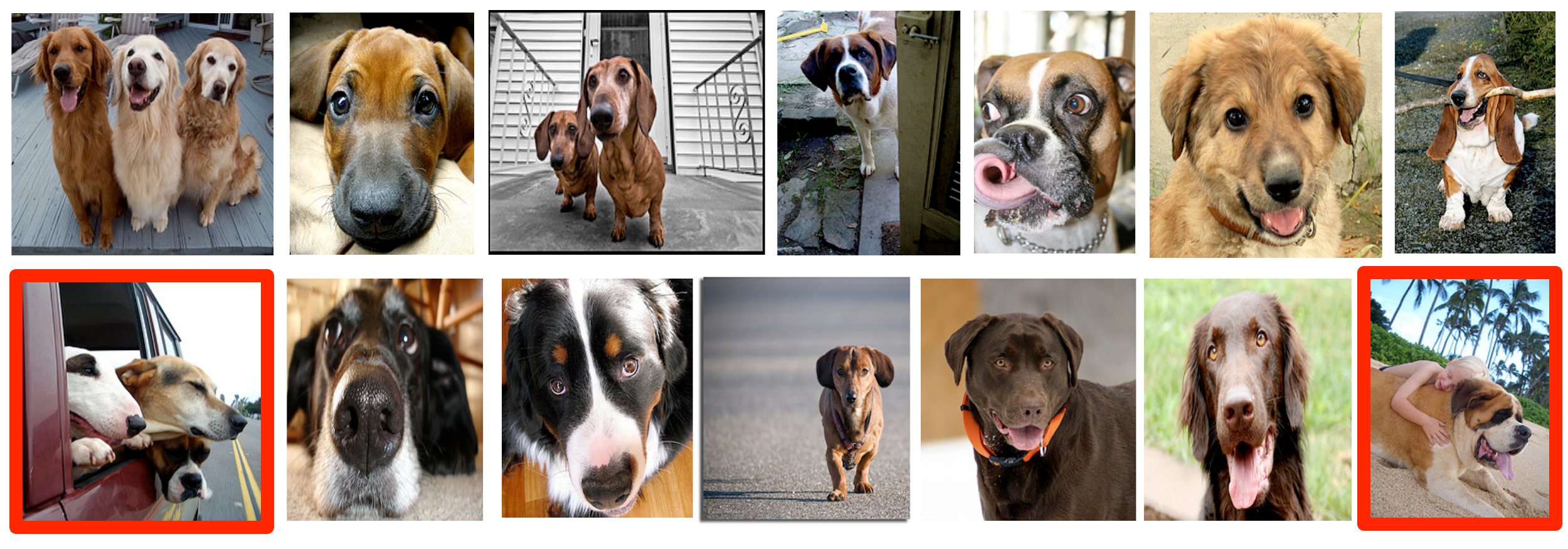}
%	\vspace{-0.25in}
	\caption{Dog}

\end{subfigure}
        
\caption{
   \label{fig:NUS_annot_issue}
NUS-WIDE annotation error examples. The top retrieved images in RLR for cat and dog categories are shown.  Red boxes are shown around images marked as negative samples in the dataset.}
%\vspace{-0.3in}
\end{figure}

Figure\ref{fig:NUS_annot_issue} shows some samples of annotation error in the NUS-WIDE dataset. Another example is car and vehicle categories: in the NUS-WIDE test set there are 431 instances of ``cars'' of which only 177 instances are also annotated as ``vehicle''.

%\section{Image retrieval and tag suggestion}
%\label{sec:retTagSugg}
%The last figure shows the image retrieval and tag suggestion for many tags. 

\addtocounter{subfigure}{-2}

\arxivcond{}{

\section{Qualitative results}

The following pages show examples of search results from F100M using our algorithm. 
For each tag, we show the top 25 retrieved images sorted from top-left to bottom-right. We also show the top 5 tag suggestions for the top 5 retrieved images of each query.   Note that many of the tags that we can query and predict are not included in any previous computer vision datasets. 
The query tags we show here are:
\tn{old+city},
\tn{cosplay},
\tn{person},
\tn{presentation},
\tn{painting},
\tn{map},
\tn{rusty},
\tn{box},
\tn{venice},
\tn{umbrella},
\tn{hdr},
\tn{vintage},
\tn{cat},
\tn{dog},
\tn{graduation},
\tn{flamingo},
\tn{tower+bridge},
\tn{yellowstone+national+park},
\tn{chain},
\tn{chair},
\tn{brooklyn+bridge},
\tn{golden+gate+bridge},
\tn{firework},
\tn{rainbow},
\tn{bicycle},
\tn{train},
\tn{bedroom},
\tn{bathroom},
\tn{swimming},
\tn{basketball},
\tn{blue},
\tn{green},
\tn{drum}, and
\tn{bass}.
}

\arxivcond{}{
\section{Algorithm Derivations}

See pages 19--22.
}

\newcommand{\mycaptionsize}{\large\bf}

\arxivcond{}{

\begin{figure*}
        \centering
	\vspace{-.3in}
        \newcommand{\tagname}{showjumping}
        \begin{subfigure}[b]{\textwidth}
		\centering
                \includegraphics[width=.9\textwidth]{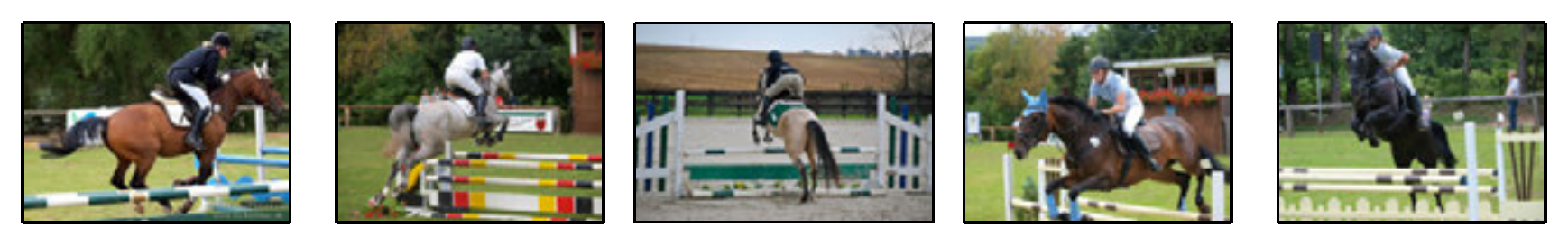}
        \end{subfigure} \\
	\vspace{-.1in}
        \begin{subfigure}[b]{\textwidth}
		\centering
                \includegraphics[width=.9\textwidth]{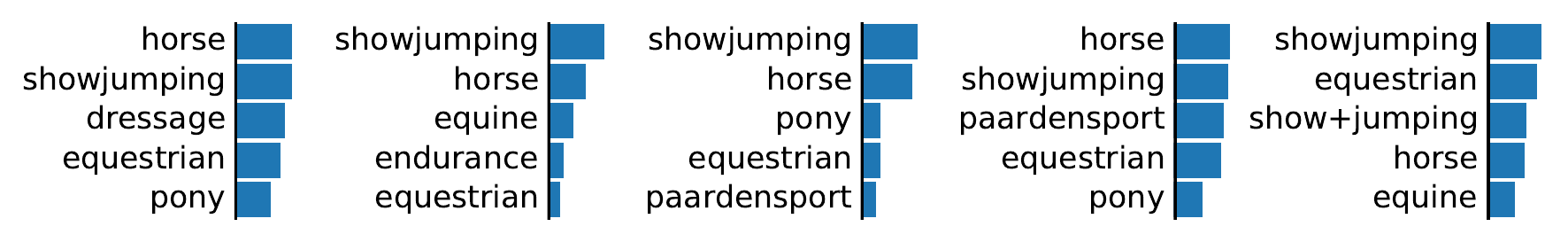}
        \end{subfigure} \\
	\vspace{-.1in}
        \begin{subfigure}[b]{\textwidth}
		\centering
                \includegraphics[width=1\textwidth]{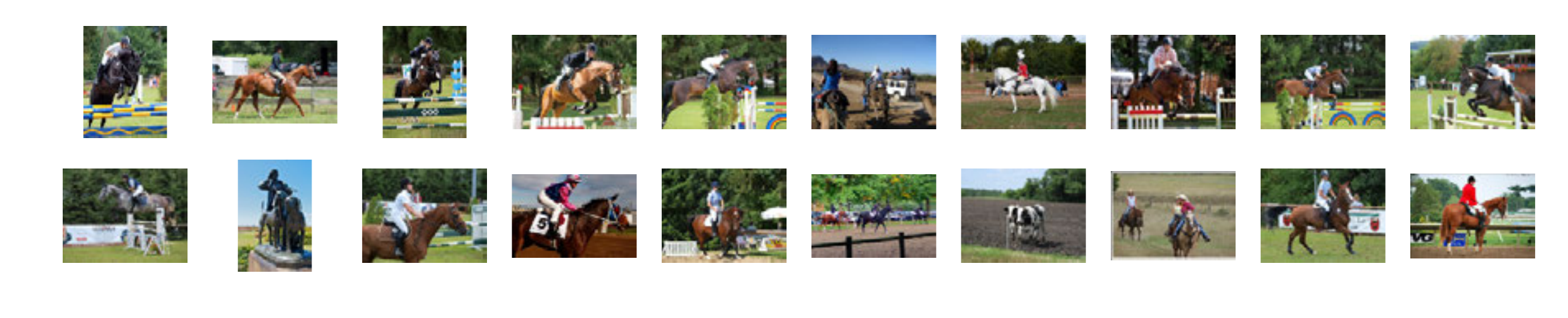}
		\vspace{-.4in}
                \caption{\mycaptionsize \tagname}
		\vspace{-.1in}
        \end{subfigure}
        \rule{500px}{1px}
	\vspace{-.1in}

        \renewcommand{\tagname}{old+city}
        \begin{subfigure}[b]{\textwidth}
		\centering
                \includegraphics[width=.9\textwidth]{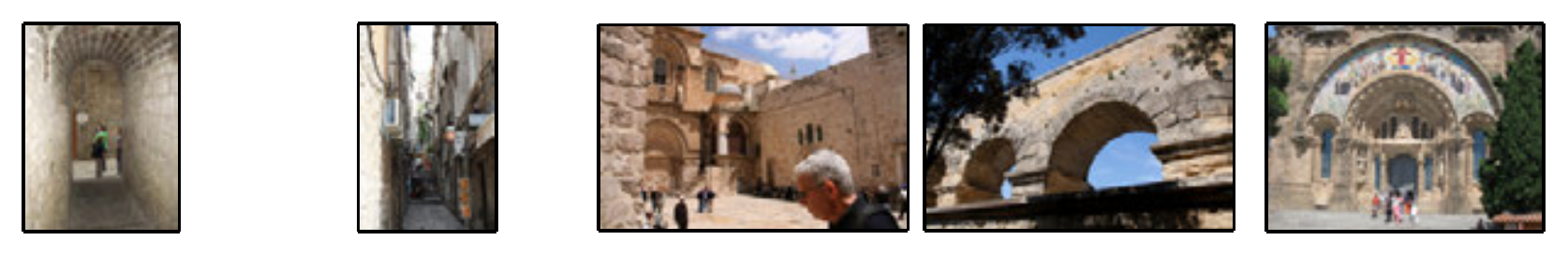}
        \end{subfigure} \\
	\vspace{-.1in}
        \begin{subfigure}[b]{\textwidth}
		\centering
                \includegraphics[width=.9\textwidth]{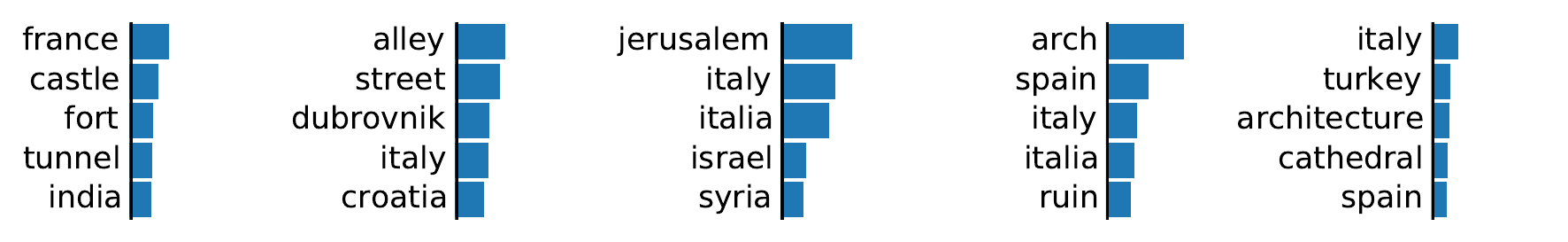}
        \end{subfigure} \\
	\vspace{-.1in}
        \begin{subfigure}[b]{\textwidth}
		\centering
                \includegraphics[width=1\textwidth]{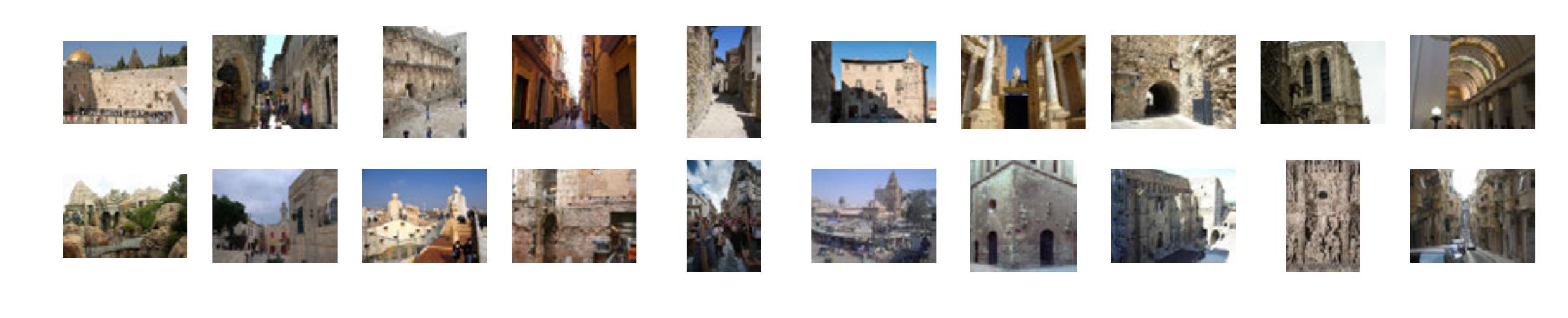}
		\vspace{-.4in}
                \caption{\mycaptionsize\tagname}
		\vspace{-.1in}
        \end{subfigure}
        \rule{500px}{1px}
	\vspace{-.1in}

        \renewcommand{\tagname}{cosplay}
        \begin{subfigure}[b]{\textwidth}
		\centering
                \includegraphics[width=.9\textwidth]{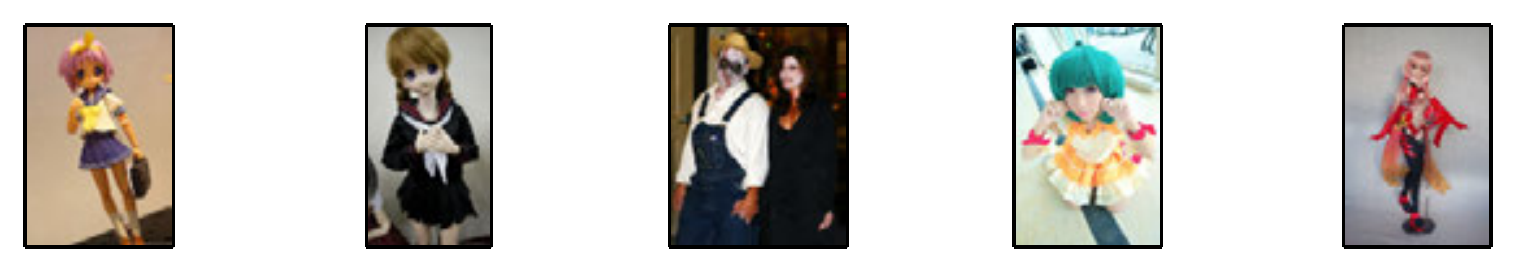}
        \end{subfigure} \\
	\vspace{-.1in}
        \begin{subfigure}[b]{\textwidth}
		\centering
                \includegraphics[width=.9\textwidth]{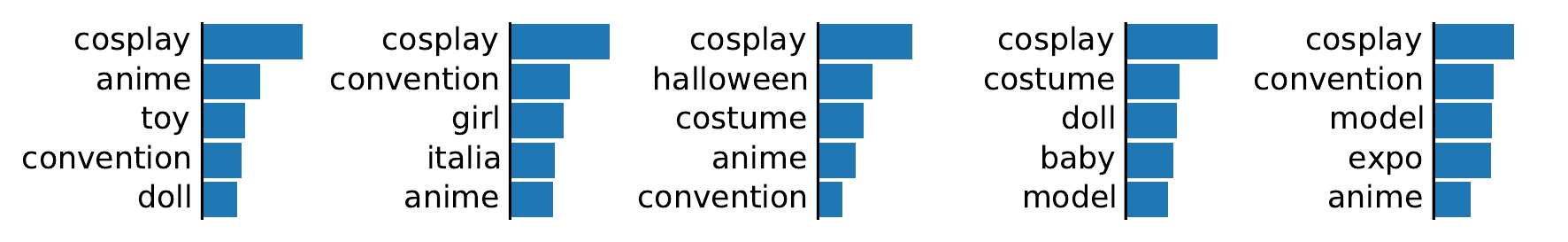}
        \end{subfigure} \\
	\vspace{-.1in}
        \begin{subfigure}[b]{\textwidth}
		\centering
                \includegraphics[width=1\textwidth]{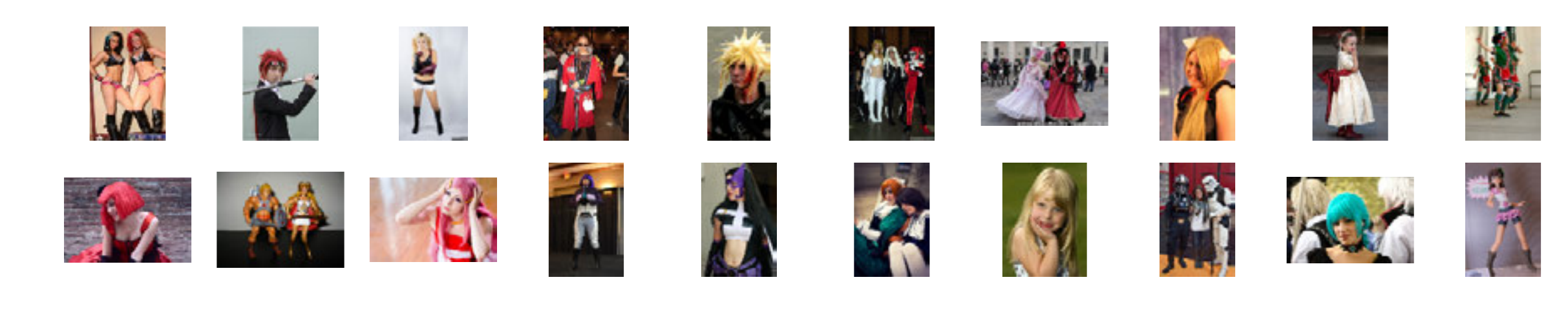}
		\vspace{-.4in}
                \caption{\mycaptionsize\tagname}
		\vspace{-.1in}
        \end{subfigure}
\end{figure*}

\begin{figure*}
        \centering
	\vspace{-.3in}
        \newcommand{\tagname}{newborn}
        \begin{subfigure}[b]{\textwidth}
		\centering
                \includegraphics[width=.9\textwidth]{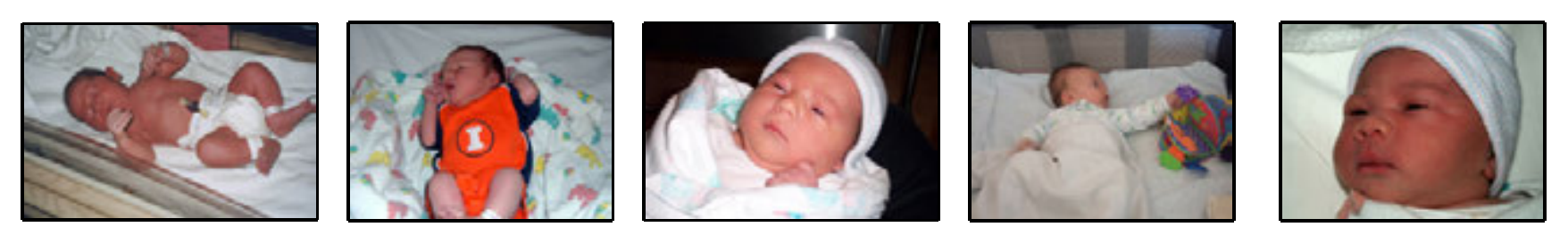}
        \end{subfigure} \\
	\vspace{-.1in}
        \begin{subfigure}[b]{\textwidth}
		\centering
                \includegraphics[width=.9\textwidth]{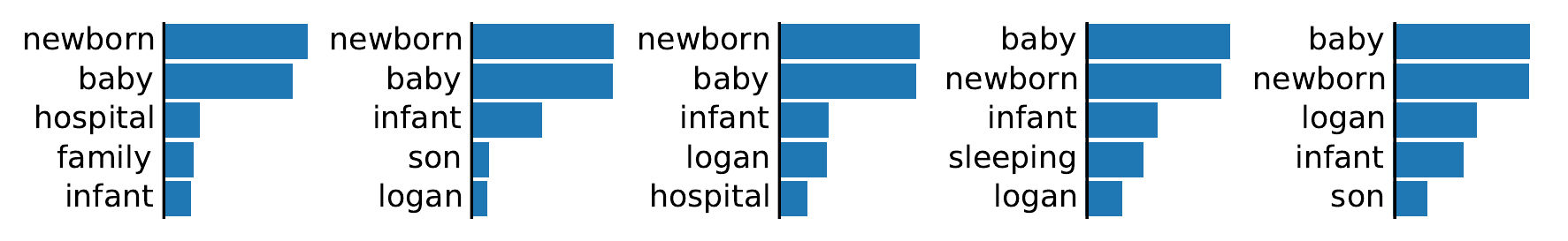}
        \end{subfigure} \\
	\vspace{-.1in}
        \begin{subfigure}[b]{\textwidth}
		\centering
                \includegraphics[width=1\textwidth]{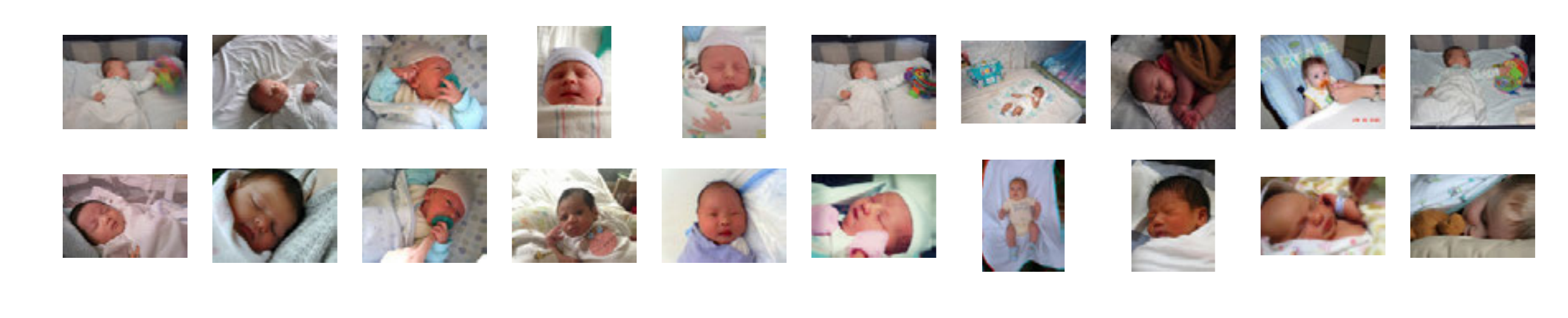}
		\vspace{-.4in}
                \caption{\mycaptionsize\tagname}
		\vspace{-.1in}
        \end{subfigure}
        \rule{500px}{1px}
	\vspace{-.1in}

        \renewcommand{\tagname}{person}
        \begin{subfigure}[b]{\textwidth}
		\centering
                \includegraphics[width=.9\textwidth]{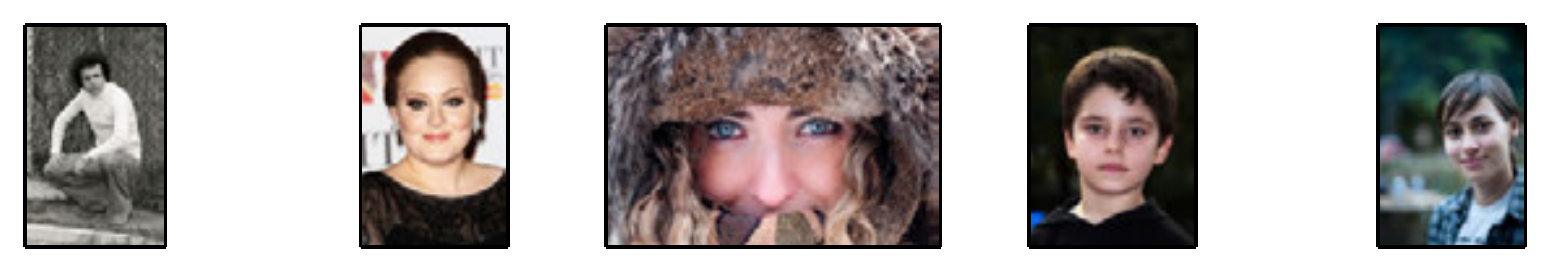}
        \end{subfigure} \\
	\vspace{-.1in}
        \begin{subfigure}[b]{\textwidth}
		\centering
                \includegraphics[width=.9\textwidth]{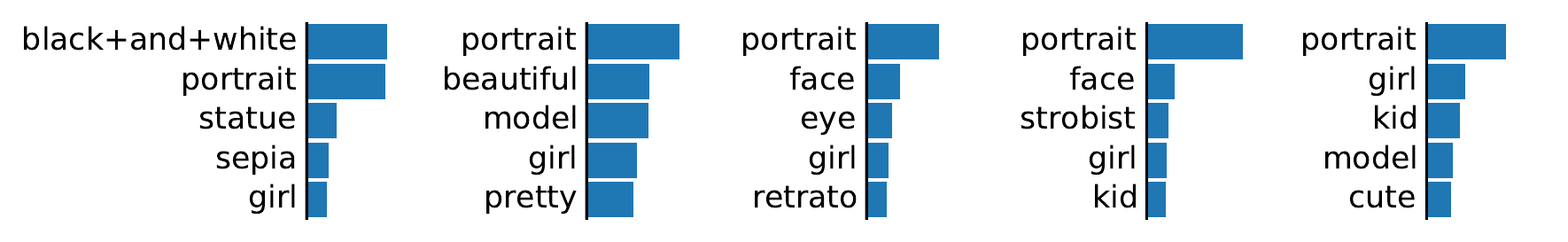}
        \end{subfigure} \\
	\vspace{-.1in}
        \begin{subfigure}[b]{\textwidth}
		\centering
                \includegraphics[width=1\textwidth]{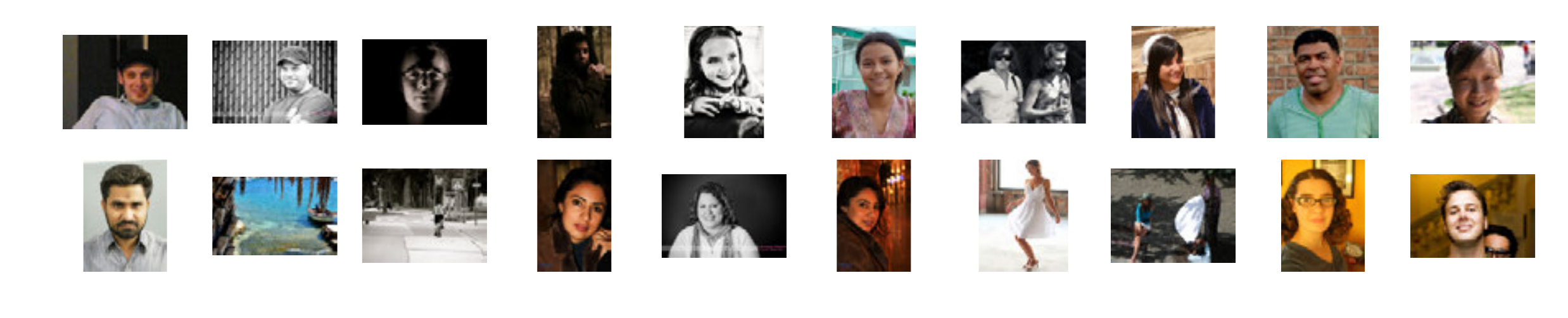}
		\vspace{-.4in}
                \caption{\mycaptionsize\tagname}
		\vspace{-.1in}
        \end{subfigure}
        \rule{500px}{1px}
	\vspace{-.1in}

        \renewcommand{\tagname}{presentation}
        \begin{subfigure}[b]{\textwidth}
		\centering
                \includegraphics[width=.9\textwidth]{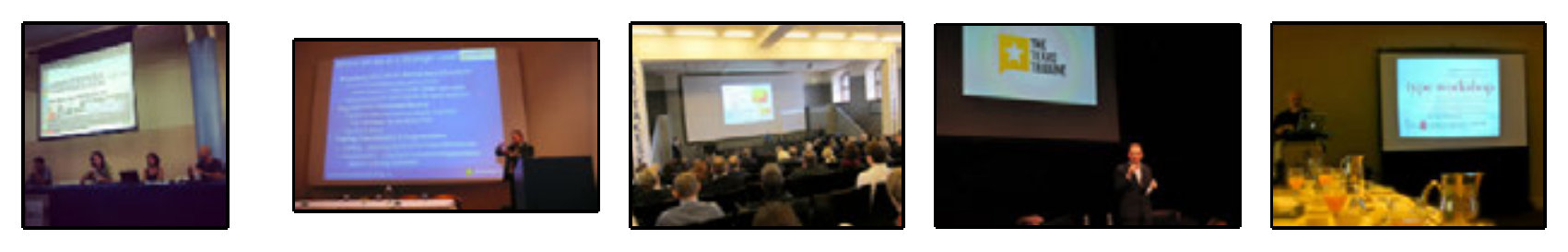}
        \end{subfigure} \\
	\vspace{-.1in}
        \begin{subfigure}[b]{\textwidth}
		\centering
                \includegraphics[width=.9\textwidth]{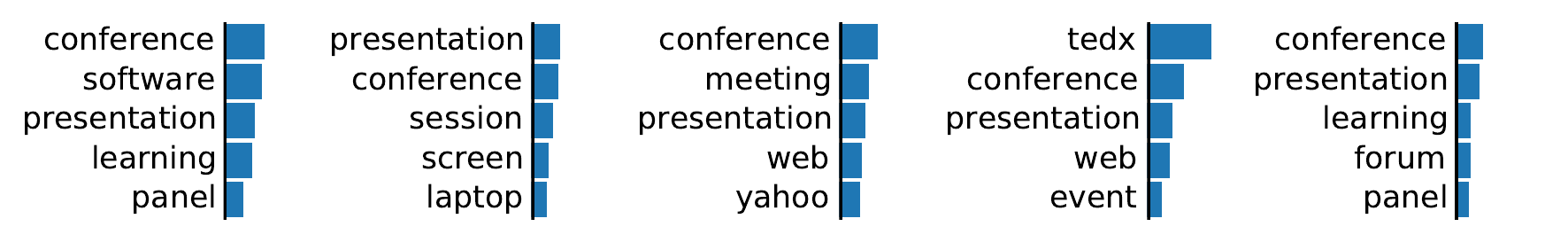}
        \end{subfigure} \\
	\vspace{-.01in}
        \begin{subfigure}[b]{\textwidth}
		\centering
                \includegraphics[width=1\textwidth]{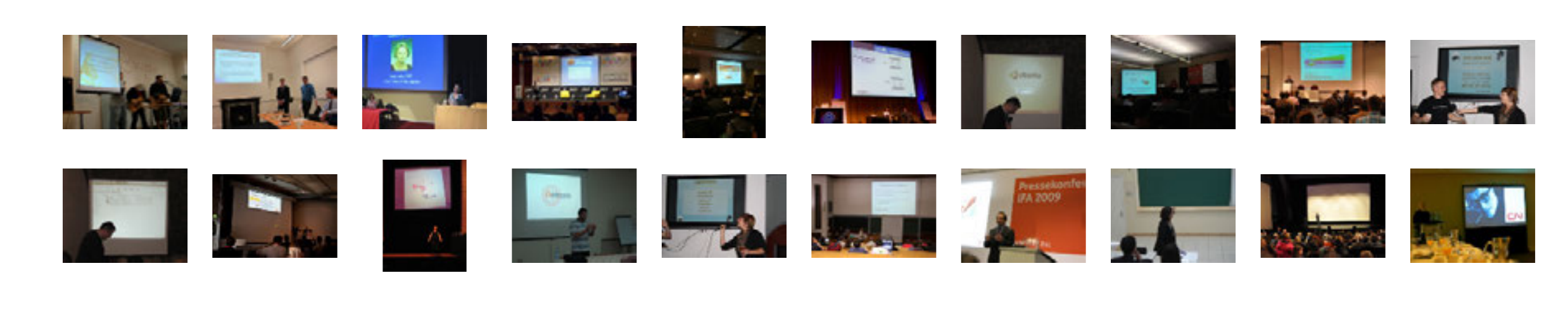}
		\vspace{-.4in}
                \caption{\mycaptionsize\tagname}
		\vspace{-.1in}
        \end{subfigure}
\end{figure*}

\begin{figure*}
        \centering
	\vspace{-.3in}
        \newcommand{\tagname}{drawing}
        \begin{subfigure}[b]{\textwidth}
		\centering
                \includegraphics[width=.9\textwidth]{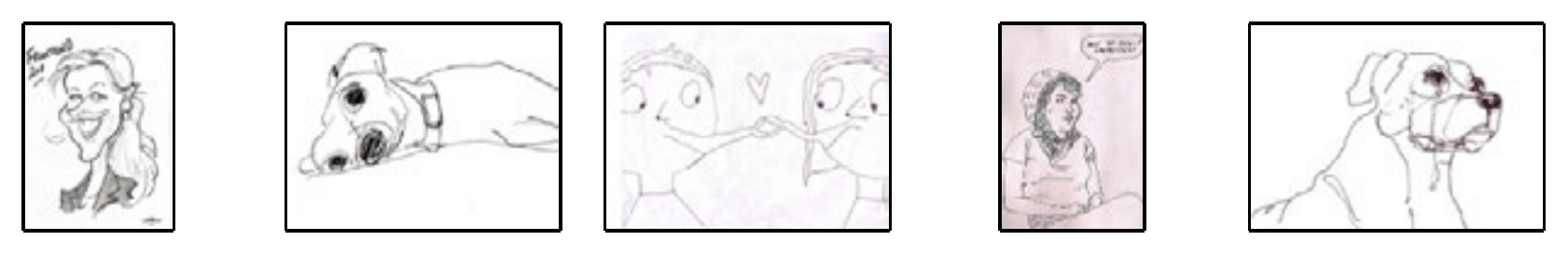}
        \end{subfigure} \\
	\vspace{-.1in}
        \begin{subfigure}[b]{\textwidth}
		\centering
                \includegraphics[width=.9\textwidth]{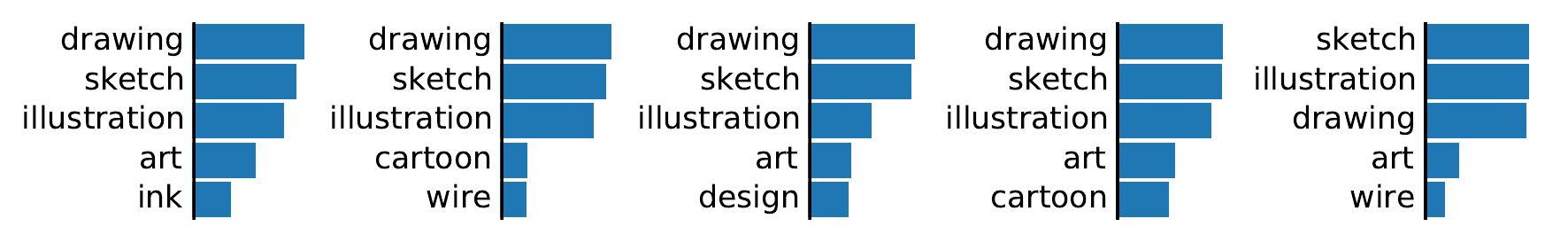}
        \end{subfigure} \\
	\vspace{-.1in}
        \begin{subfigure}[b]{\textwidth}
		\centering
                \includegraphics[width=1\textwidth]{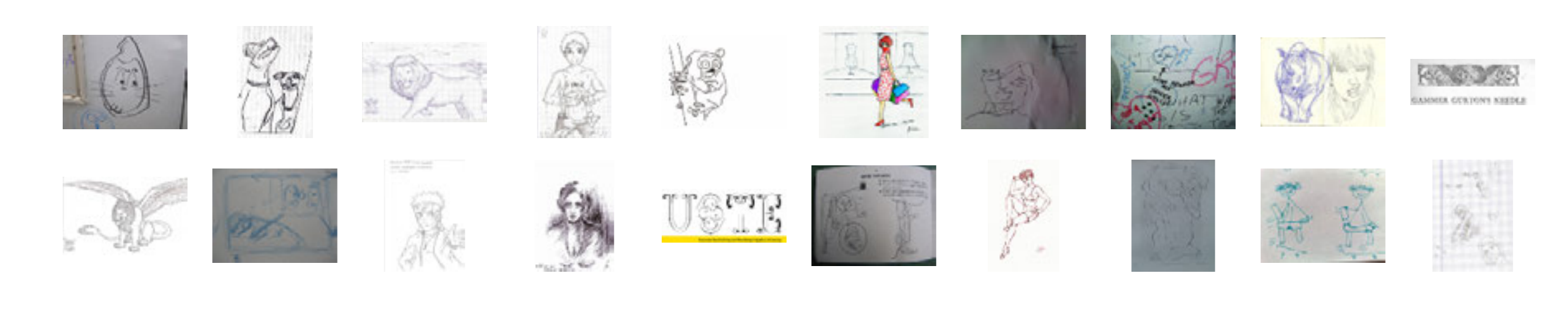}
		\vspace{-.4in}
                \caption{\mycaptionsize\tagname}
		\vspace{-.1in}
        \end{subfigure}
        \rule{500px}{1px}
	\vspace{-.1in}

        \renewcommand{\tagname}{painting}
        \begin{subfigure}[b]{\textwidth}
		\centering
                \includegraphics[width=.9\textwidth]{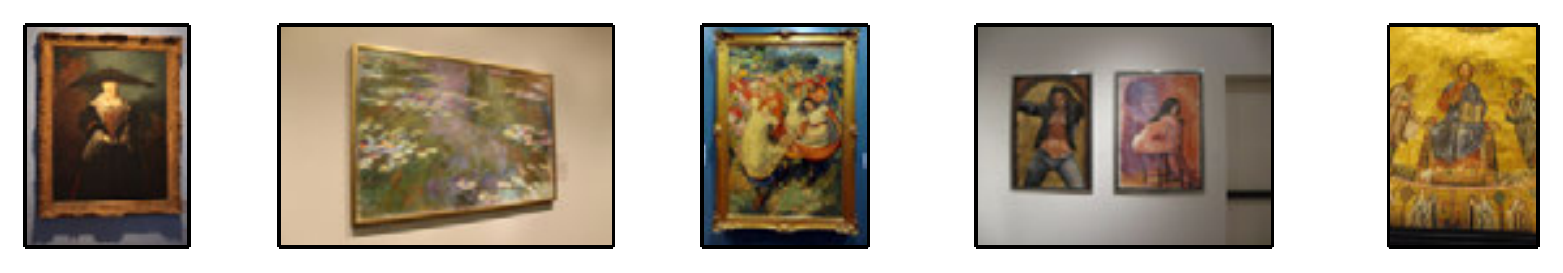}
        \end{subfigure} \\
	\vspace{-.1in}
        \begin{subfigure}[b]{\textwidth}
		\centering
                \includegraphics[width=.9\textwidth]{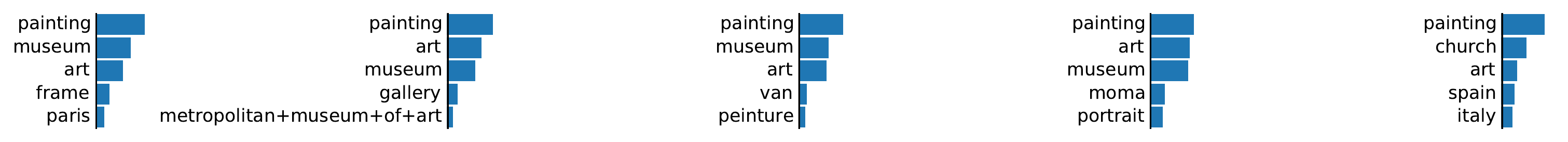}
        \end{subfigure} \\
	\vspace{-.01in}
        \begin{subfigure}[b]{\textwidth}
		\centering
                \includegraphics[width=1\textwidth]{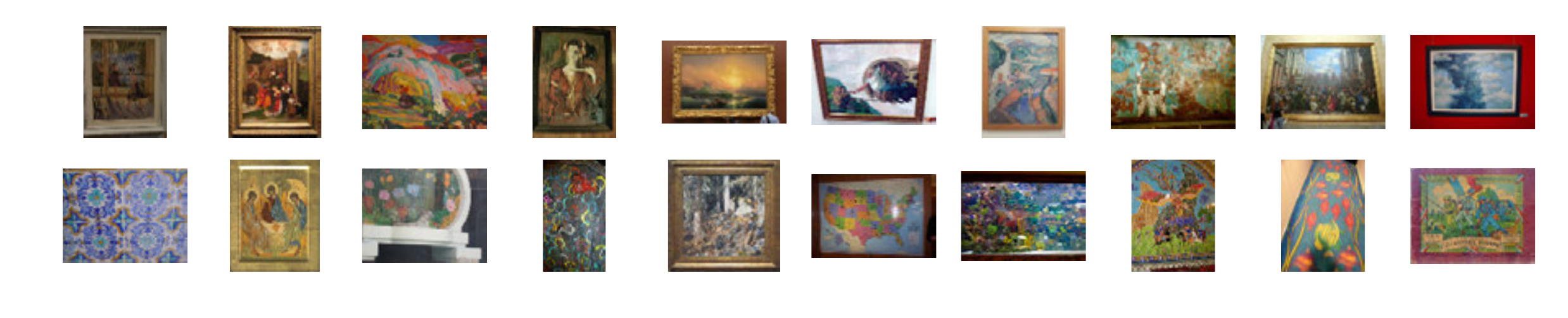}
		\vspace{-.4in}
                \caption{\mycaptionsize\tagname}
		\vspace{-.1in}
        \end{subfigure}
        \rule{500px}{1px}
	\vspace{-.1in}

        \renewcommand{\tagname}{map}
        \begin{subfigure}[b]{\textwidth}
		\centering
                \includegraphics[width=.9\textwidth]{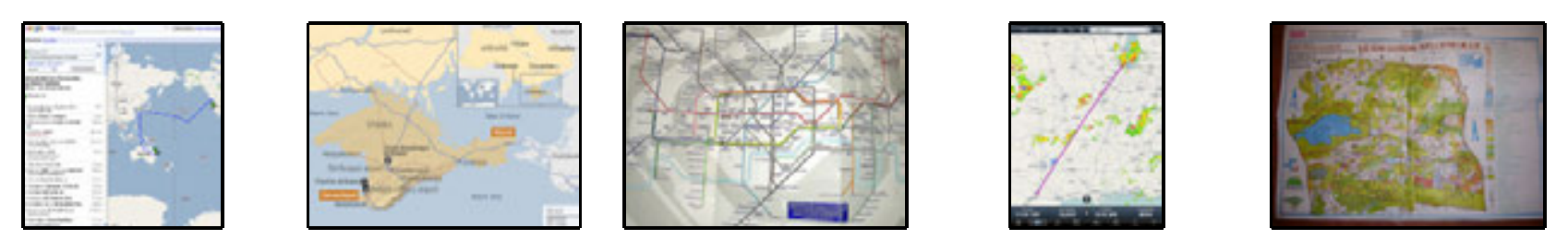}
        \end{subfigure} \\
	\vspace{-.1in}
        \begin{subfigure}[b]{\textwidth}
		\centering
                \includegraphics[width=.9\textwidth]{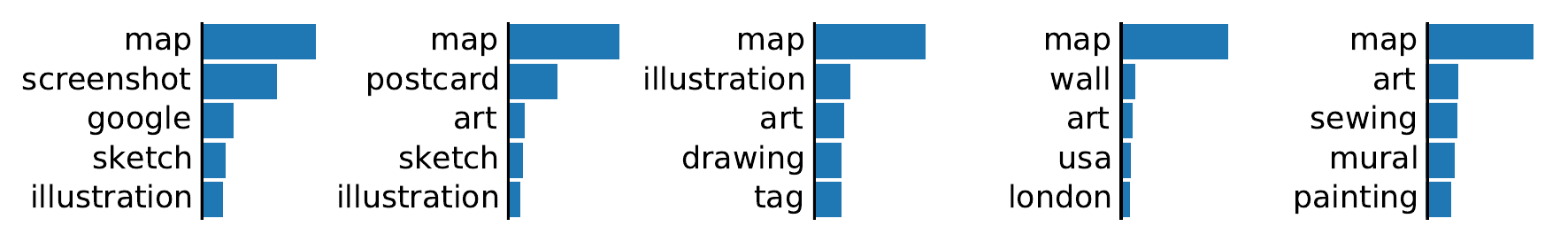}
        \end{subfigure} \\
	\vspace{-.01in}
        \begin{subfigure}[b]{\textwidth}
		\centering
                \includegraphics[width=1\textwidth]{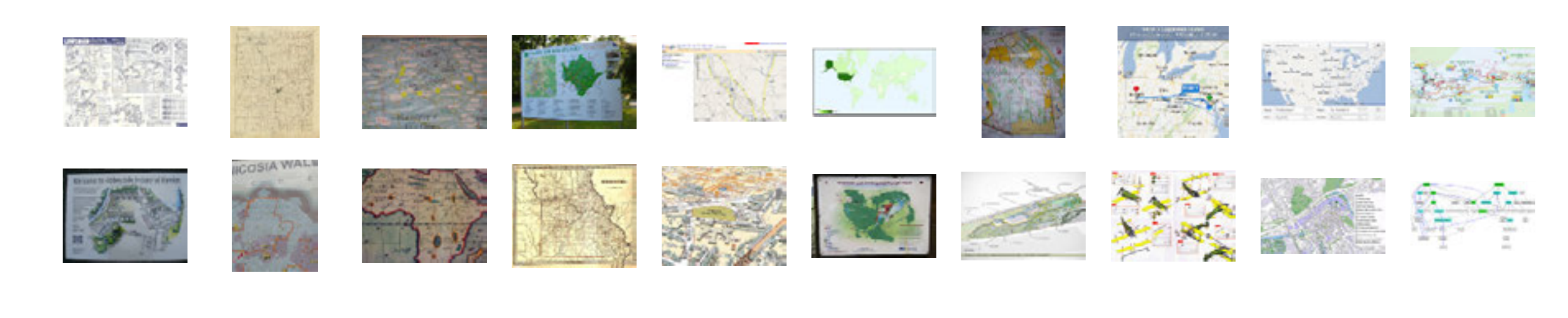}
		\vspace{-.4in}
                \caption{\mycaptionsize\tagname}
		\vspace{-.1in}
        \end{subfigure}
\end{figure*}

\begin{figure*}
        \centering
	\vspace{-.3in}
        \newcommand{\tagname}{furry}
        \begin{subfigure}[b]{\textwidth}
		\centering
                \includegraphics[width=.9\textwidth]{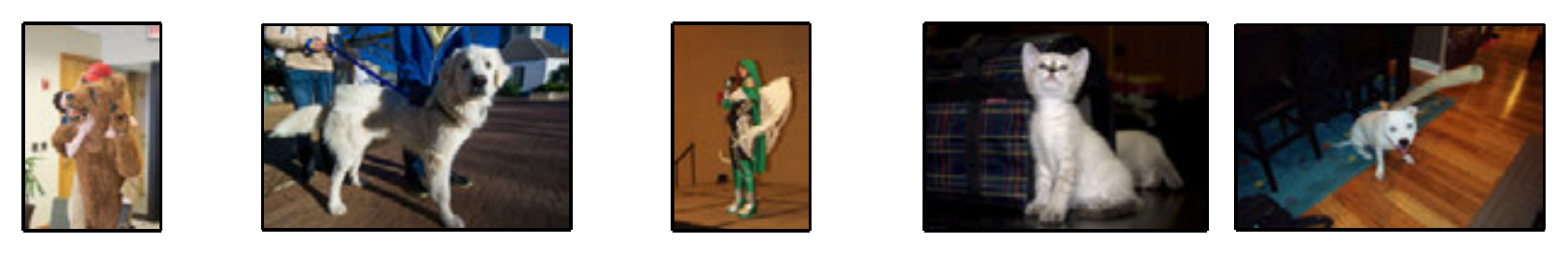}
        \end{subfigure} \\
	\vspace{-.1in}
        \begin{subfigure}[b]{\textwidth}
		\centering
                \includegraphics[width=.9\textwidth]{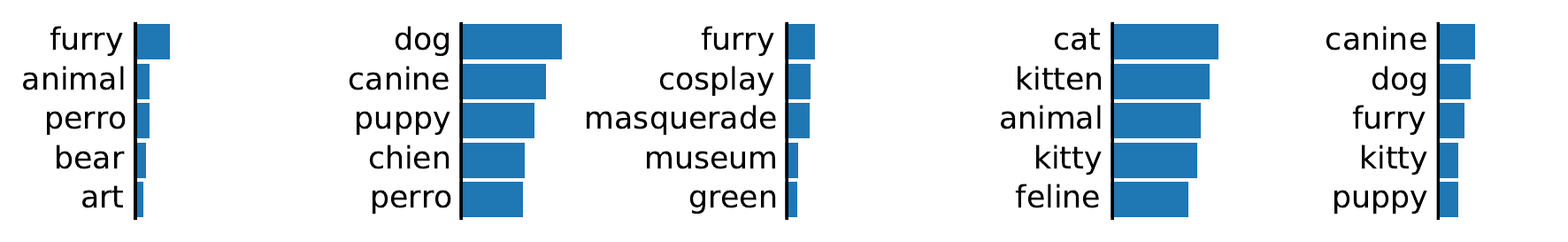}
        \end{subfigure} \\
	\vspace{-.1in}
        \begin{subfigure}[b]{\textwidth}
		\centering
                \includegraphics[width=1\textwidth]{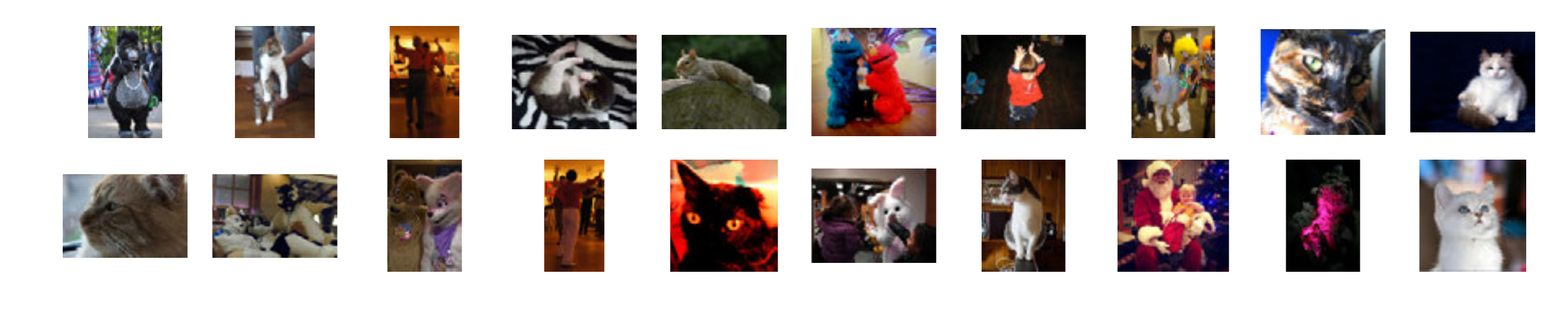}
		\vspace{-.4in}
                \caption{\mycaptionsize\tagname}
		\vspace{-.1in}
        \end{subfigure}
        \rule{500px}{1px}
	\vspace{-.1in}

        \renewcommand{\tagname}{rusty}
        \begin{subfigure}[b]{\textwidth}
		\centering
                \includegraphics[width=.9\textwidth]{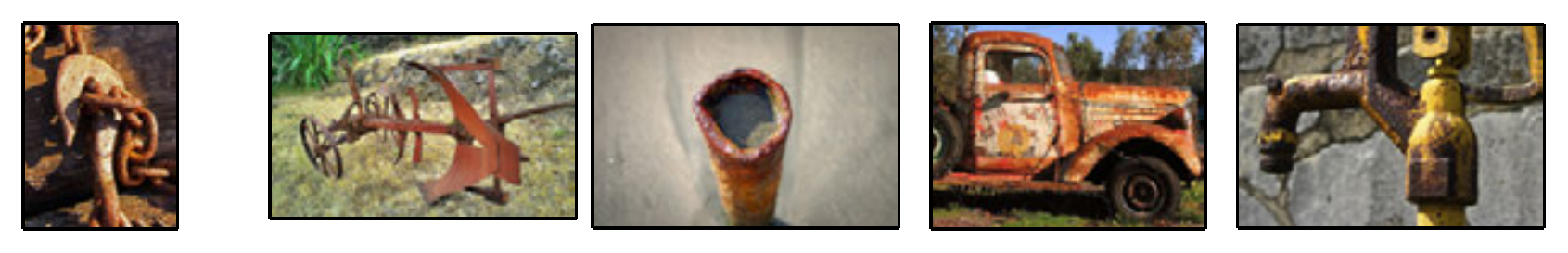}
        \end{subfigure} \\
	\vspace{-.1in}
        \begin{subfigure}[b]{\textwidth}
		\centering
                \includegraphics[width=.9\textwidth]{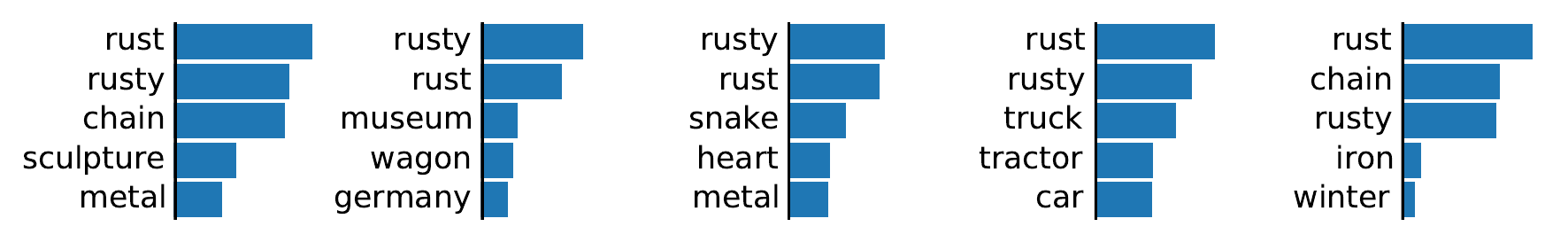}
        \end{subfigure} \\
	\vspace{-.1in}
        \begin{subfigure}[b]{\textwidth}
		\centering
                \includegraphics[width=1\textwidth]{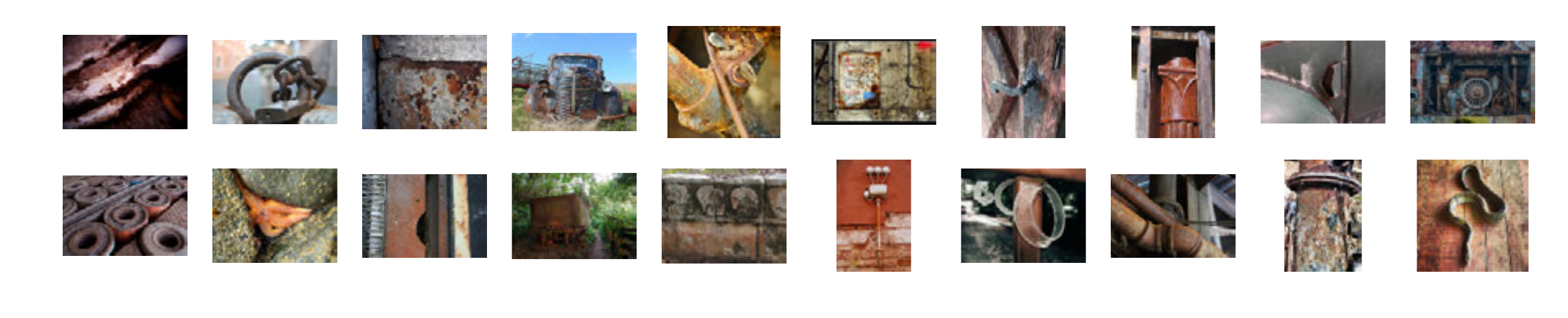}
		\vspace{-.4in}
                \caption{\mycaptionsize\tagname}
		\vspace{-.1in}
        \end{subfigure}
        \rule{500px}{1px}
	\vspace{-.1in}

        \renewcommand{\tagname}{box}
        \begin{subfigure}[b]{\textwidth}
		\centering
                \includegraphics[width=.9\textwidth]{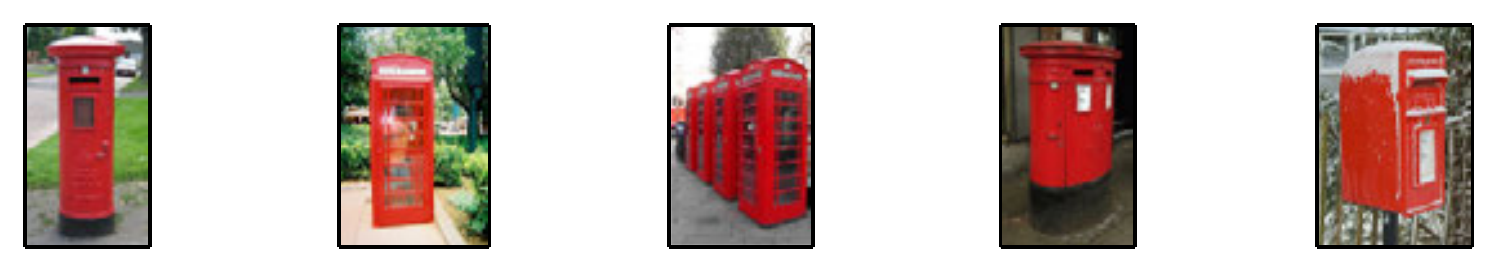}
        \end{subfigure} \\
	\vspace{-.1in}
        \begin{subfigure}[b]{\textwidth}
		\centering
                \includegraphics[width=.9\textwidth]{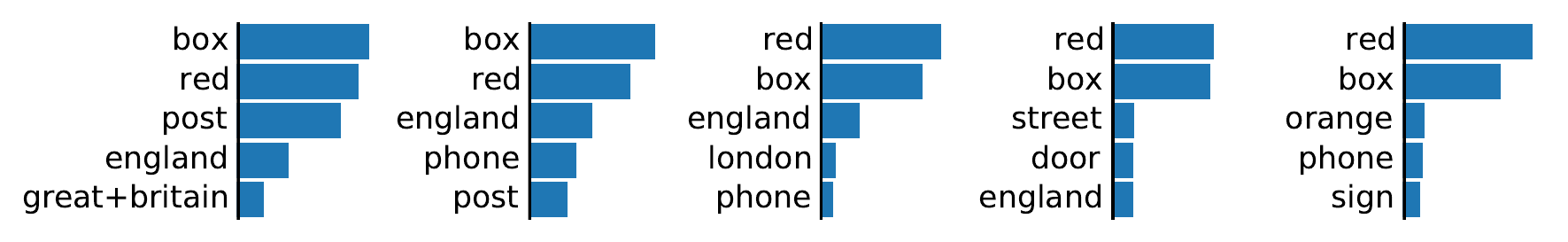}
        \end{subfigure} \\
	\vspace{-.01in}
        \begin{subfigure}[b]{\textwidth}
		\centering
                \includegraphics[width=1\textwidth]{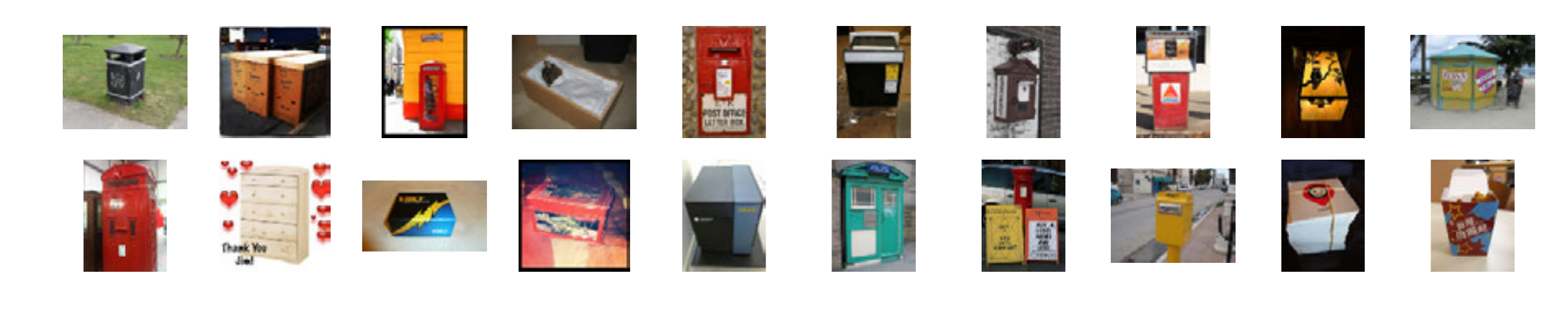}
		\vspace{-.4in}
                \caption{\mycaptionsize\tagname}
		\vspace{-.1in}
        \end{subfigure}
\end{figure*}

\begin{figure*}
        \centering
	\vspace{-.3in}
        \newcommand{\tagname}{wedding}
        \begin{subfigure}[b]{\textwidth}
		\centering
                \includegraphics[width=.9\textwidth]{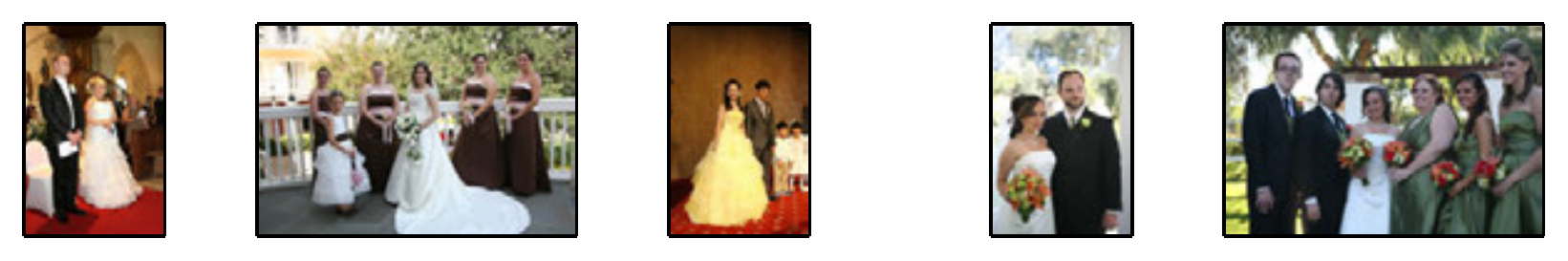}
        \end{subfigure} \\
	\vspace{-.1in}
        \begin{subfigure}[b]{\textwidth}
		\centering
                \includegraphics[width=.9\textwidth]{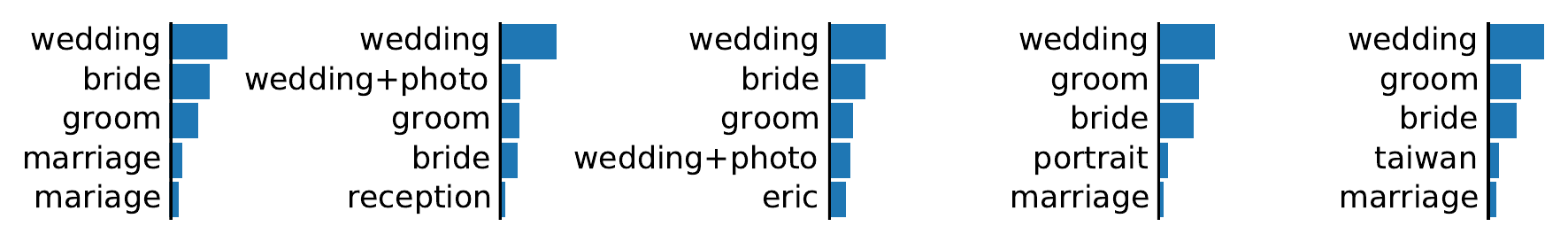}
        \end{subfigure} \\
	\vspace{-.1in}
        \begin{subfigure}[b]{\textwidth}
		\centering
                \includegraphics[width=1\textwidth]{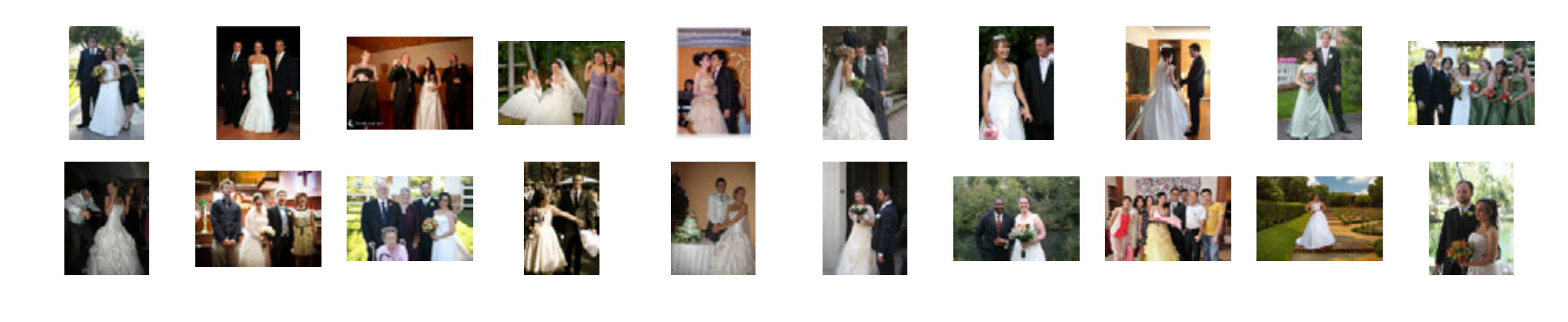}
		\vspace{-.4in}
                \caption{\mycaptionsize\tagname}
		\vspace{-.1in}
        \end{subfigure}
        \rule{500px}{1px}
	\vspace{-.1in}

        \renewcommand{\tagname}{venice}
        \begin{subfigure}[b]{\textwidth}
		\centering
                \includegraphics[width=.9\textwidth]{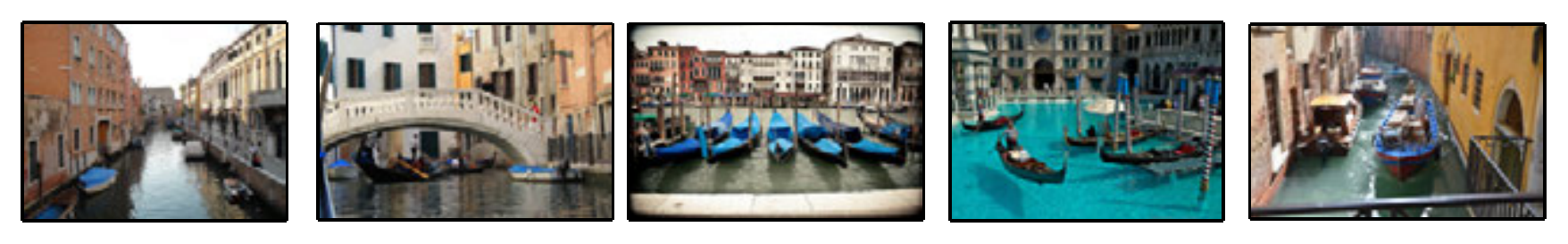}
        \end{subfigure} \\
	\vspace{-.1in}
        \begin{subfigure}[b]{\textwidth}
		\centering
                \includegraphics[width=.9\textwidth]{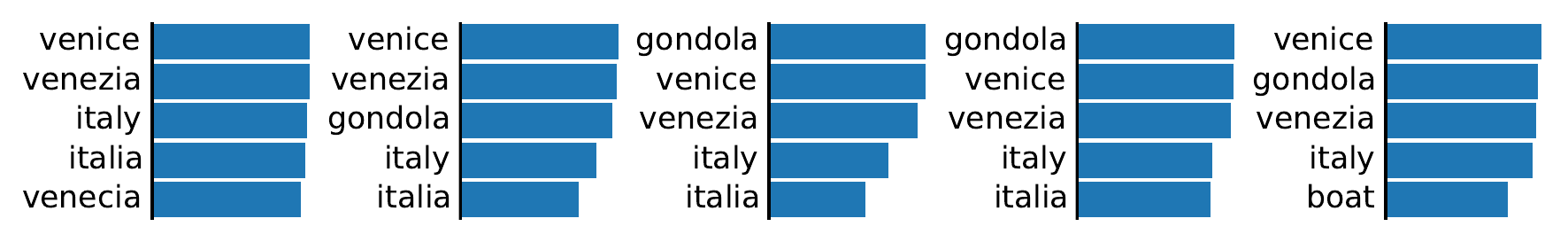}
        \end{subfigure} \\
	\vspace{-.1in}
        \begin{subfigure}[b]{\textwidth}
		\centering
                \includegraphics[width=1\textwidth]{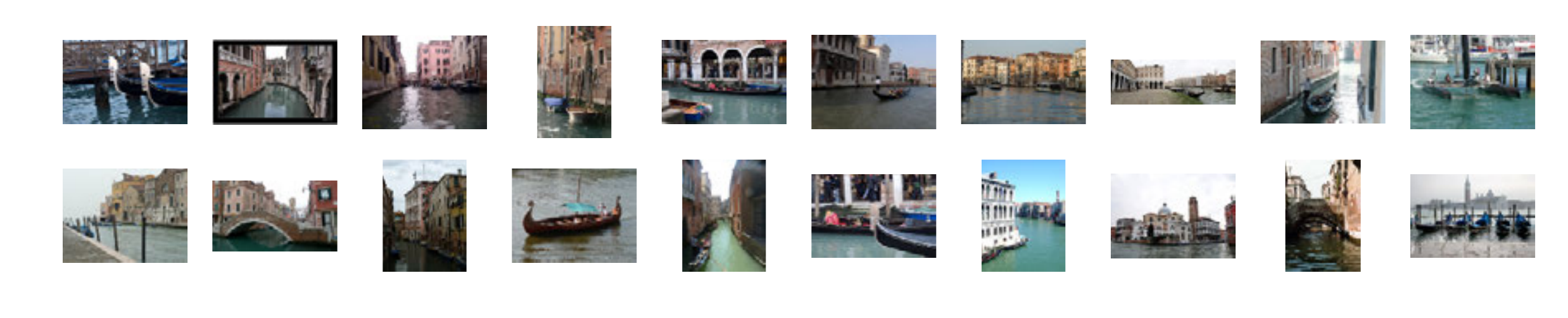}
		\vspace{-.4in}
                \caption{\mycaptionsize\tagname}
		\vspace{-.1in}
        \end{subfigure}
        \rule{500px}{1px}
	\vspace{-.1in}

        \renewcommand{\tagname}{umbrella}
        \begin{subfigure}[b]{\textwidth}
		\centering
                \includegraphics[width=.9\textwidth]{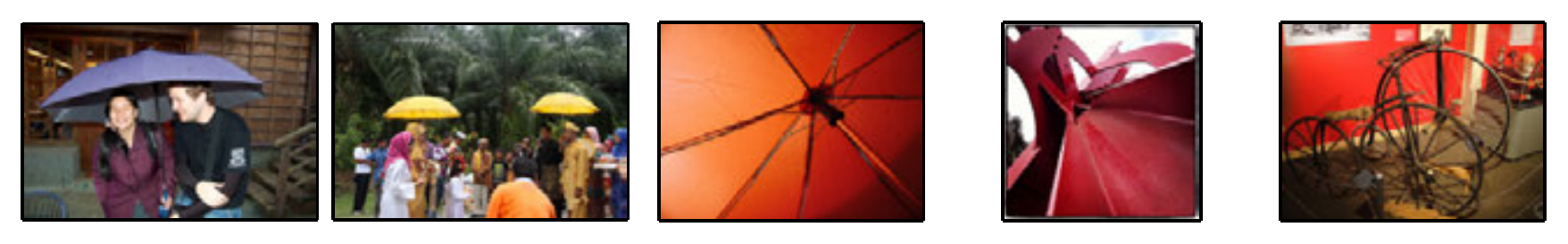}
        \end{subfigure} \\
	\vspace{-.1in}
        \begin{subfigure}[b]{\textwidth}
		\centering
                \includegraphics[width=.9\textwidth]{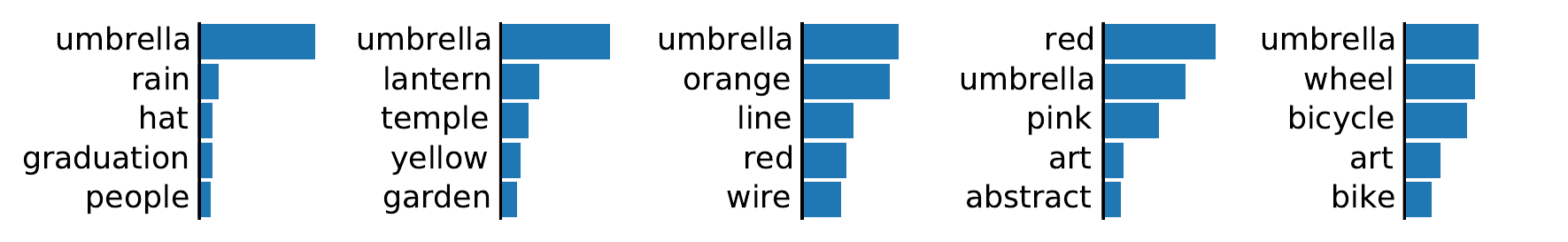}
        \end{subfigure} \\
	\vspace{-.1in}
        \begin{subfigure}[b]{\textwidth}
		\centering
                \includegraphics[width=1\textwidth]{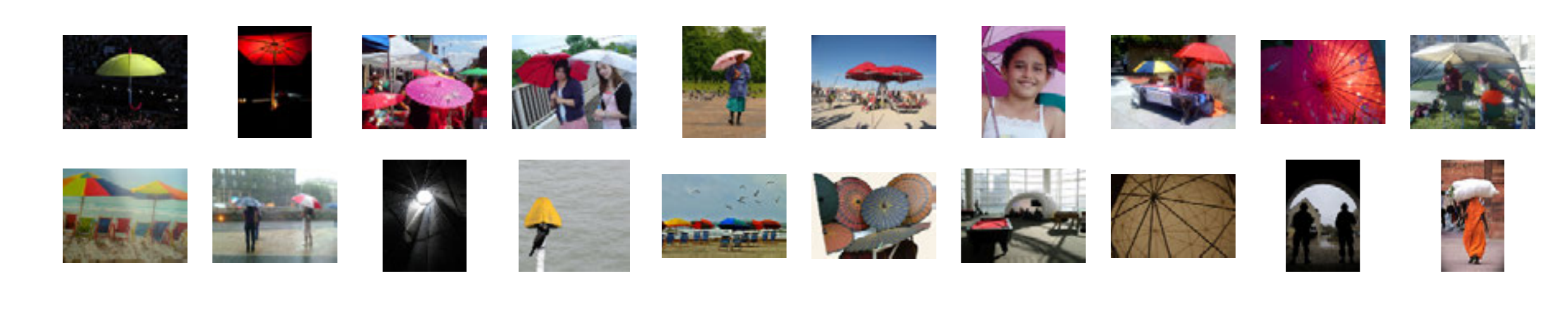}
		\vspace{-.4in}
                \caption{\mycaptionsize\tagname}
		\vspace{-.1in}
        \end{subfigure}
\end{figure*}

\begin{figure*}
        \centering
	\vspace{-.3in}
        \newcommand{\tagname}{macro}
        \begin{subfigure}[b]{\textwidth}
		\centering
                \includegraphics[width=.9\textwidth]{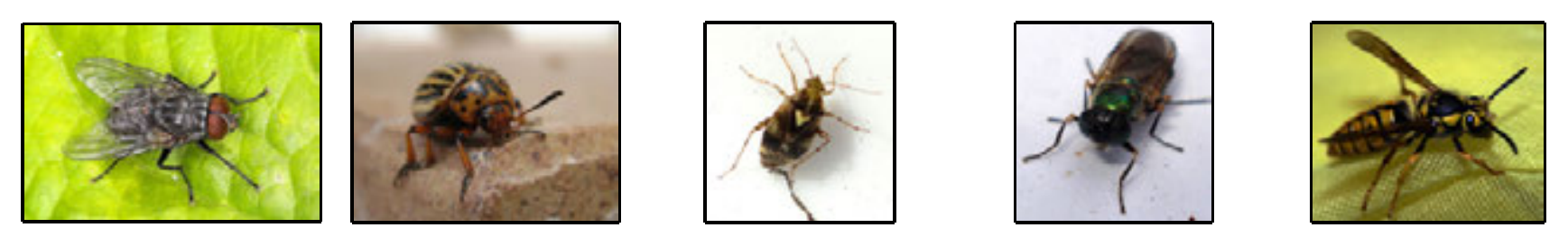}
        \end{subfigure} \\
	\vspace{-.1in}
        \begin{subfigure}[b]{\textwidth}
		\centering
                \includegraphics[width=.9\textwidth]{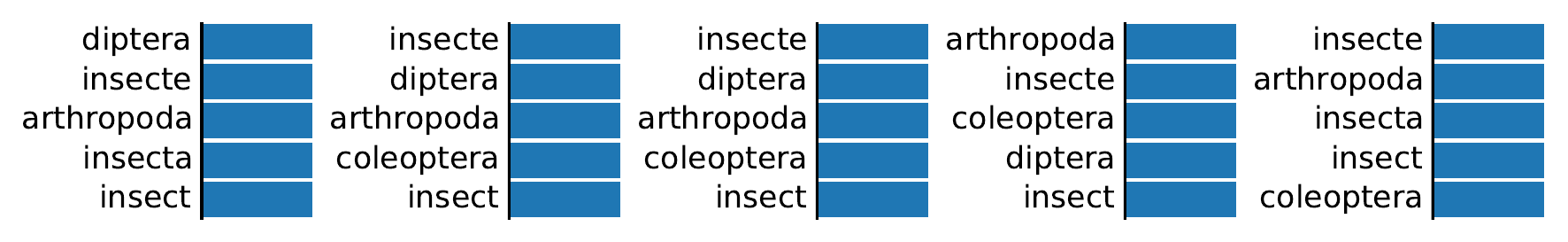}
        \end{subfigure} \\
	\vspace{-.1in}
        \begin{subfigure}[b]{\textwidth}
		\centering
                \includegraphics[width=1\textwidth]{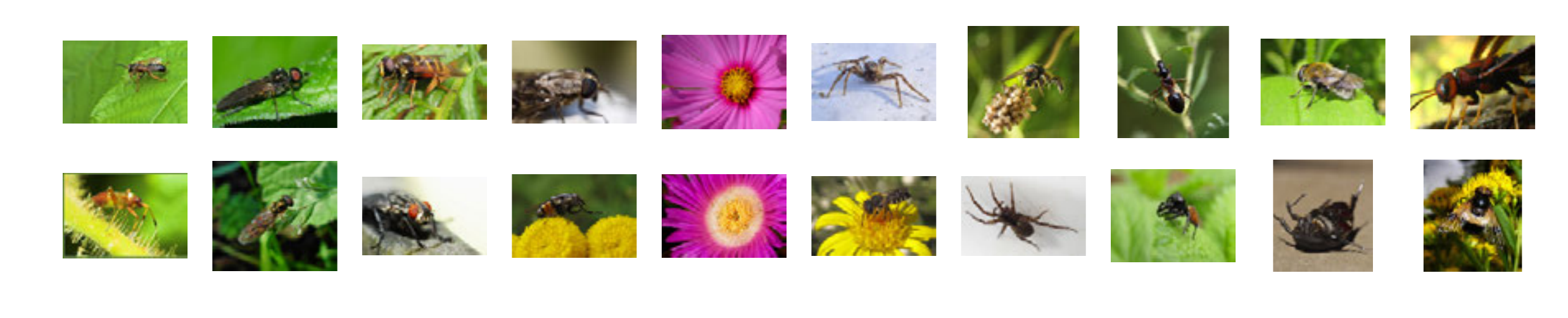}
		\vspace{-.4in}
                \caption{\mycaptionsize\tagname}
		\vspace{-.1in}
        \end{subfigure}
        \rule{500px}{1px}
	\vspace{-.1in}

        \renewcommand{\tagname}{hdr}
        \begin{subfigure}[b]{\textwidth}
		\centering
                \includegraphics[width=.9\textwidth]{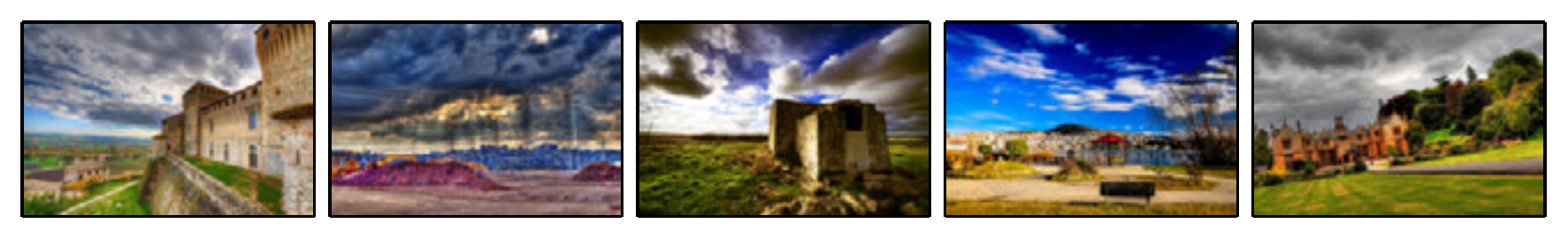}
        \end{subfigure} \\
	\vspace{-.1in}
        \begin{subfigure}[b]{\textwidth}
		\centering
                \includegraphics[width=.9\textwidth]{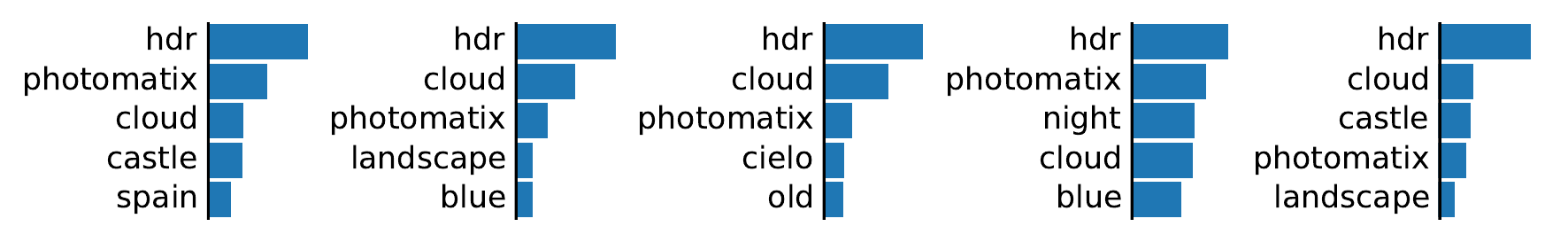}
        \end{subfigure} \\
	\vspace{-.1in}
        \begin{subfigure}[b]{\textwidth}
		\centering
                \includegraphics[width=1\textwidth]{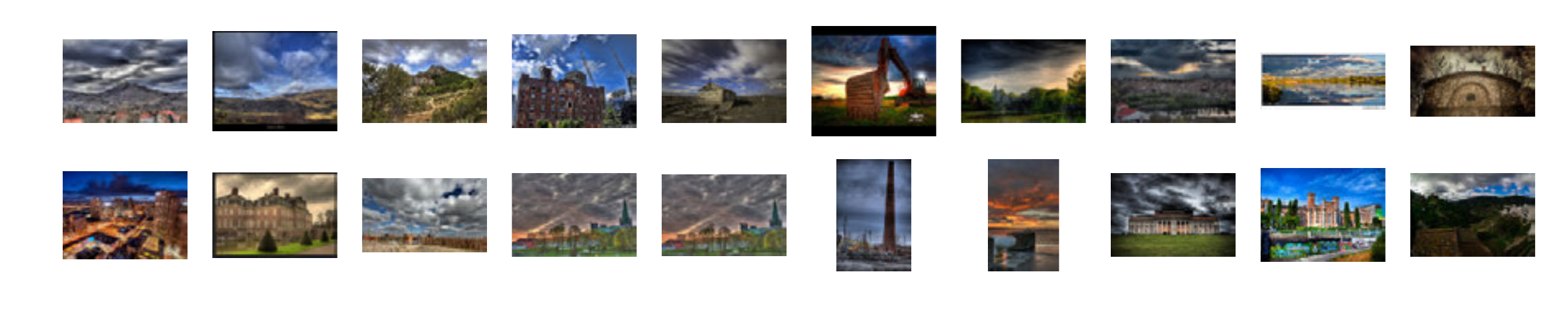}
		\vspace{-.4in}
                \caption{\mycaptionsize\tagname}
		\vspace{-.1in}
        \end{subfigure}
        \rule{500px}{1px}
	\vspace{-.1in}

        \renewcommand{\tagname}{vintage}
        \begin{subfigure}[b]{\textwidth}
		\centering
                \includegraphics[width=.9\textwidth]{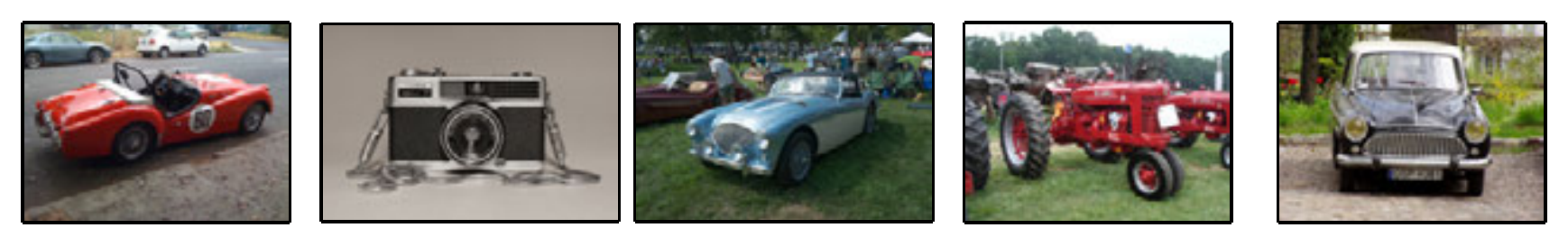}
        \end{subfigure} \\
	\vspace{-.1in}
        \begin{subfigure}[b]{\textwidth}
		\centering
                \includegraphics[width=.9\textwidth]{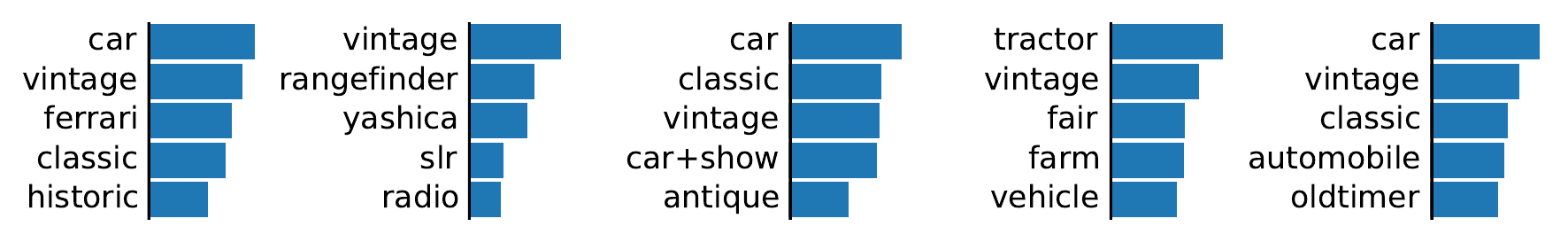}
        \end{subfigure} \\
	\vspace{-.1in}
        \begin{subfigure}[b]{\textwidth}
		\centering
                \includegraphics[width=1\textwidth]{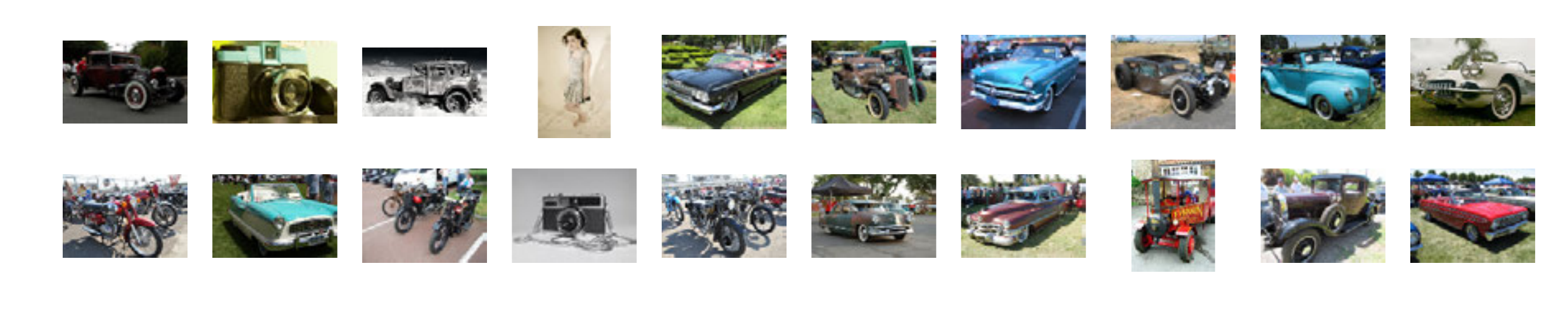}
		\vspace{-.4in}
                \caption{\mycaptionsize\tagname}
		\vspace{-.1in}
        \end{subfigure}
\end{figure*}

\begin{figure*}
        \centering
	\vspace{-.3in}
        \newcommand{\tagname}{horse}
        \begin{subfigure}[b]{\textwidth}
		\centering
                \includegraphics[width=.9\textwidth]{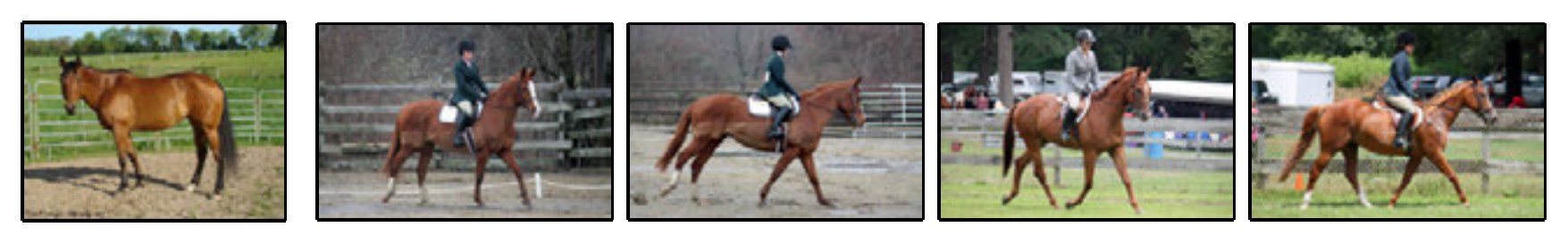}
        \end{subfigure} \\
	\vspace{-.1in}
        \begin{subfigure}[b]{\textwidth}
		\centering
                \includegraphics[width=.9\textwidth]{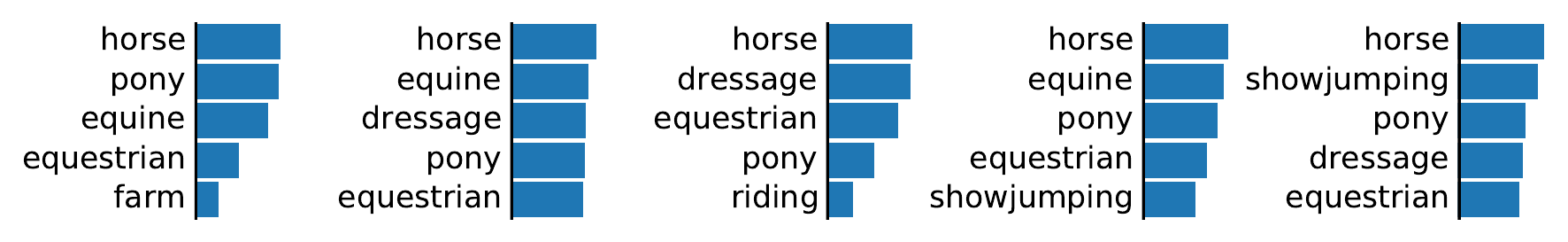}
        \end{subfigure} \\
	\vspace{-.1in}
        \begin{subfigure}[b]{\textwidth}
		\centering
                \includegraphics[width=1\textwidth]{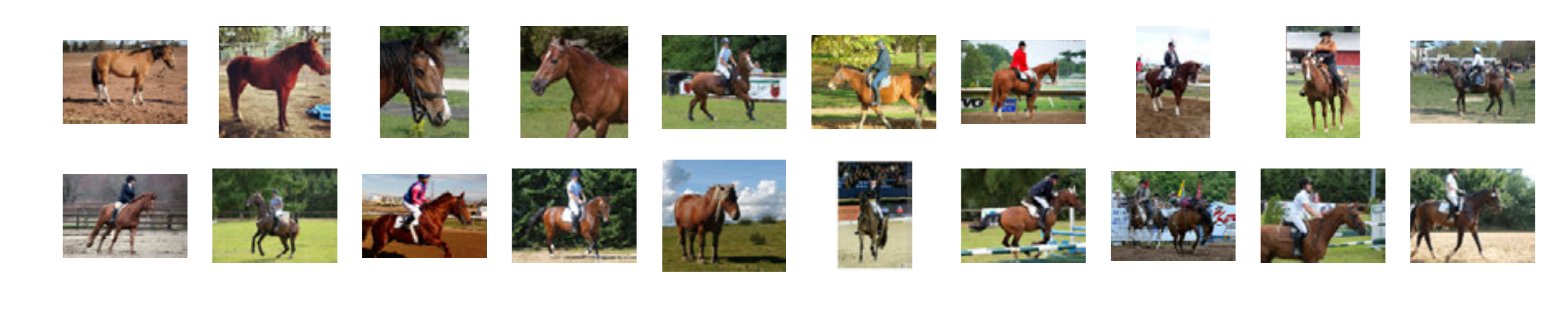}
		\vspace{-.4in}
                \caption{\mycaptionsize\tagname}
		\vspace{-.1in}
        \end{subfigure}
        \rule{500px}{1px}
	\vspace{-.1in}

        \renewcommand{\tagname}{cat}
        \begin{subfigure}[b]{\textwidth}
		\centering
                \includegraphics[width=.9\textwidth]{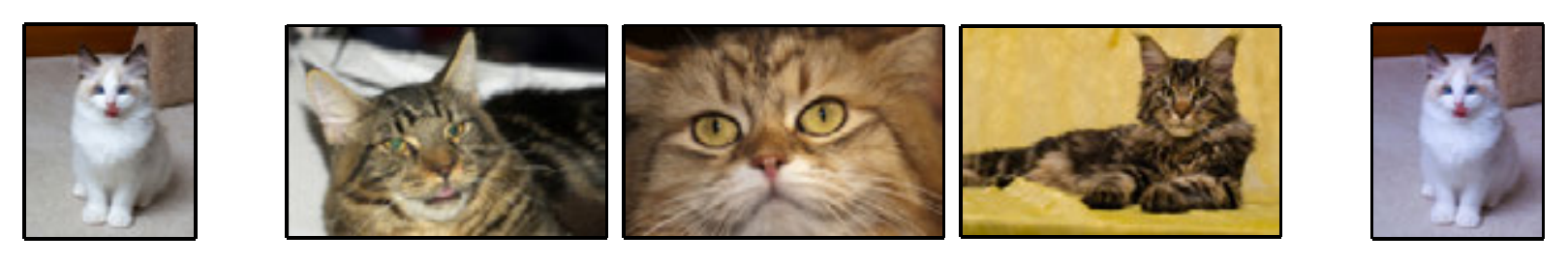}
        \end{subfigure} \\
	\vspace{-.1in}
        \begin{subfigure}[b]{\textwidth}
		\centering
                \includegraphics[width=.9\textwidth]{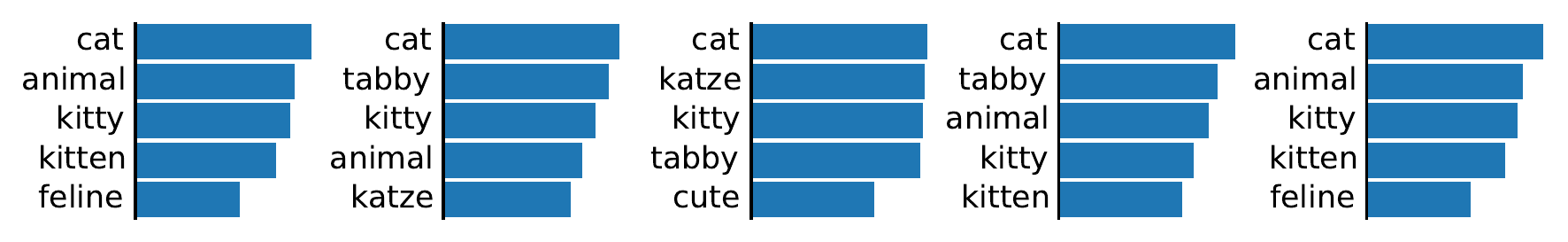}
        \end{subfigure} \\
	\vspace{-.1in}
        \begin{subfigure}[b]{\textwidth}
		\centering
                \includegraphics[width=1\textwidth]{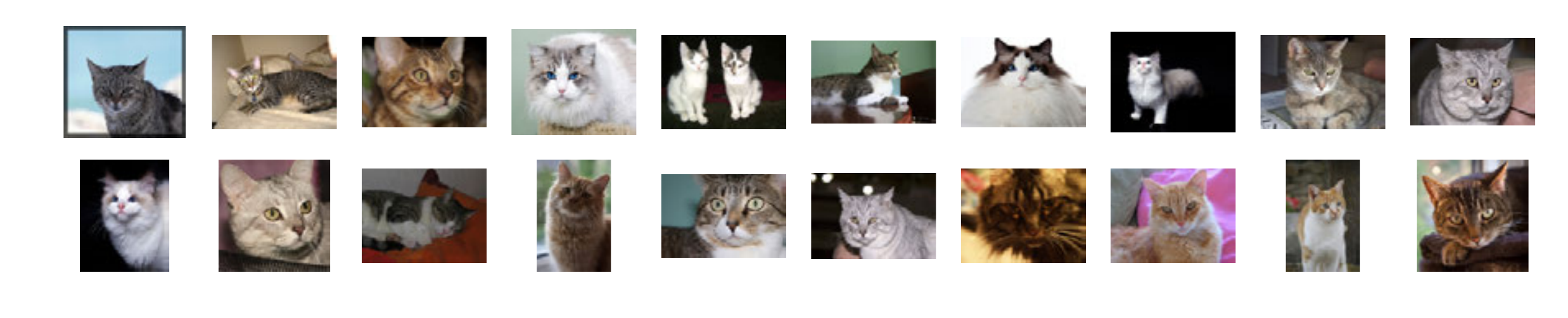}
		\vspace{-.4in}
                \caption{\mycaptionsize\tagname}
		\vspace{-.1in}
        \end{subfigure}
        \rule{500px}{1px}
	\vspace{-.1in}

        \renewcommand{\tagname}{dog}
        \begin{subfigure}[b]{\textwidth}
		\centering
                \includegraphics[width=.9\textwidth]{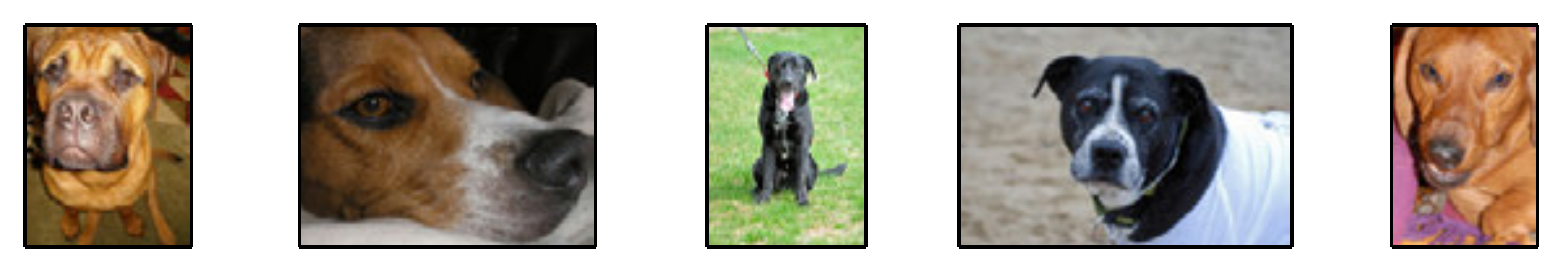}
        \end{subfigure} \\
	\vspace{-.1in}
        \begin{subfigure}[b]{\textwidth}
		\centering
                \includegraphics[width=.9\textwidth]{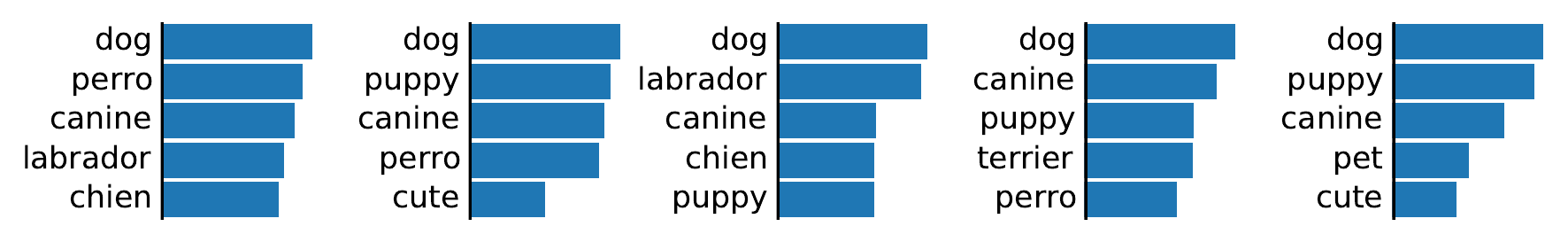}
        \end{subfigure} \\
	\vspace{-.1in}
        \begin{subfigure}[b]{\textwidth}
		\centering
                \includegraphics[width=1\textwidth]{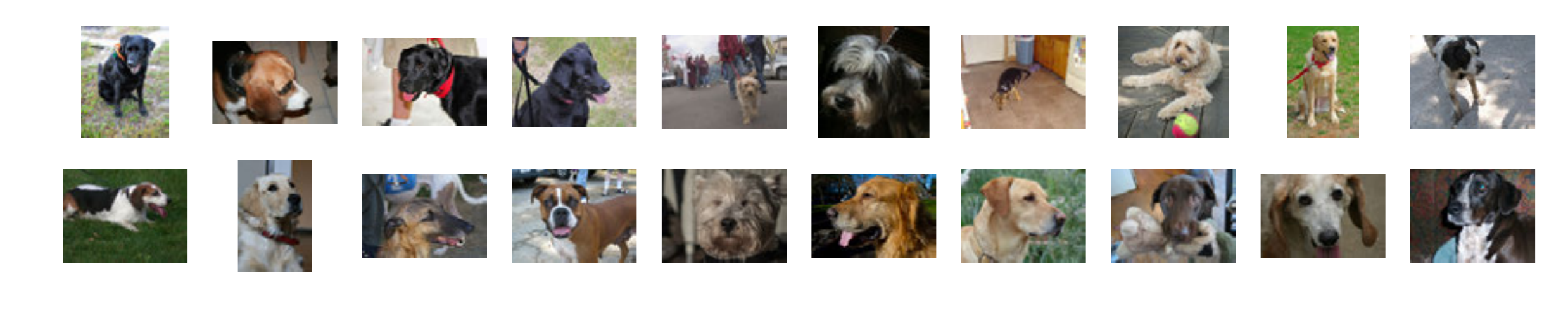}
		\vspace{-.4in}
                \caption{\mycaptionsize\tagname}
		\vspace{-.1in}
        \end{subfigure}
\end{figure*}

\begin{figure*}
        \centering
	\vspace{-.3in}
        \newcommand{\tagname}{rally}
        \begin{subfigure}[b]{\textwidth}
		\centering
                \includegraphics[width=.9\textwidth]{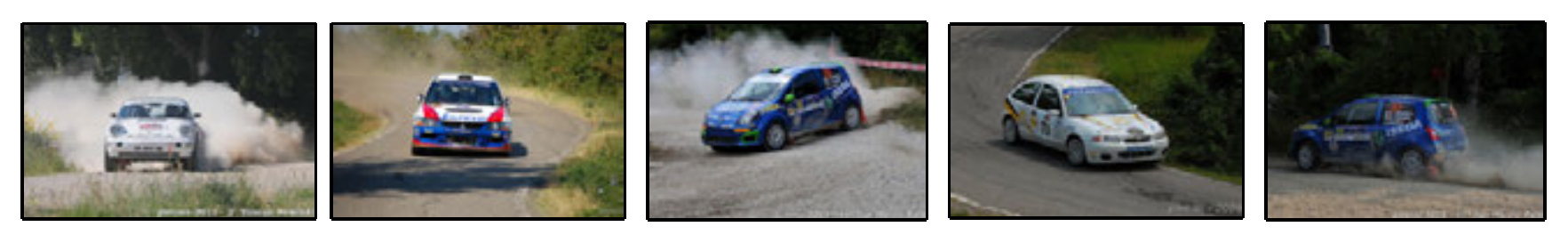}
        \end{subfigure} \\
	\vspace{-.1in}
        \begin{subfigure}[b]{\textwidth}
		\centering
                \includegraphics[width=.9\textwidth]{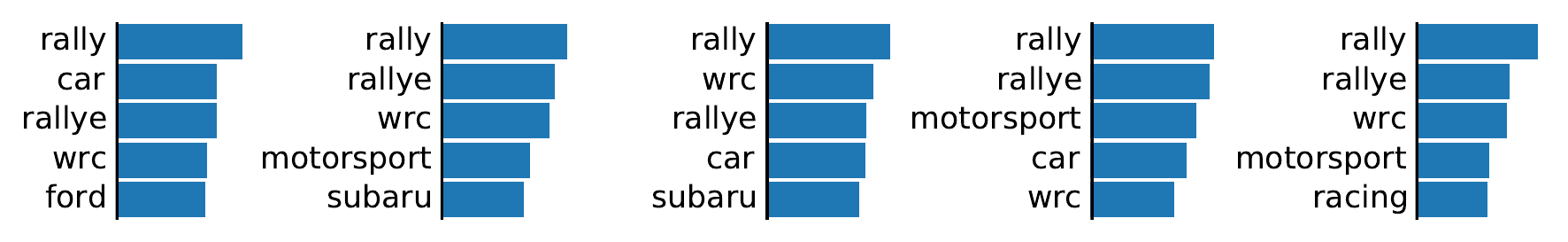}
        \end{subfigure} \\
	\vspace{-.1in}
        \begin{subfigure}[b]{\textwidth}
		\centering
                \includegraphics[width=1\textwidth]{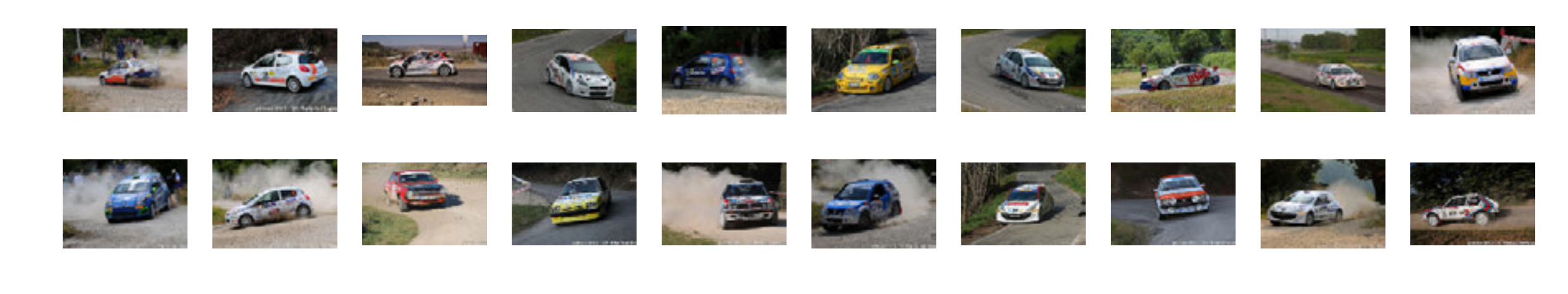}
		\vspace{-.4in}
                \caption{\mycaptionsize\tagname}
		\vspace{-.1in}
        \end{subfigure}
        \rule{500px}{1px}
	\vspace{-.1in}

        \renewcommand{\tagname}{graduation}
        \begin{subfigure}[b]{\textwidth}
		\centering
                \includegraphics[width=.9\textwidth]{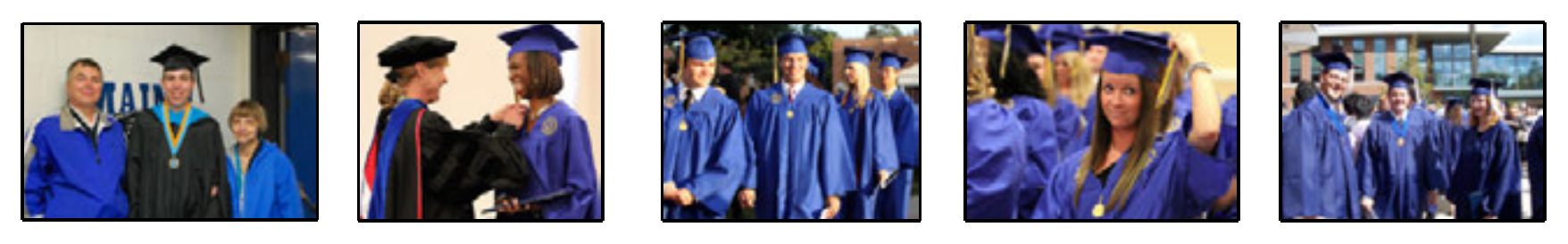}
        \end{subfigure} \\
	\vspace{-.1in}
        \begin{subfigure}[b]{\textwidth}
		\centering
                \includegraphics[width=.9\textwidth]{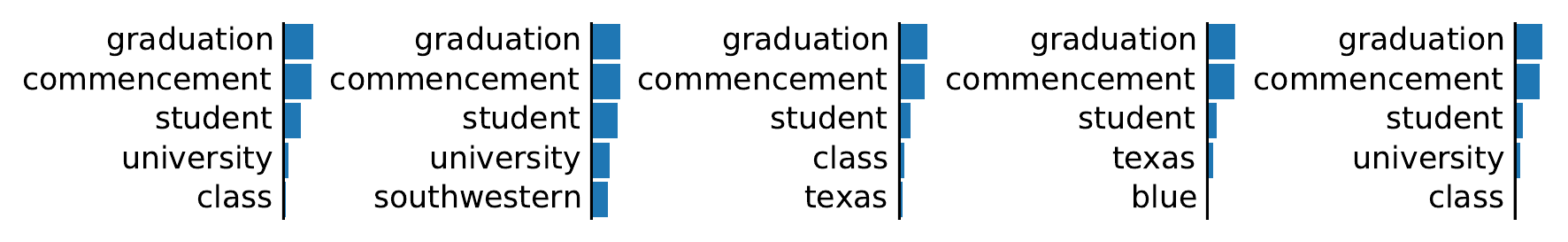}
        \end{subfigure} \\
	\vspace{-.1in}
        \begin{subfigure}[b]{\textwidth}
		\centering
                \includegraphics[width=1\textwidth]{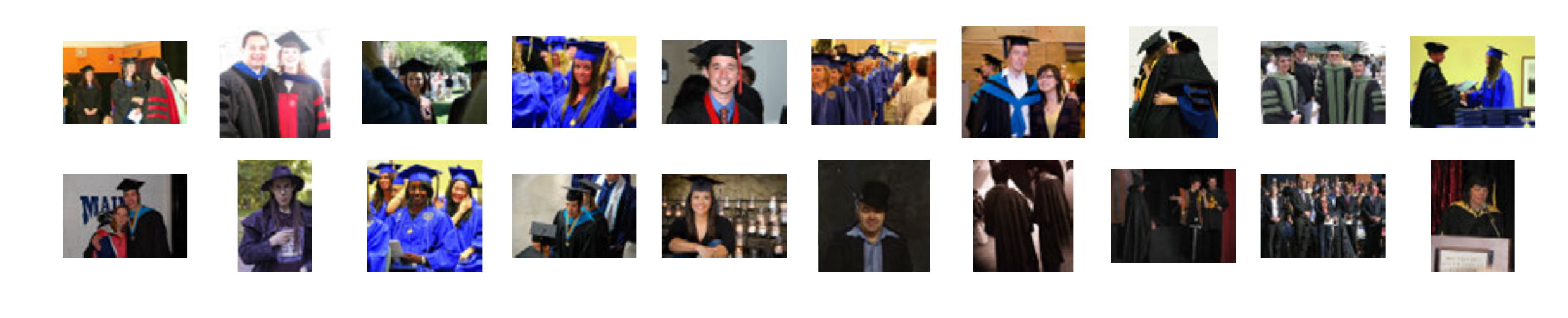}
		\vspace{-.4in}
                \caption{\mycaptionsize\tagname}
		\vspace{-.1in}
        \end{subfigure}
        \rule{500px}{1px}
	\vspace{-.1in}

        \renewcommand{\tagname}{flamingo}
        \begin{subfigure}[b]{\textwidth}
		\centering
                \includegraphics[width=.9\textwidth]{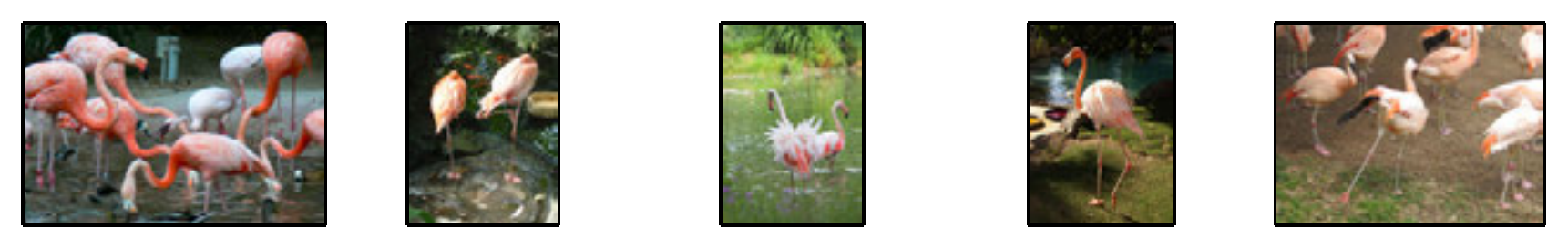}
        \end{subfigure} \\
	\vspace{-.1in}
        \begin{subfigure}[b]{\textwidth}
		\centering
                \includegraphics[width=.9\textwidth]{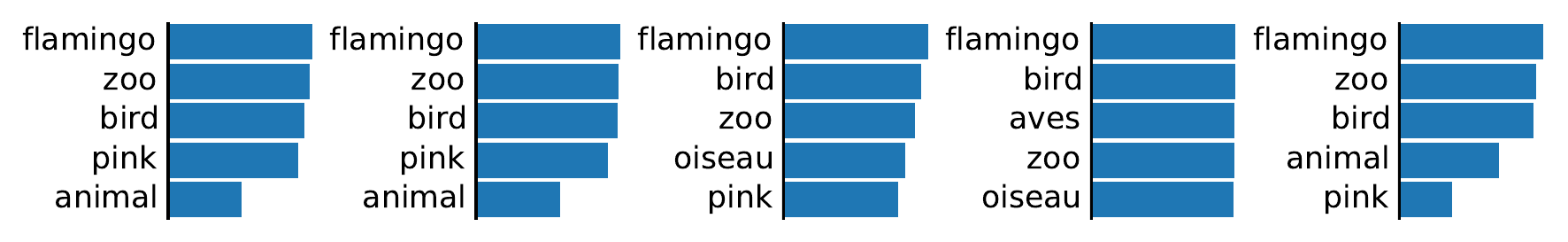}
        \end{subfigure} \\
	\vspace{-.1in}
        \begin{subfigure}[b]{\textwidth}
		\centering
                \includegraphics[width=1\textwidth]{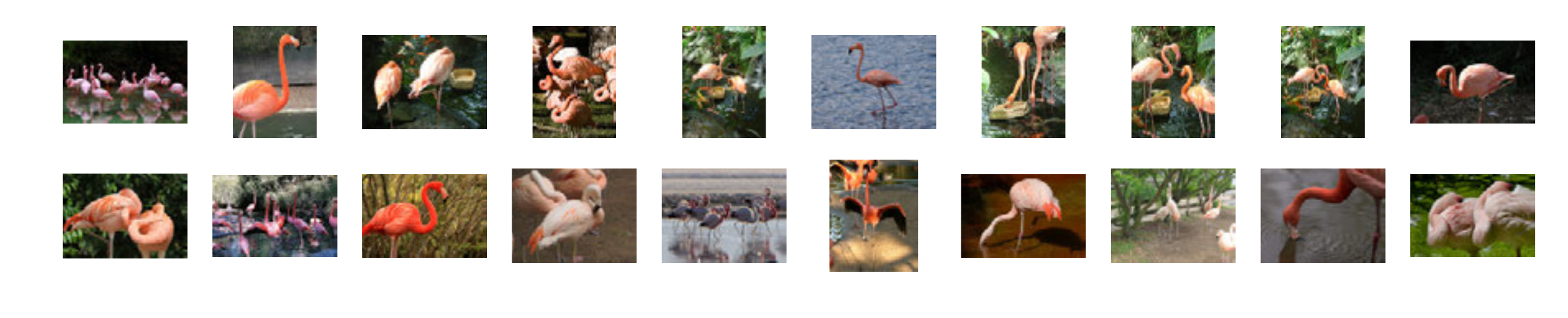}
		\vspace{-.4in}
                \caption{\mycaptionsize\tagname}
		\vspace{-.1in}
        \end{subfigure}
\end{figure*}

\addtocounter{subfigure}{-24}
\clearpage

\begin{figure*}
        \centering
	\vspace{-.3in}
        \newcommand{\tagname}{eiffel}
        \begin{subfigure}[b]{\textwidth}
		\centering
                \includegraphics[width=.9\textwidth]{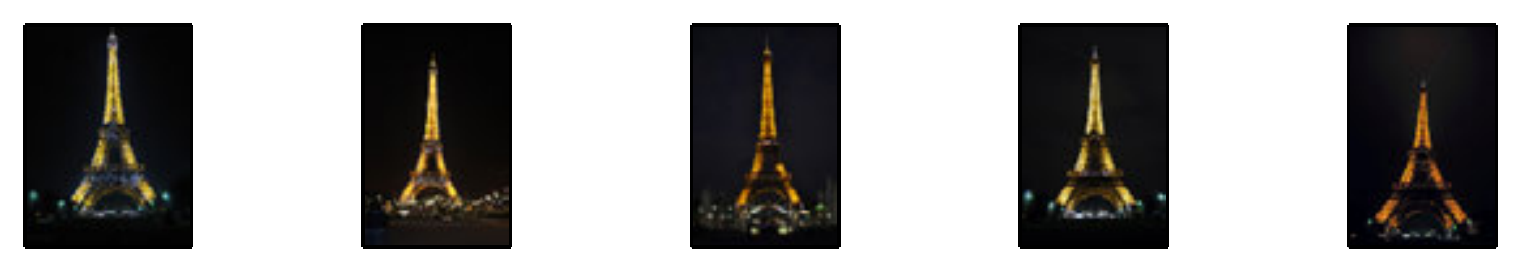}
        \end{subfigure} \\
	\vspace{-.1in}
        \begin{subfigure}[b]{\textwidth}
		\centering
                \includegraphics[width=.9\textwidth]{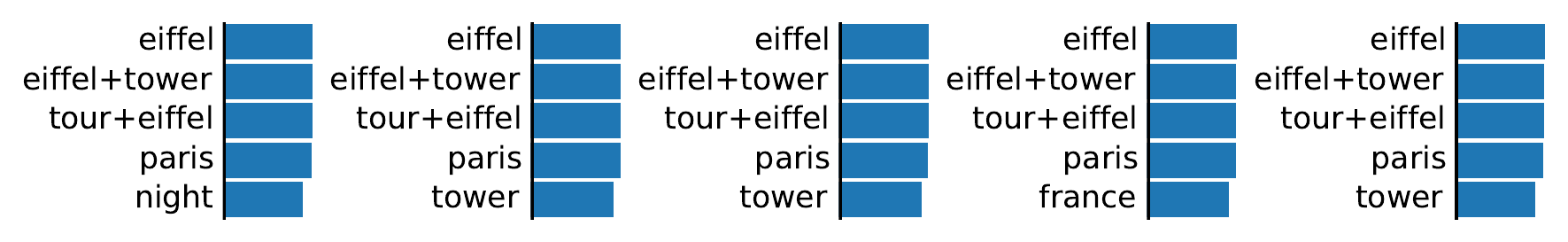}
        \end{subfigure} \\
	\vspace{-.1in}
        \begin{subfigure}[b]{\textwidth}
		\centering
                \includegraphics[width=1\textwidth]{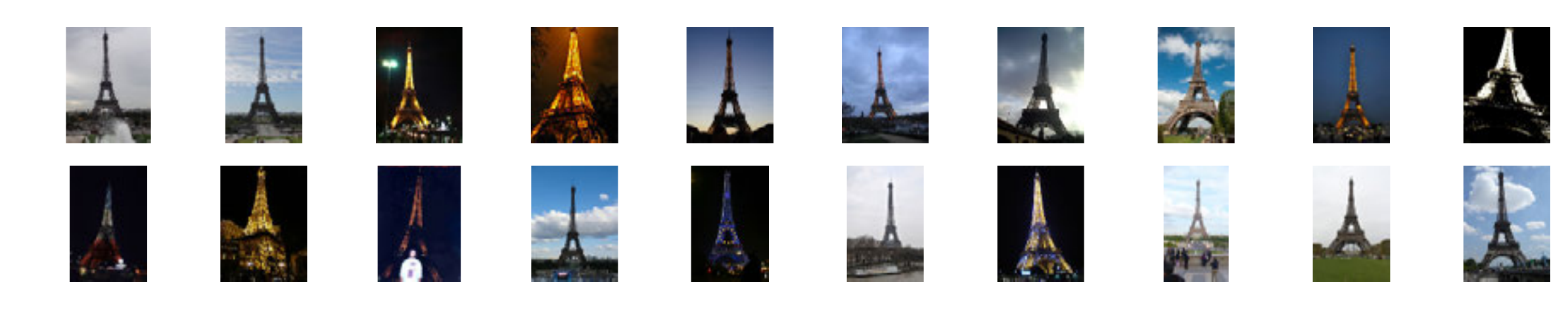}
		\vspace{-.4in}
                \caption{\mycaptionsize\tagname}
		\vspace{-.1in}
        \end{subfigure}
        \rule{500px}{1px}
	\vspace{-.1in}

        \renewcommand{\tagname}{tower+bridge}
        \begin{subfigure}[b]{\textwidth}
		\centering
                \includegraphics[width=.9\textwidth]{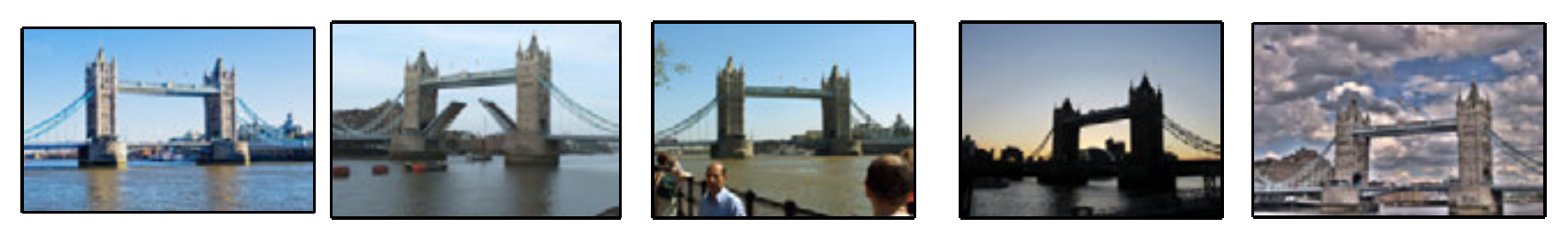}
        \end{subfigure} \\
	\vspace{-.1in}
        \begin{subfigure}[b]{\textwidth}
		\centering
                \includegraphics[width=.9\textwidth]{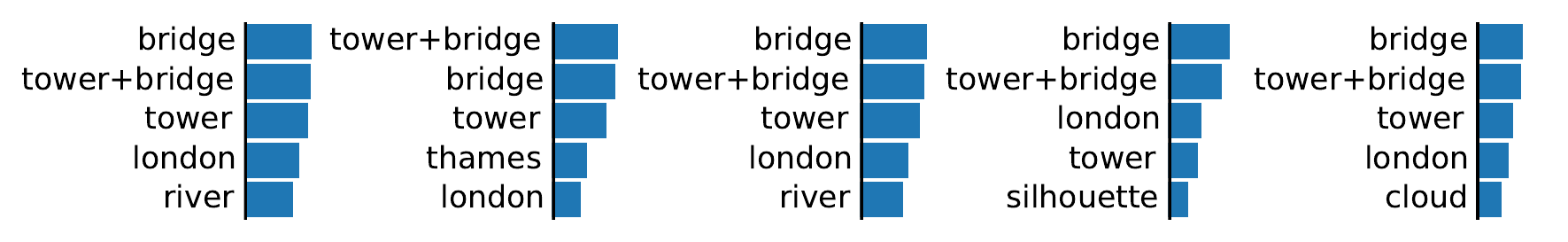}
        \end{subfigure} \\
	\vspace{-.1in}
        \begin{subfigure}[b]{\textwidth}
		\centering
                \includegraphics[width=1\textwidth]{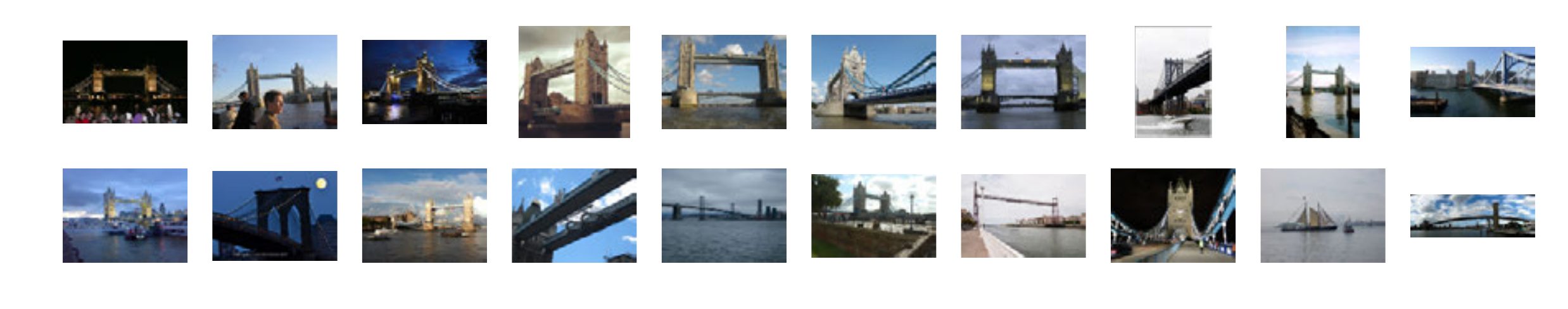}
		\vspace{-.4in}
                \caption{\mycaptionsize\tagname}
		\vspace{-.1in}
        \end{subfigure}
        \rule{500px}{1px}
	\vspace{-.1in}

        \renewcommand{\tagname}{yellowstone+national+park}
        \begin{subfigure}[b]{\textwidth}
		\centering
                \includegraphics[width=.9\textwidth]{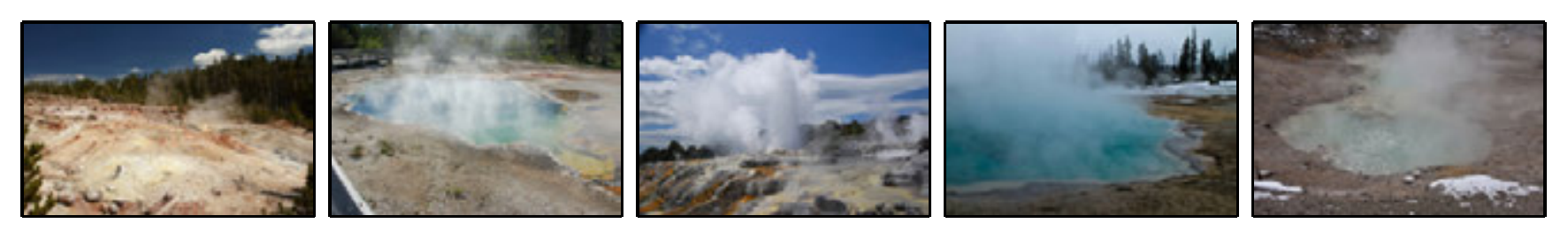}
        \end{subfigure} \\
	\vspace{-.1in}
        \begin{subfigure}[b]{\textwidth}
		\centering
                \includegraphics[width=.9\textwidth]{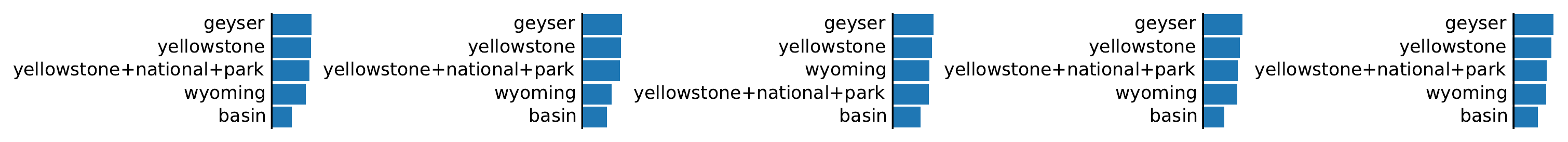}
        \end{subfigure} \\
	\vspace{-.01in}
        \begin{subfigure}[b]{\textwidth}
		\centering
                \includegraphics[width=1\textwidth]{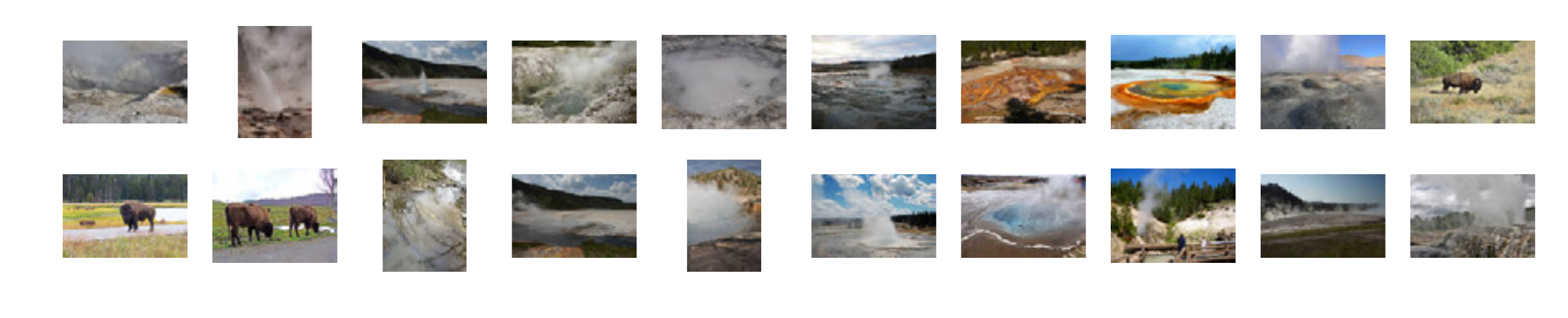}
		\vspace{-.4in}
                \caption{\mycaptionsize\tagname}
		\vspace{-.1in}
        \end{subfigure}
\end{figure*}

\begin{figure*}
        \centering
	\vspace{-.3in}
        \newcommand{\tagname}{bead}
        \begin{subfigure}[b]{\textwidth}
		\centering
                \includegraphics[width=.9\textwidth]{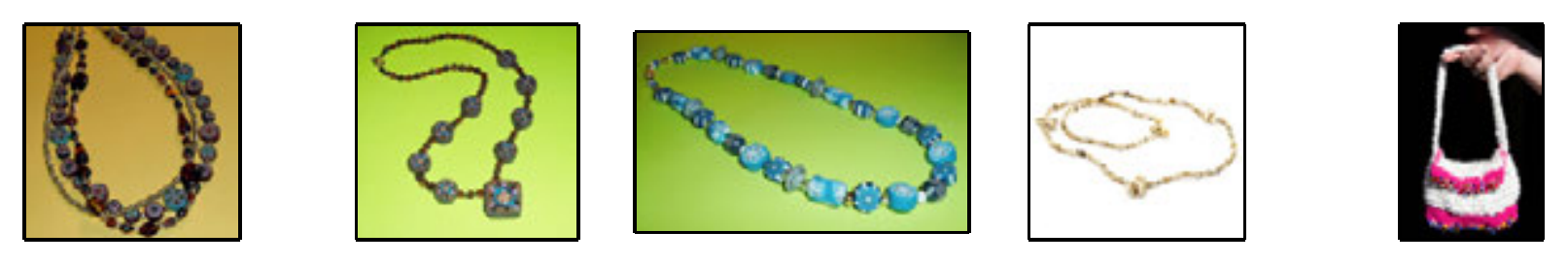}
        \end{subfigure} \\
	\vspace{-.1in}
        \begin{subfigure}[b]{\textwidth}
		\centering
                \includegraphics[width=.9\textwidth]{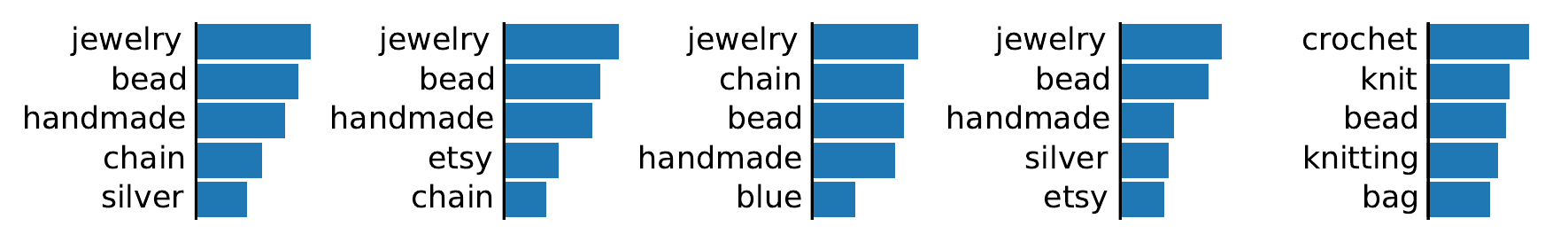}
        \end{subfigure} \\
	\vspace{-.1in}
        \begin{subfigure}[b]{\textwidth}
		\centering
                \includegraphics[width=1\textwidth]{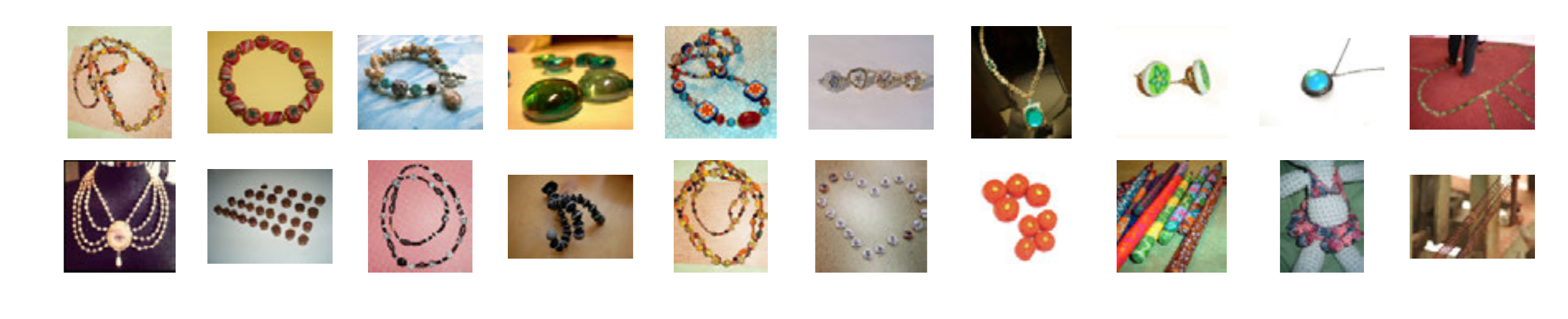}
		\vspace{-.4in}
                \caption{\mycaptionsize\tagname}
		\vspace{-.1in}
        \end{subfigure}
        \rule{500px}{1px}
	\vspace{-.1in}

        \renewcommand{\tagname}{chain}
        \begin{subfigure}[b]{\textwidth}
		\centering
                \includegraphics[width=.9\textwidth]{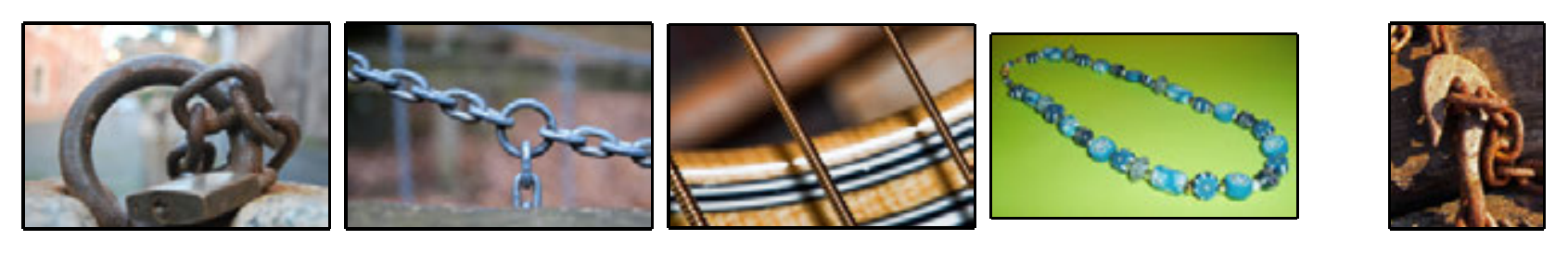}
        \end{subfigure} \\
	\vspace{-.1in}
        \begin{subfigure}[b]{\textwidth}
		\centering
                \includegraphics[width=.9\textwidth]{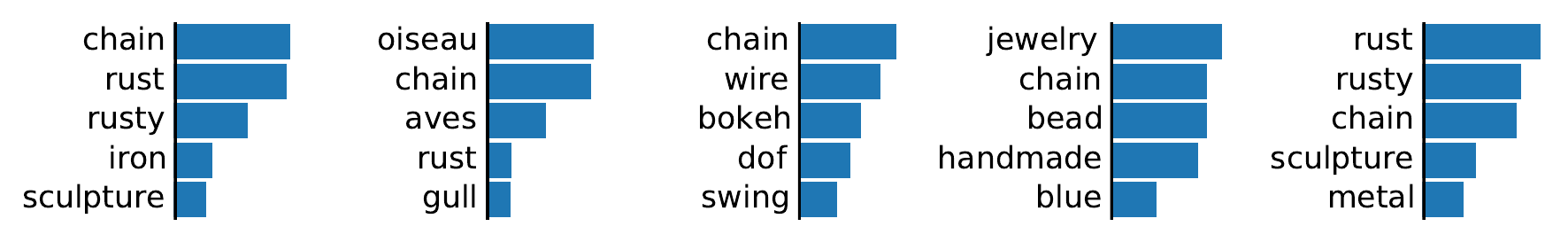}
        \end{subfigure} \\
	\vspace{-.1in}
        \begin{subfigure}[b]{\textwidth}
		\centering
                \includegraphics[width=1\textwidth]{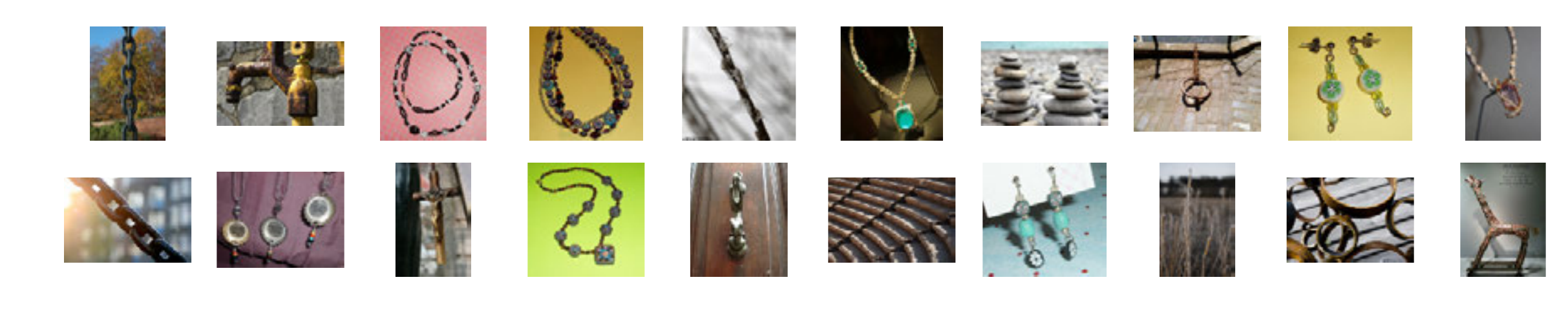}
		\vspace{-.4in}
                \caption{\mycaptionsize\tagname}
		\vspace{-.1in}
        \end{subfigure}
        \rule{500px}{1px}
	\vspace{-.1in}

        \renewcommand{\tagname}{chair}
        \begin{subfigure}[b]{\textwidth}
		\centering
                \includegraphics[width=.9\textwidth]{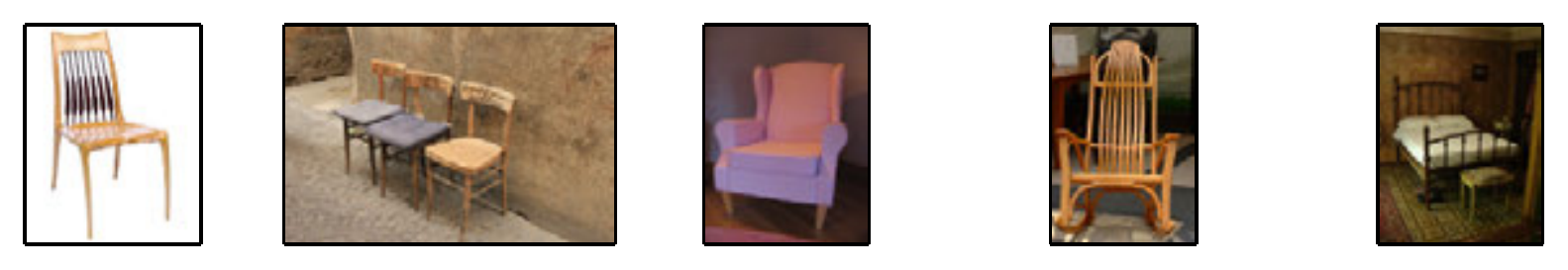}
        \end{subfigure} \\
	\vspace{-.1in}
        \begin{subfigure}[b]{\textwidth}
		\centering
                \includegraphics[width=.9\textwidth]{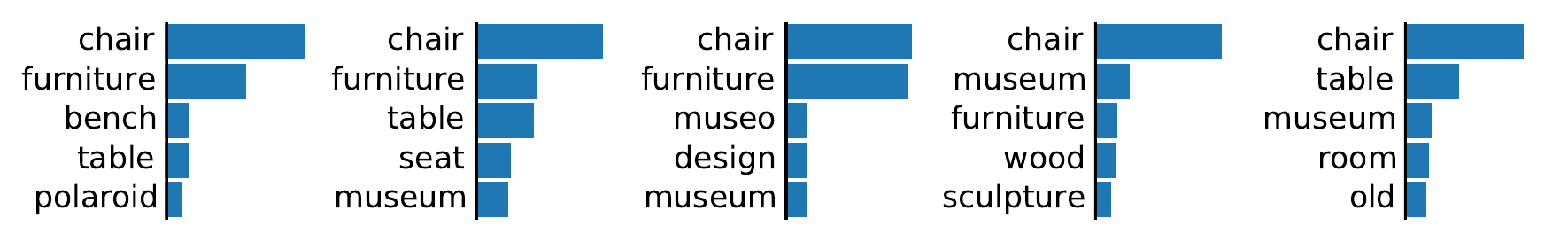}
        \end{subfigure} \\
	\vspace{-.01in}
        \begin{subfigure}[b]{\textwidth}
		\centering
                \includegraphics[width=1\textwidth]{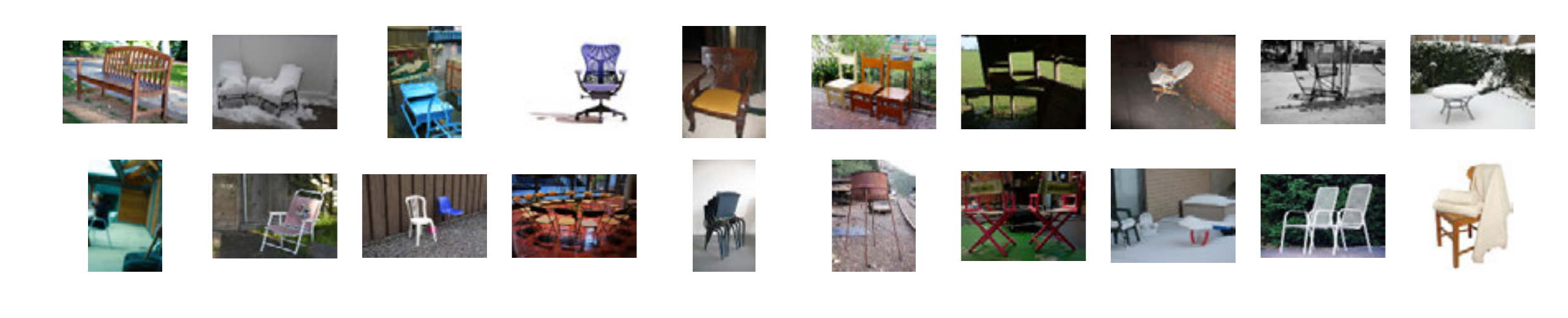}
		\vspace{-.4in}
                \caption{\mycaptionsize\tagname}
		\vspace{-.1in}
        \end{subfigure}
\end{figure*}

\begin{figure*}
        \centering
	\vspace{-.3in}
        \newcommand{\tagname}{hawaii}
        \begin{subfigure}[b]{\textwidth}
		\centering
                \includegraphics[width=.9\textwidth]{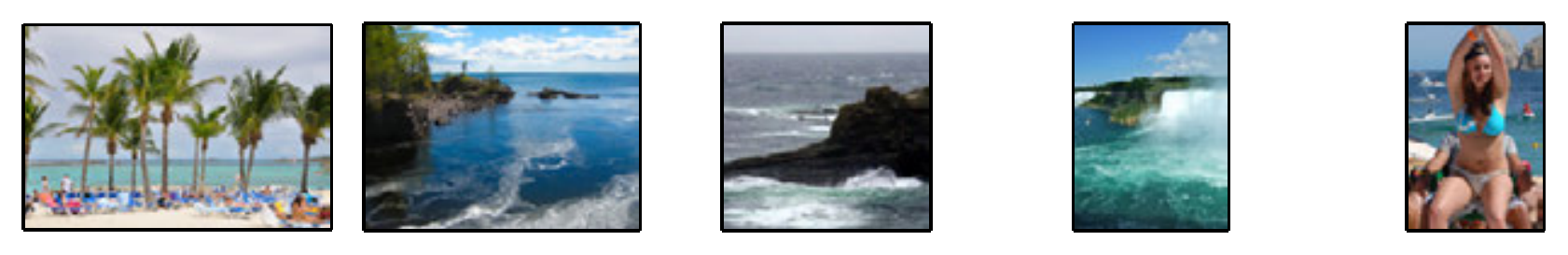}
        \end{subfigure} \\
	\vspace{-.1in}
        \begin{subfigure}[b]{\textwidth}
		\centering
                \includegraphics[width=.9\textwidth]{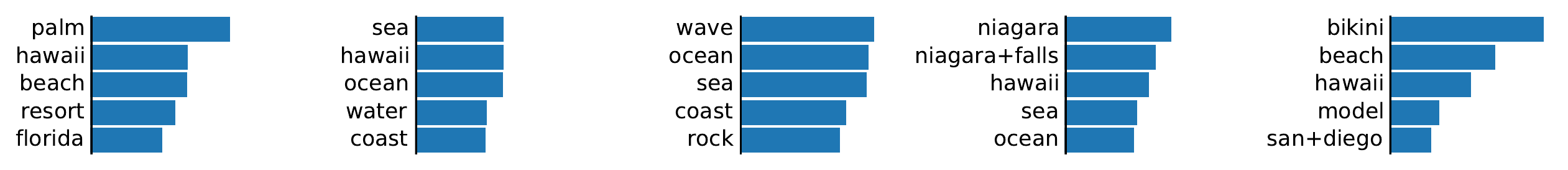}
        \end{subfigure} \\
	\vspace{-.1in}
        \begin{subfigure}[b]{\textwidth}
		\centering
                \includegraphics[width=1\textwidth]{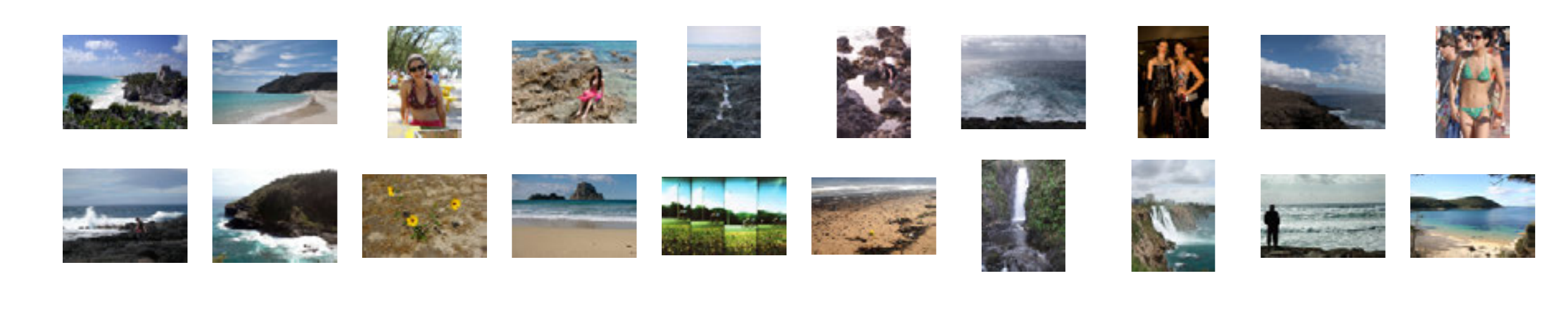}
		\vspace{-.4in}
                \caption{\mycaptionsize\tagname}
		\vspace{-.1in}
        \end{subfigure}
        \rule{500px}{1px}
	\vspace{-.1in}

        \renewcommand{\tagname}{brooklyn+bridge}
        \begin{subfigure}[b]{\textwidth}
		\centering
                \includegraphics[width=.9\textwidth]{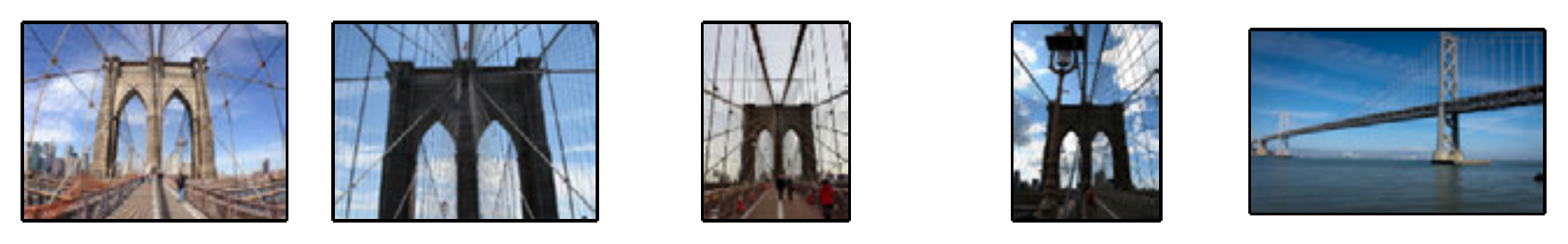}
        \end{subfigure} \\
	\vspace{-.1in}
        \begin{subfigure}[b]{\textwidth}
		\centering
                \includegraphics[width=.9\textwidth]{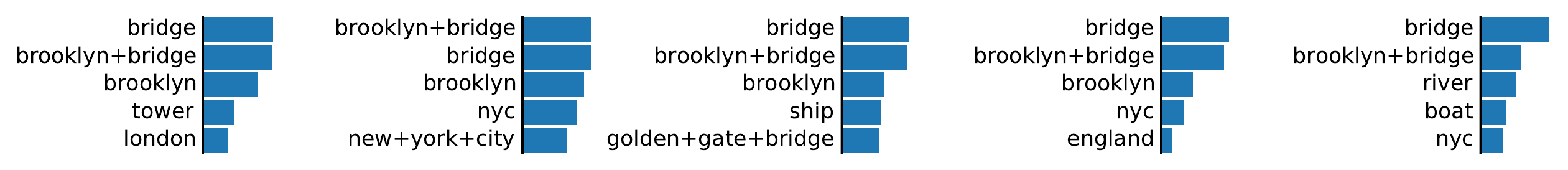}
        \end{subfigure} \\
	\vspace{-.1in}
        \begin{subfigure}[b]{\textwidth}
		\centering
                \includegraphics[width=1\textwidth]{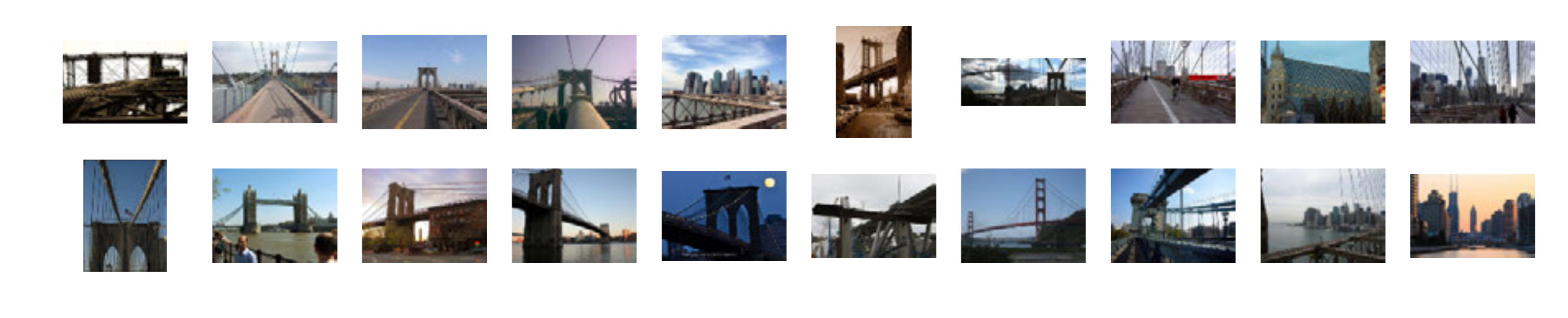}
		\vspace{-.4in}
                \caption{\mycaptionsize\tagname}
		\vspace{-.1in}
        \end{subfigure}
        \rule{500px}{1px}
	\vspace{-.1in}

        \renewcommand{\tagname}{golden+gate+bridge}
        \begin{subfigure}[b]{\textwidth}
		\centering
                \includegraphics[width=.9\textwidth]{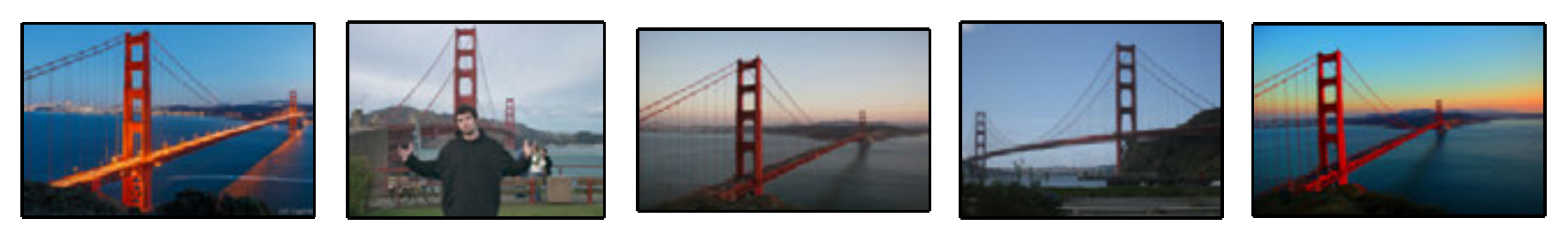}
        \end{subfigure} \\
	\vspace{-.1in}
        \begin{subfigure}[b]{\textwidth}
		\centering
                \includegraphics[width=.9\textwidth]{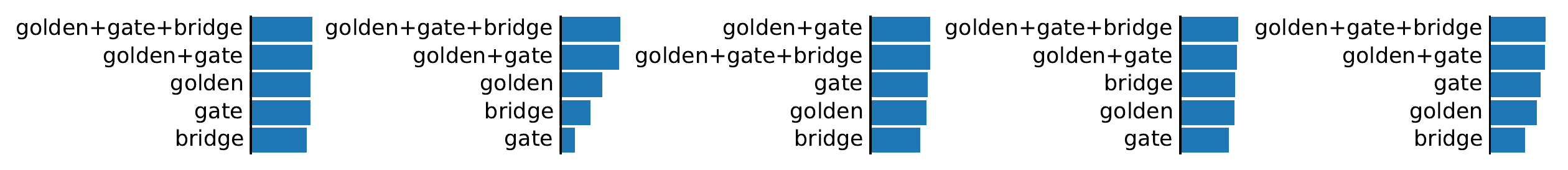}
        \end{subfigure} \\
	\vspace{-.01in}
        \begin{subfigure}[b]{\textwidth}
		\centering
                \includegraphics[width=1\textwidth]{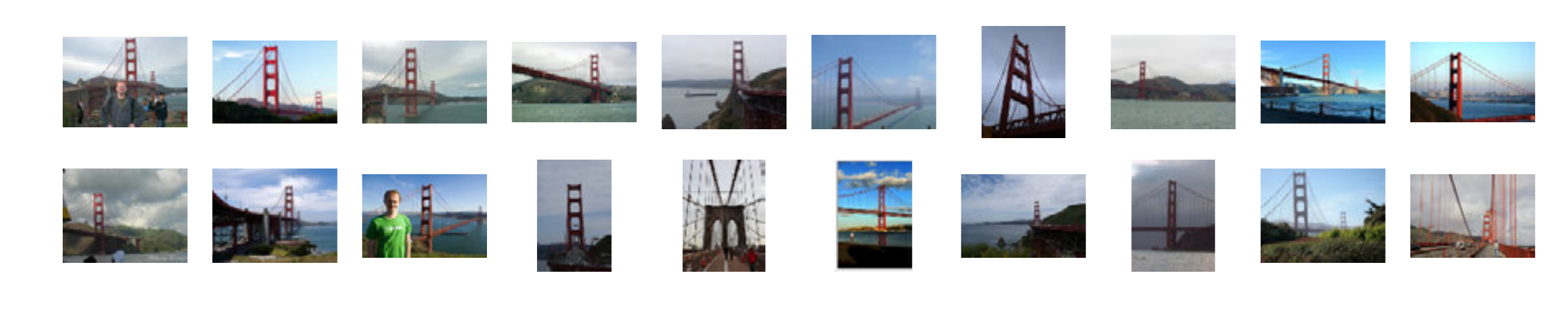}
		\vspace{-.4in}
                \caption{\mycaptionsize\tagname}
		\vspace{-.1in}
        \end{subfigure}
\end{figure*}

\begin{figure*}
        \centering
	\vspace{-.3in}
        \newcommand{\tagname}{graffiti}
        \begin{subfigure}[b]{\textwidth}
		\centering
                \includegraphics[width=.9\textwidth]{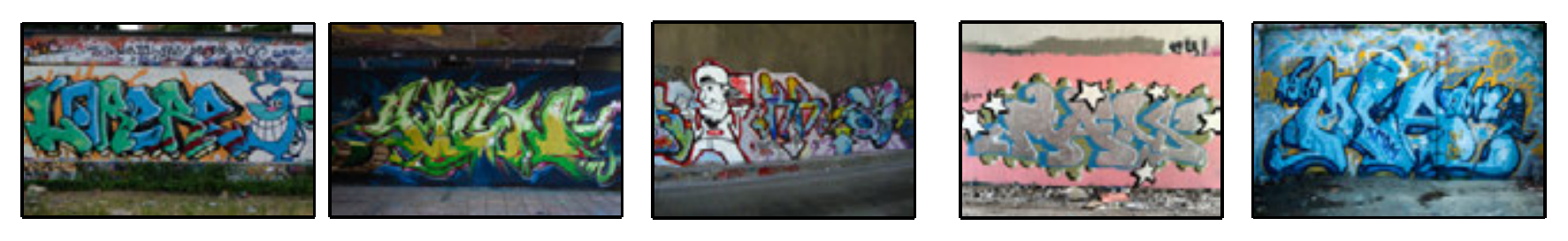}
        \end{subfigure} \\
	\vspace{-.1in}
        \begin{subfigure}[b]{\textwidth}
		\centering
                \includegraphics[width=.9\textwidth]{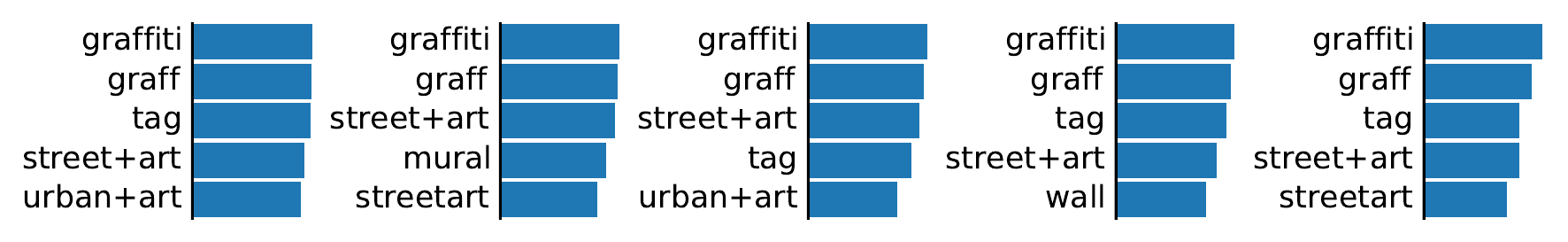}
        \end{subfigure} \\
	\vspace{-.1in}
        \begin{subfigure}[b]{\textwidth}
		\centering
                \includegraphics[width=1\textwidth]{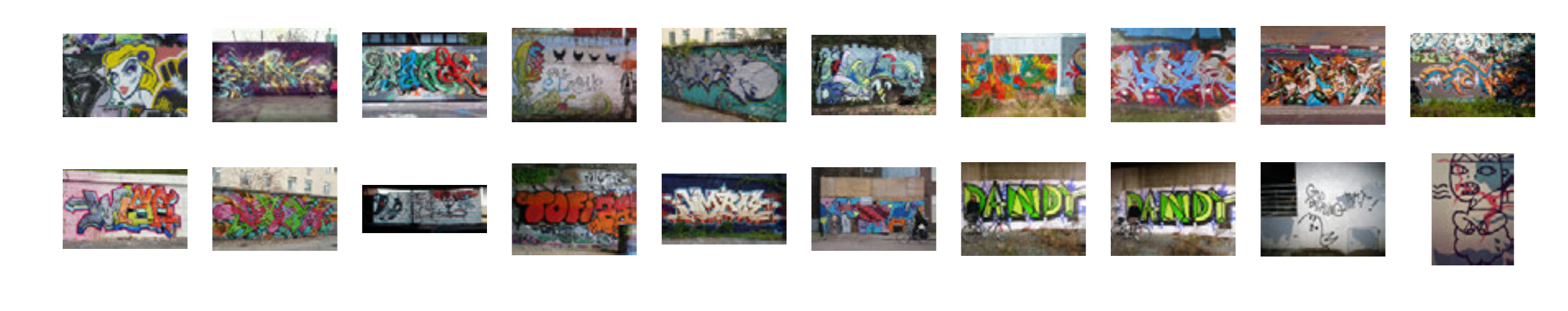}
		\vspace{-.4in}
                \caption{\mycaptionsize\tagname}
		\vspace{-.1in}
        \end{subfigure}
        \rule{500px}{1px}
	\vspace{-.1in}

        \renewcommand{\tagname}{firework}
        \begin{subfigure}[b]{\textwidth}
		\centering
                \includegraphics[width=.9\textwidth]{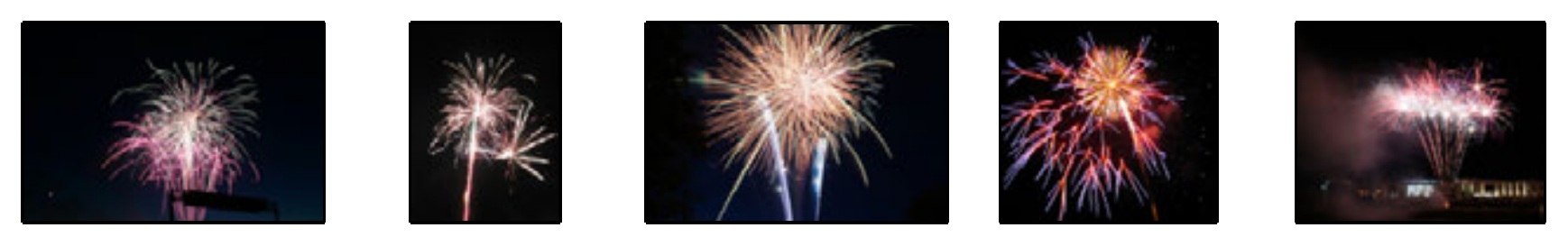}
        \end{subfigure} \\
	\vspace{-.1in}
        \begin{subfigure}[b]{\textwidth}
		\centering
                \includegraphics[width=.9\textwidth]{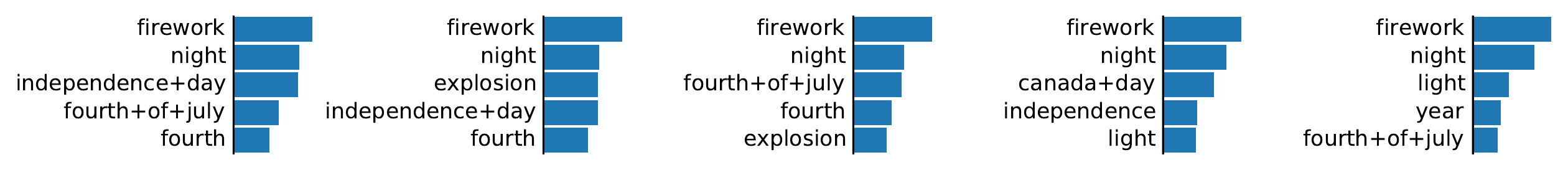}
        \end{subfigure} \\
	\vspace{-.1in}
        \begin{subfigure}[b]{\textwidth}
		\centering
                \includegraphics[width=1\textwidth]{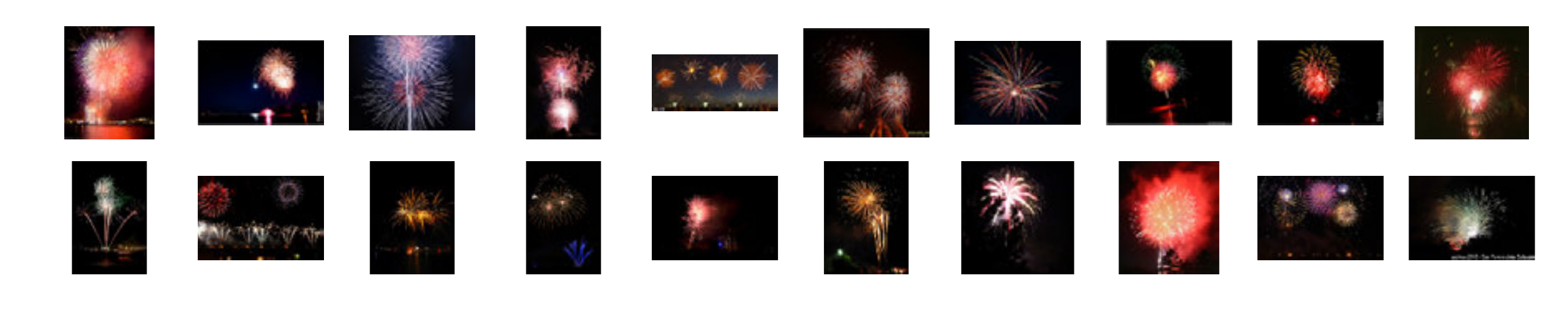}
		\vspace{-.4in}
                \caption{\mycaptionsize\tagname}
		\vspace{-.1in}
        \end{subfigure}
        \rule{500px}{1px}
	\vspace{-.1in}

        \renewcommand{\tagname}{rainbow}
        \begin{subfigure}[b]{\textwidth}
		\centering
                \includegraphics[width=.9\textwidth]{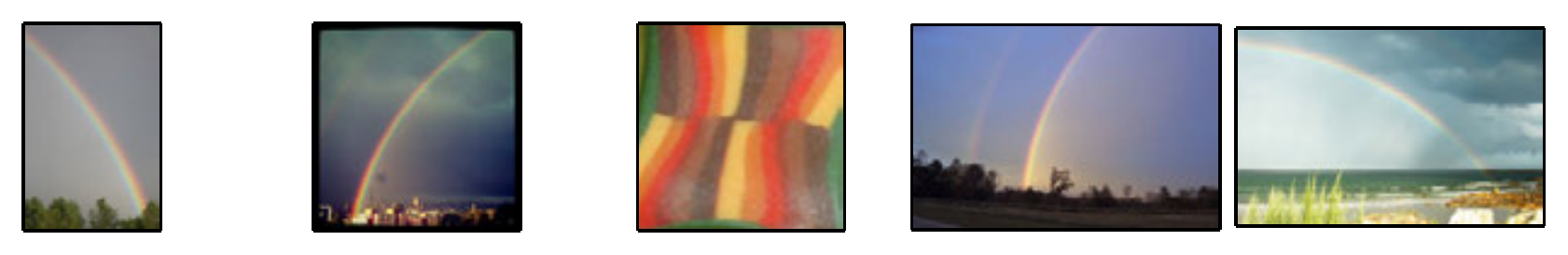}
        \end{subfigure} \\
	\vspace{-.1in}
        \begin{subfigure}[b]{\textwidth}
		\centering
                \includegraphics[width=.9\textwidth]{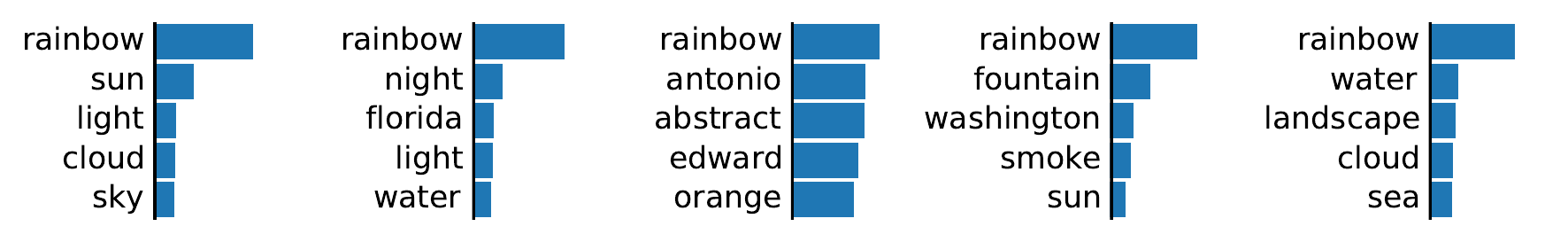}
        \end{subfigure} \\
	\vspace{-.01in}
        \begin{subfigure}[b]{\textwidth}
		\centering
                \includegraphics[width=1\textwidth]{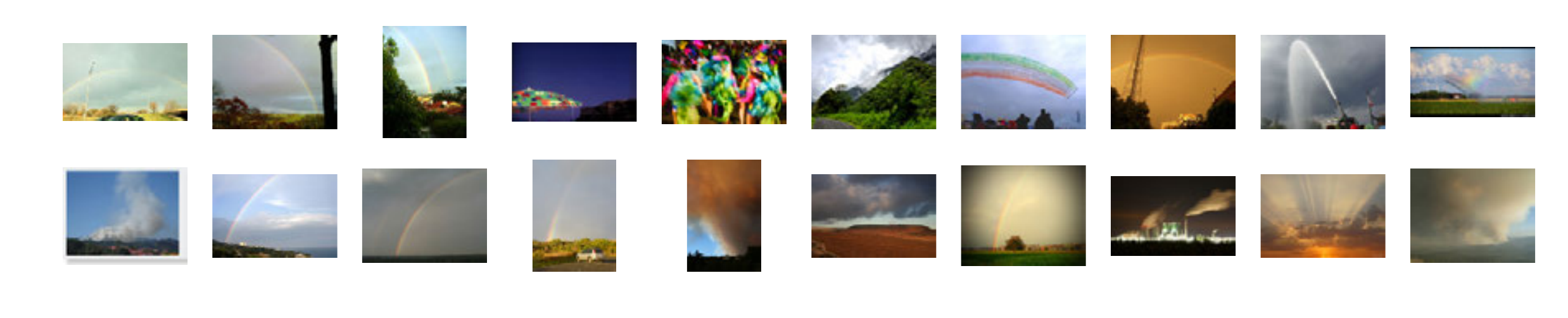}
		\vspace{-.4in}
                \caption{\mycaptionsize\tagname}
		\vspace{-.1in}
        \end{subfigure}
\end{figure*}

\begin{figure*}
        \centering
	\vspace{-.3in}
        \newcommand{\tagname}{bus}
        \begin{subfigure}[b]{\textwidth}
		\centering
                \includegraphics[width=.9\textwidth]{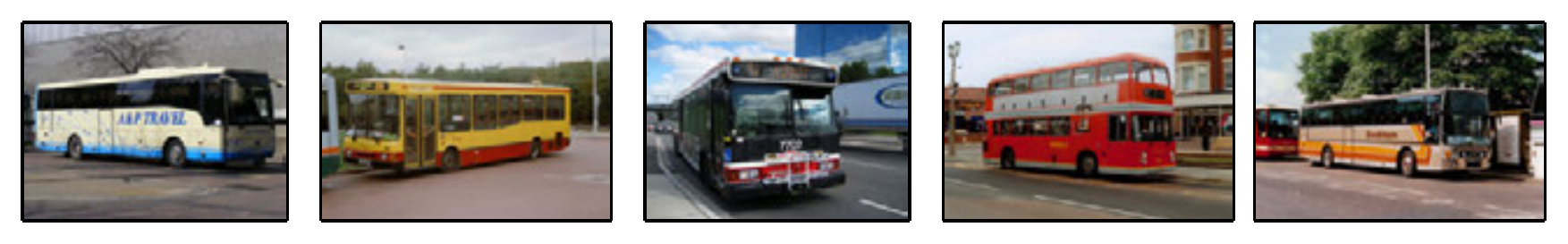}
        \end{subfigure} \\
	\vspace{-.1in}
        \begin{subfigure}[b]{\textwidth}
		\centering
                \includegraphics[width=.9\textwidth]{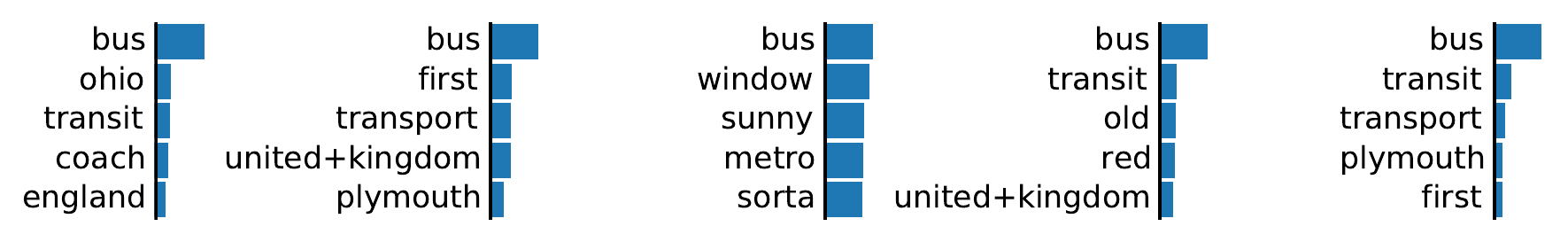}
        \end{subfigure} \\
	\vspace{-.1in}
        \begin{subfigure}[b]{\textwidth}
		\centering
                \includegraphics[width=1\textwidth]{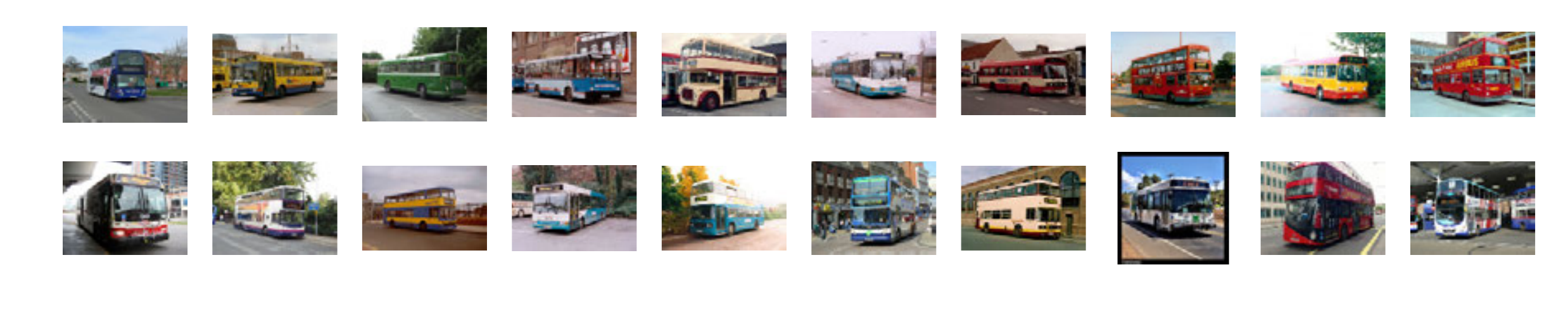}
		\vspace{-.4in}
                \caption{\mycaptionsize\tagname}
		\vspace{-.1in}
        \end{subfigure}
        \rule{500px}{1px}
	\vspace{-.1in}

        \renewcommand{\tagname}{bicycle}
        \begin{subfigure}[b]{\textwidth}
		\centering
                \includegraphics[width=.9\textwidth]{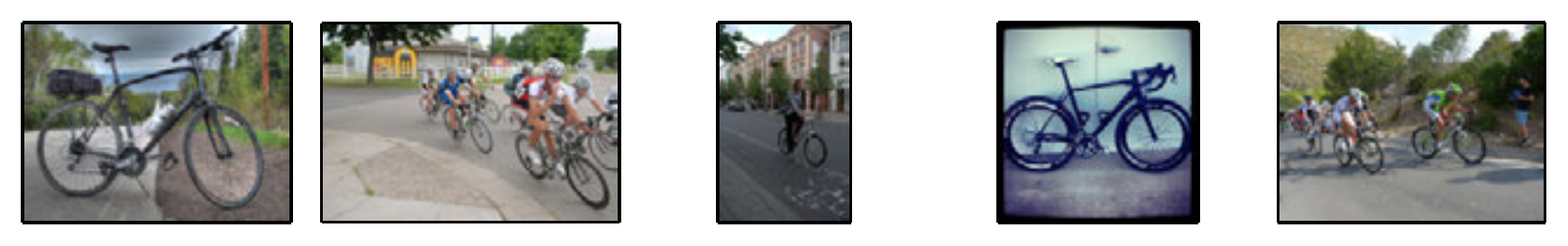}
        \end{subfigure} \\
	\vspace{-.1in}
        \begin{subfigure}[b]{\textwidth}
		\centering
                \includegraphics[width=.9\textwidth]{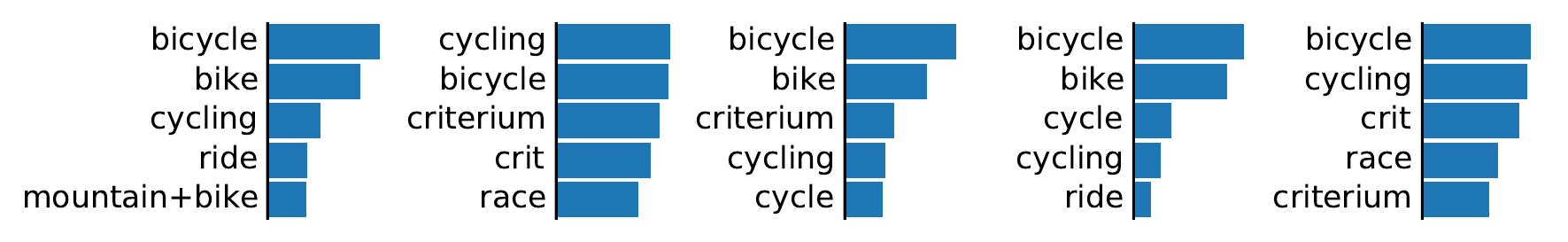}
        \end{subfigure} \\
	\vspace{-.1in}
        \begin{subfigure}[b]{\textwidth}
		\centering
                \includegraphics[width=1\textwidth]{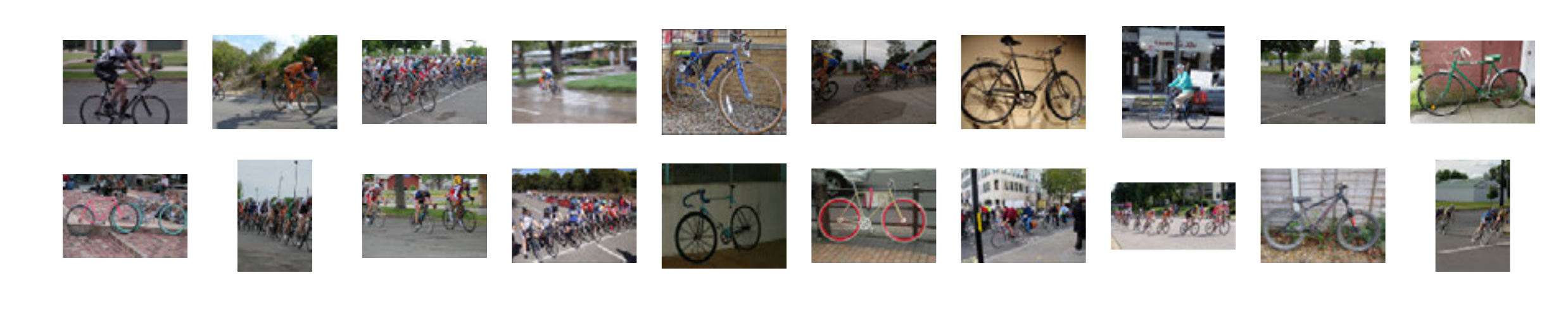}
		\vspace{-.4in}
                \caption{\mycaptionsize\tagname}
		\vspace{-.1in}
        \end{subfigure}
        \rule{500px}{1px}
	\vspace{-.1in}

        \renewcommand{\tagname}{train}
        \begin{subfigure}[b]{\textwidth}
		\centering
                \includegraphics[width=.9\textwidth]{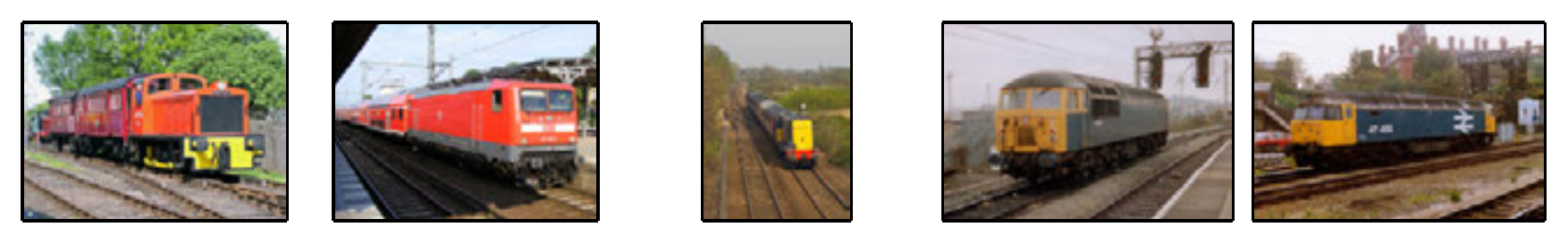}
        \end{subfigure} \\
	\vspace{-.1in}
        \begin{subfigure}[b]{\textwidth}
		\centering
                \includegraphics[width=.9\textwidth]{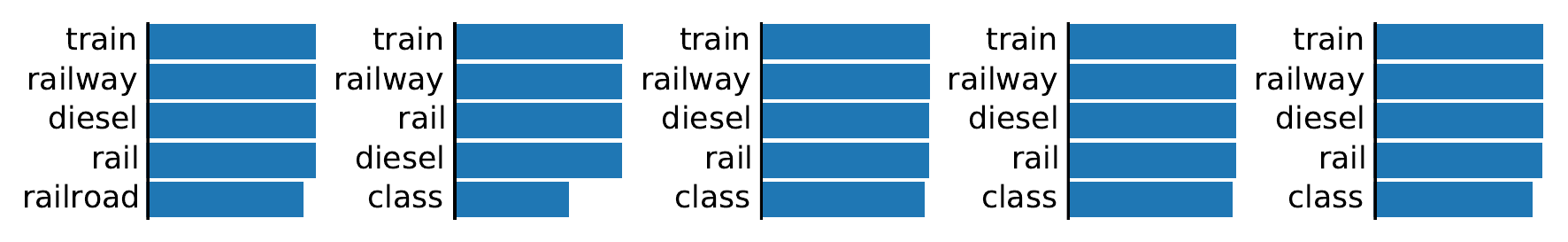}
        \end{subfigure} \\
	\vspace{-.01in}
        \begin{subfigure}[b]{\textwidth}
		\centering
                \includegraphics[width=1\textwidth]{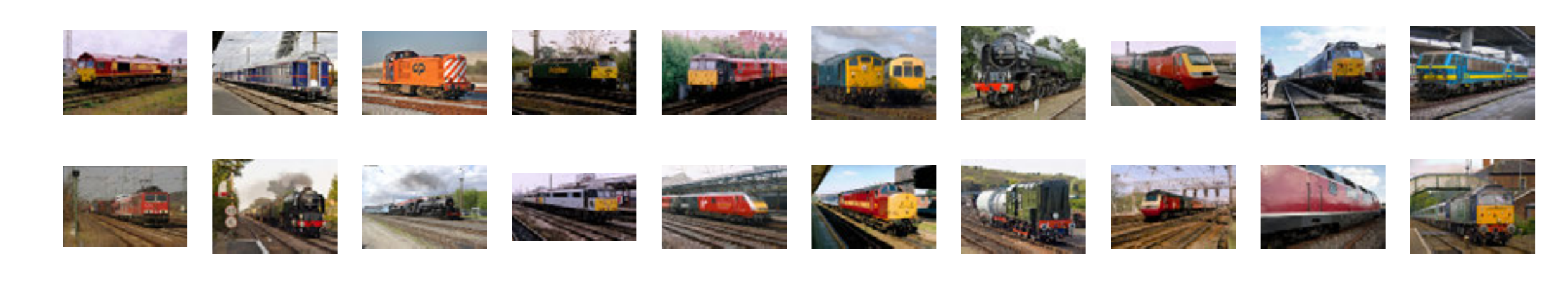}
		\vspace{-.4in}
                \caption{\mycaptionsize\tagname}
		\vspace{-.1in}
        \end{subfigure}
\end{figure*}

\begin{figure*}
        \centering
	\vspace{-.3in}
        \newcommand{\tagname}{kitchen}
        \begin{subfigure}[b]{\textwidth}
		\centering
                \includegraphics[width=.9\textwidth]{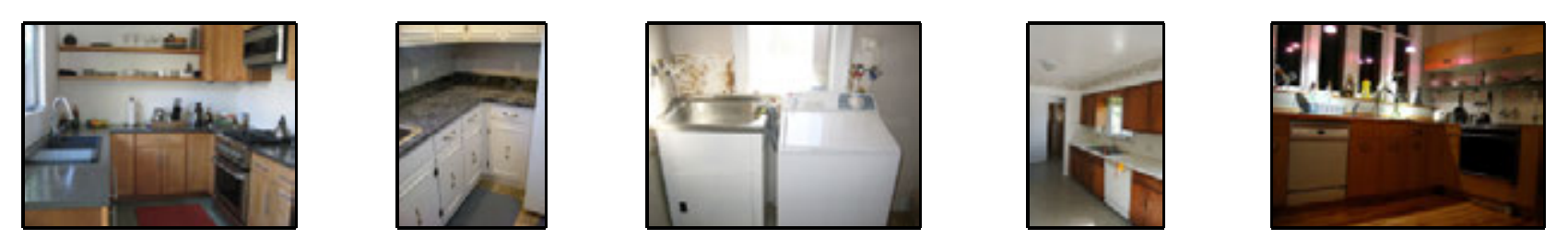}
        \end{subfigure} \\
	\vspace{-.1in}
        \begin{subfigure}[b]{\textwidth}
		\centering
                \includegraphics[width=.9\textwidth]{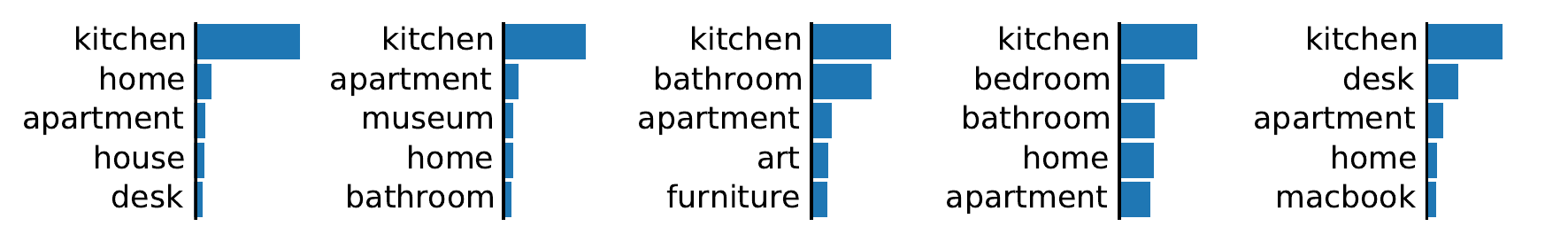}
        \end{subfigure} \\
	\vspace{-.1in}
        \begin{subfigure}[b]{\textwidth}
		\centering
                \includegraphics[width=1\textwidth]{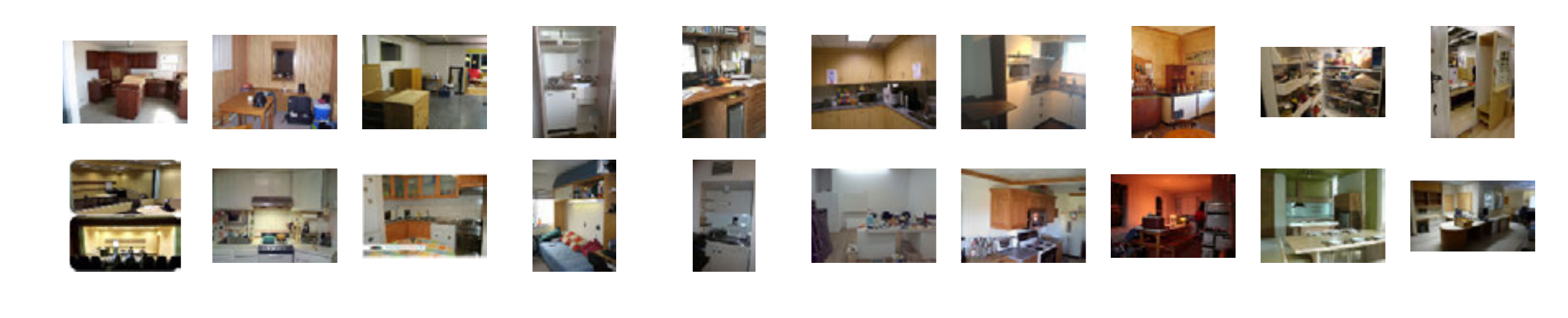}
		\vspace{-.4in}
                \caption{\mycaptionsize\tagname}
		\vspace{-.1in}
        \end{subfigure}
        \rule{500px}{1px}
	\vspace{-.1in}

        \renewcommand{\tagname}{bedroom}
        \begin{subfigure}[b]{\textwidth}
		\centering
                \includegraphics[width=.9\textwidth]{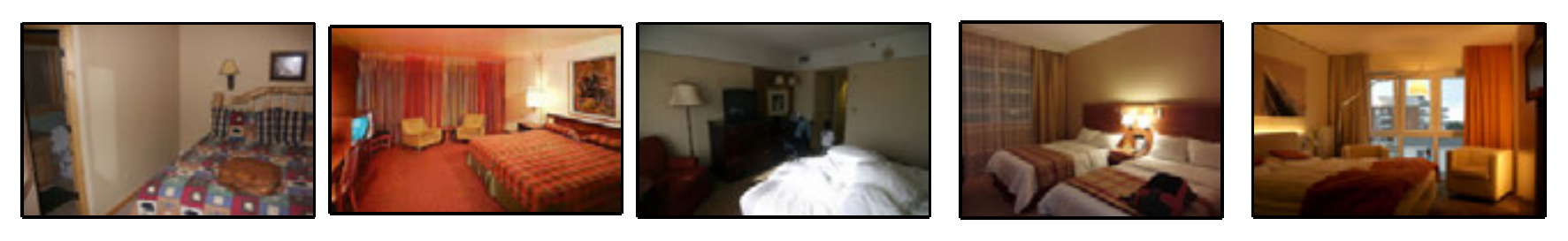}
        \end{subfigure} \\
	\vspace{-.1in}
        \begin{subfigure}[b]{\textwidth}
		\centering
                \includegraphics[width=.9\textwidth]{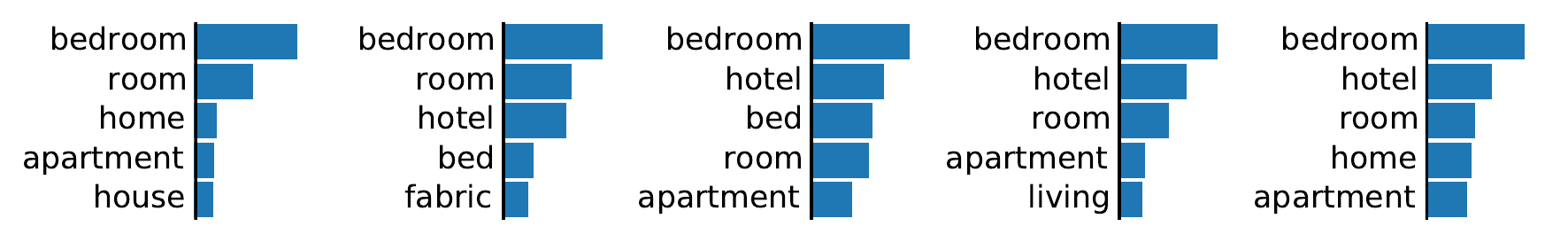}
        \end{subfigure} \\
	\vspace{-.1in}
        \begin{subfigure}[b]{\textwidth}
		\centering
                \includegraphics[width=1\textwidth]{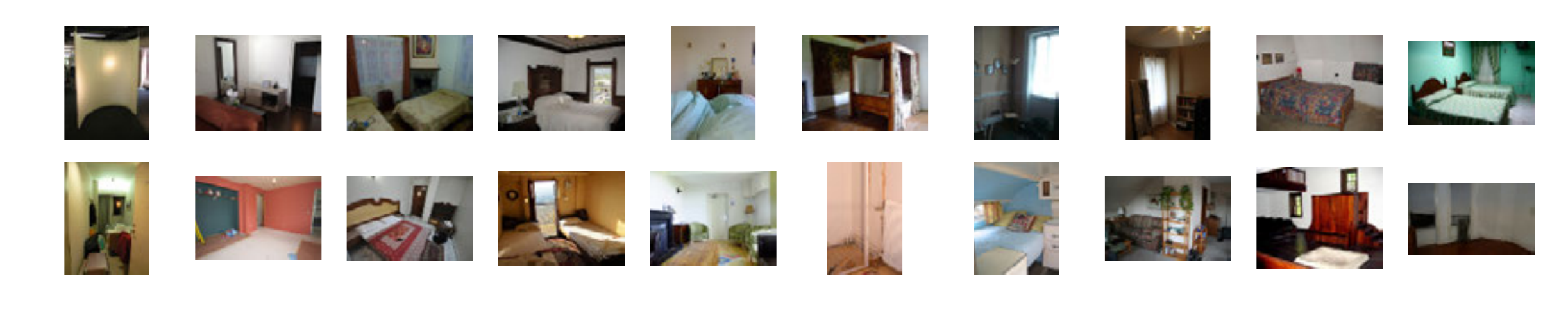}
		\vspace{-.4in}
                \caption{\mycaptionsize\tagname}
		\vspace{-.1in}
        \end{subfigure}
        \rule{500px}{1px}
	\vspace{-.1in}

        \renewcommand{\tagname}{bathroom}
        \begin{subfigure}[b]{\textwidth}
		\centering
                \includegraphics[width=.9\textwidth]{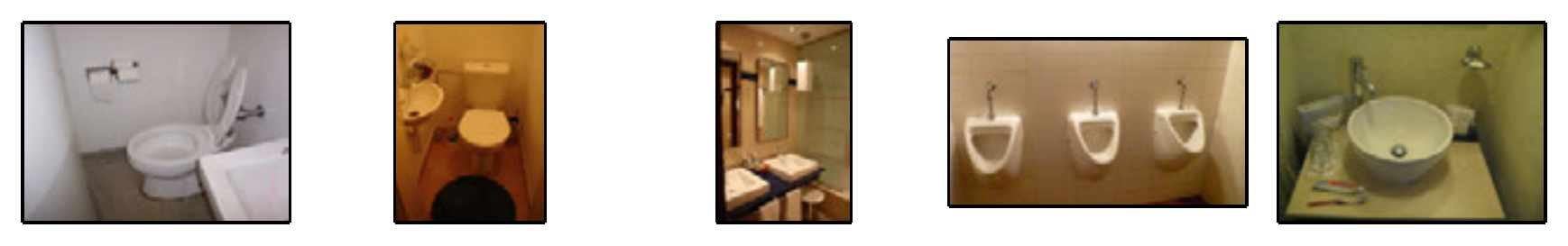}
        \end{subfigure} \\
	\vspace{-.1in}
        \begin{subfigure}[b]{\textwidth}
		\centering
                \includegraphics[width=.9\textwidth]{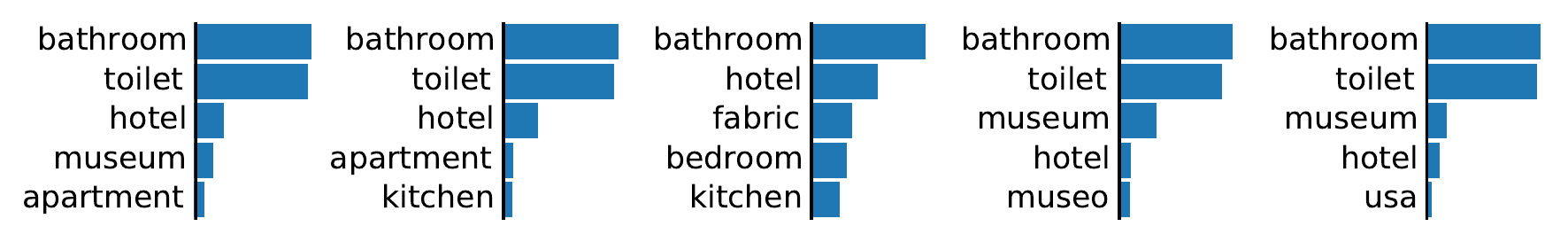}
        \end{subfigure} \\
	\vspace{-.01in}
        \begin{subfigure}[b]{\textwidth}
		\centering
                \includegraphics[width=1\textwidth]{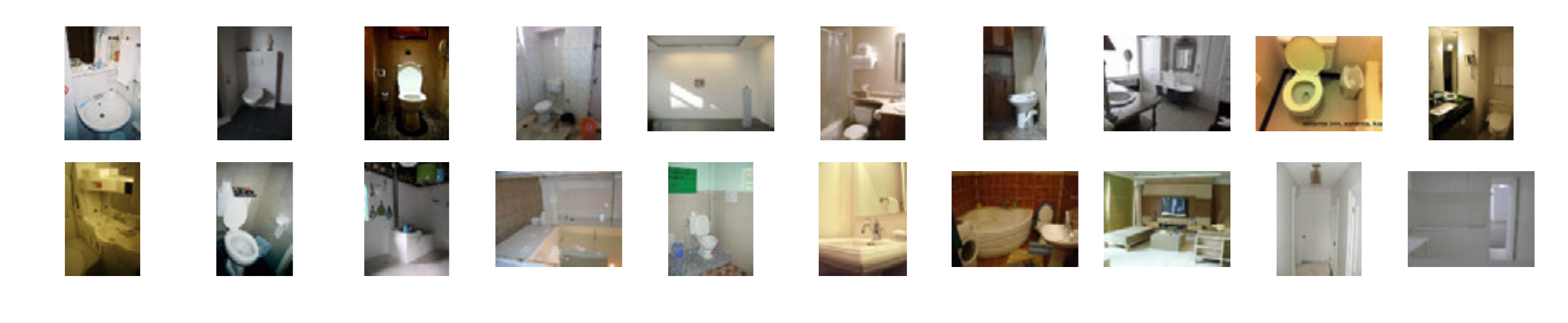}
		\vspace{-.4in}
                \caption{\mycaptionsize\tagname}
		\vspace{-.1in}
        \end{subfigure}
\end{figure*}

\begin{figure*}
        \centering
	\vspace{-.3in}
        \newcommand{\tagname}{baseball}
        \begin{subfigure}[b]{\textwidth}
		\centering
                \includegraphics[width=.9\textwidth]{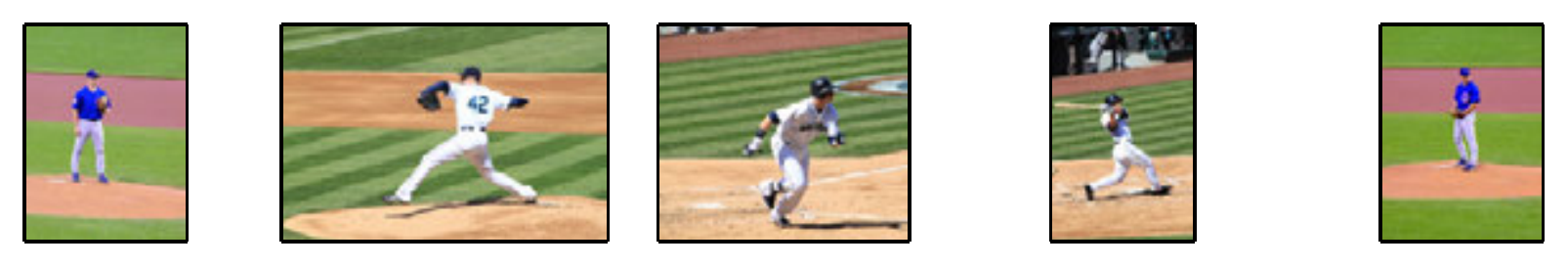}
        \end{subfigure} \\
	\vspace{-.1in}
        \begin{subfigure}[b]{\textwidth}
		\centering
                \includegraphics[width=.9\textwidth]{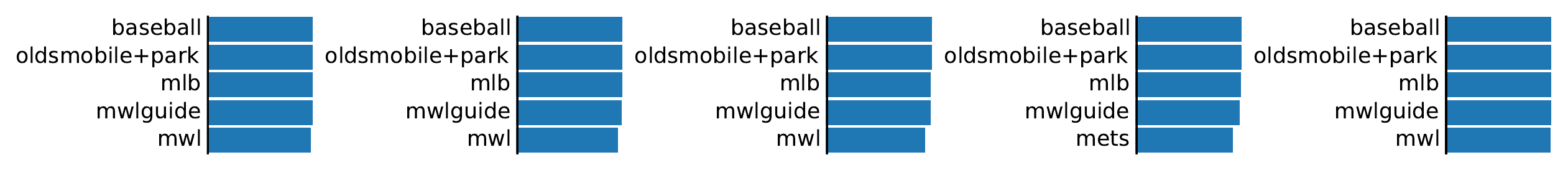}
        \end{subfigure} \\
	\vspace{-.1in}
        \begin{subfigure}[b]{\textwidth}
		\centering
                \includegraphics[width=1\textwidth]{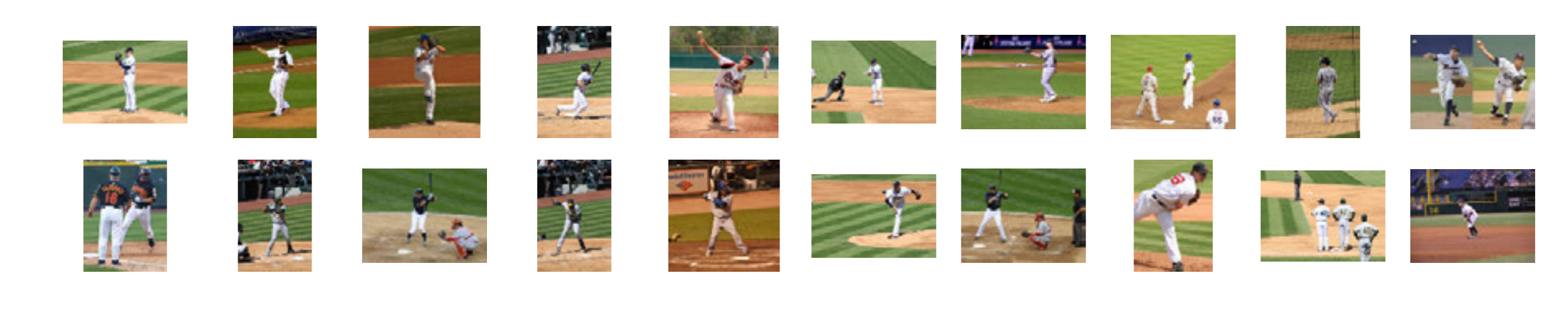}
		\vspace{-.4in}
                \caption{\mycaptionsize\tagname}
		\vspace{-.1in}
        \end{subfigure}
        \rule{500px}{1px}
	\vspace{-.1in}

        \renewcommand{\tagname}{swimming}
        \begin{subfigure}[b]{\textwidth}
		\centering
                \includegraphics[width=.9\textwidth]{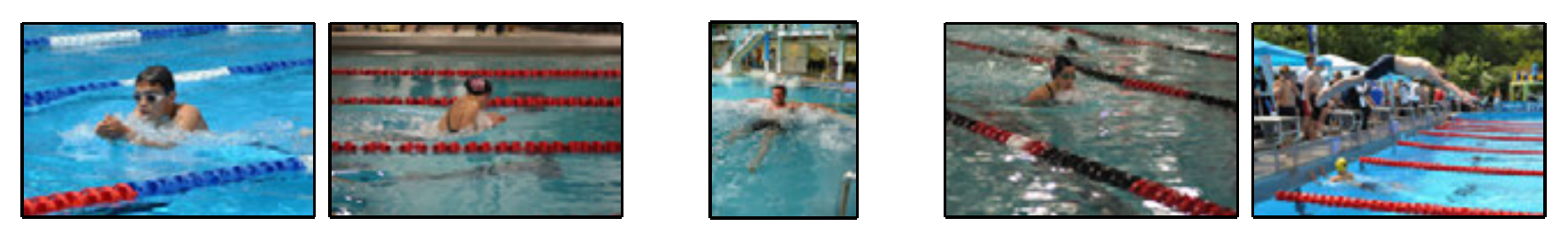}
        \end{subfigure} \\
	\vspace{-.1in}
        \begin{subfigure}[b]{\textwidth}
		\centering
                \includegraphics[width=.9\textwidth]{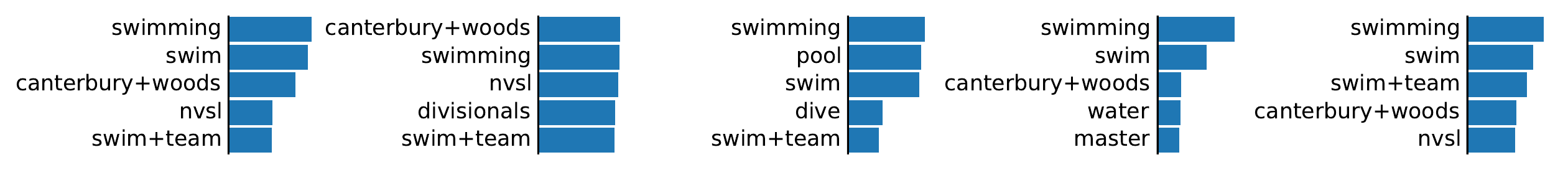}
        \end{subfigure} \\
	\vspace{-.1in}
        \begin{subfigure}[b]{\textwidth}
		\centering
                \includegraphics[width=1\textwidth]{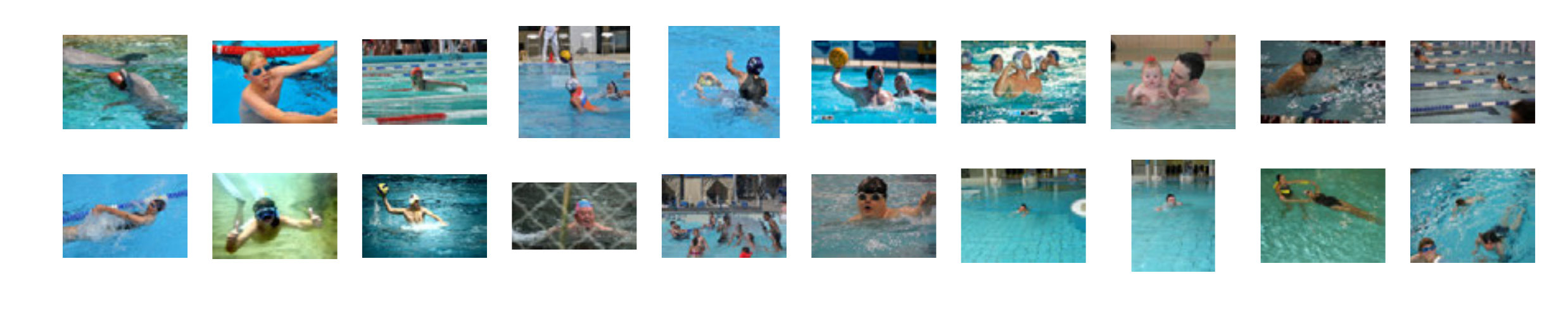}
		\vspace{-.4in}
                \caption{\mycaptionsize\tagname}
		\vspace{-.1in}
        \end{subfigure}
        \rule{500px}{1px}
	\vspace{-.1in}

        \renewcommand{\tagname}{basketball}
        \begin{subfigure}[b]{\textwidth}
		\centering
                \includegraphics[width=.9\textwidth]{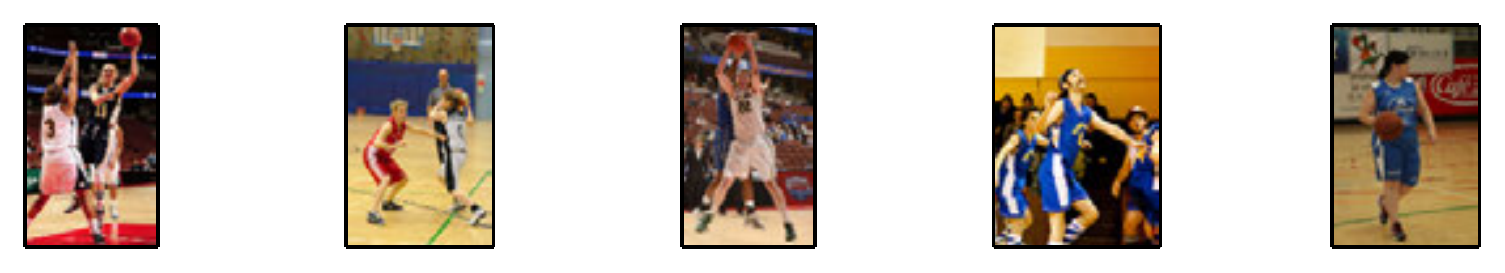}
        \end{subfigure} \\
	\vspace{-.1in}
        \begin{subfigure}[b]{\textwidth}
		\centering
                \includegraphics[width=.9\textwidth]{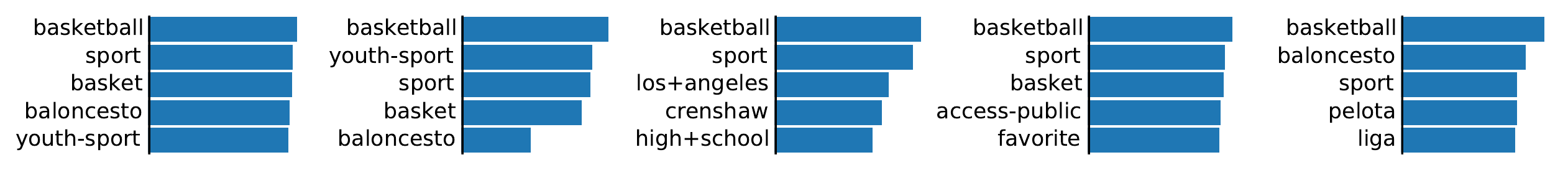}
        \end{subfigure} \\
	\vspace{-.01in}
        \begin{subfigure}[b]{\textwidth}
		\centering
                \includegraphics[width=1\textwidth]{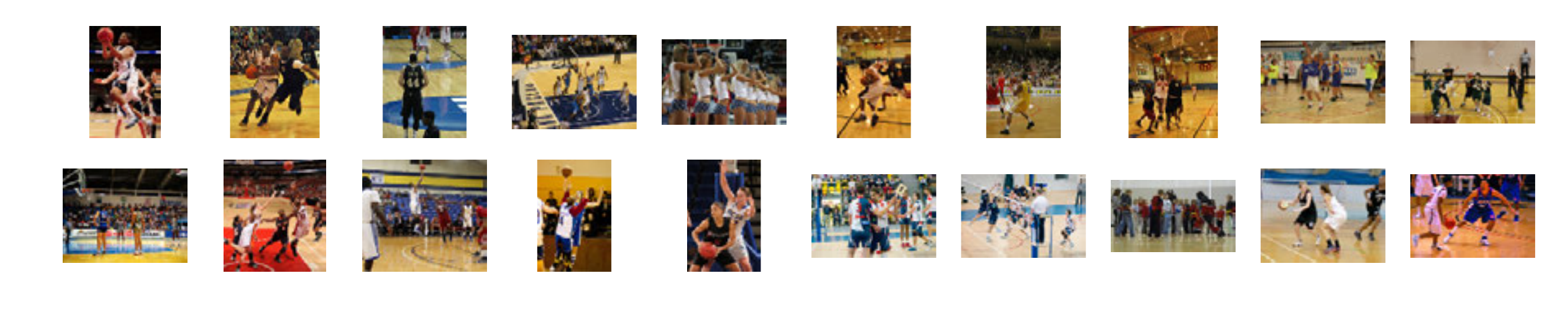}
		\vspace{-.4in}
                \caption{\mycaptionsize\tagname}
		\vspace{-.1in}
        \end{subfigure}
\end{figure*}

\begin{figure*}
        \centering
	\vspace{-.3in}
        \newcommand{\tagname}{brown}
        \begin{subfigure}[b]{\textwidth}
		\centering
                \includegraphics[width=.9\textwidth]{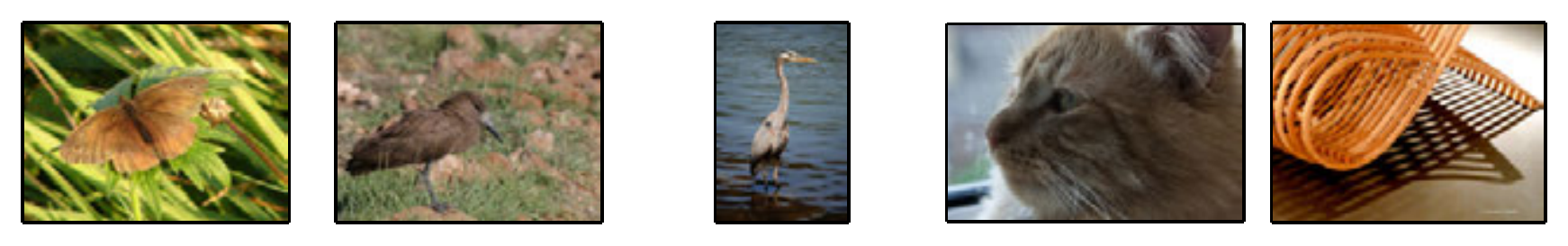}
        \end{subfigure} \\
	\vspace{-.1in}
        \begin{subfigure}[b]{\textwidth}
		\centering
                \includegraphics[width=.9\textwidth]{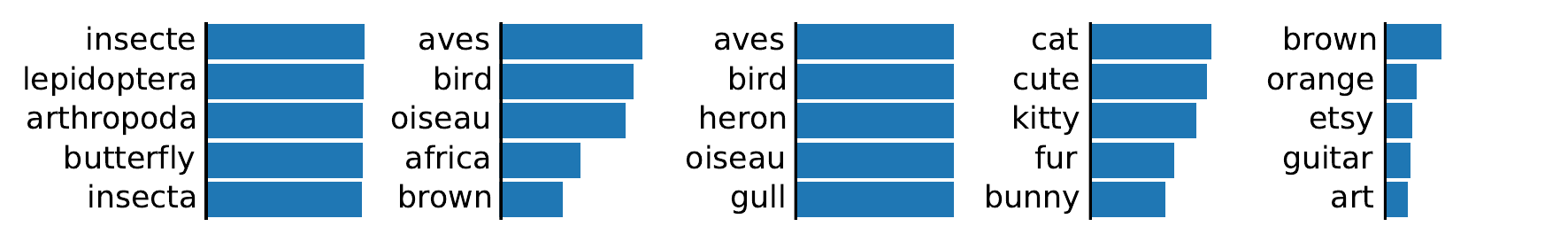}
        \end{subfigure} \\
	\vspace{-.1in}
        \begin{subfigure}[b]{\textwidth}
		\centering
                \includegraphics[width=1\textwidth]{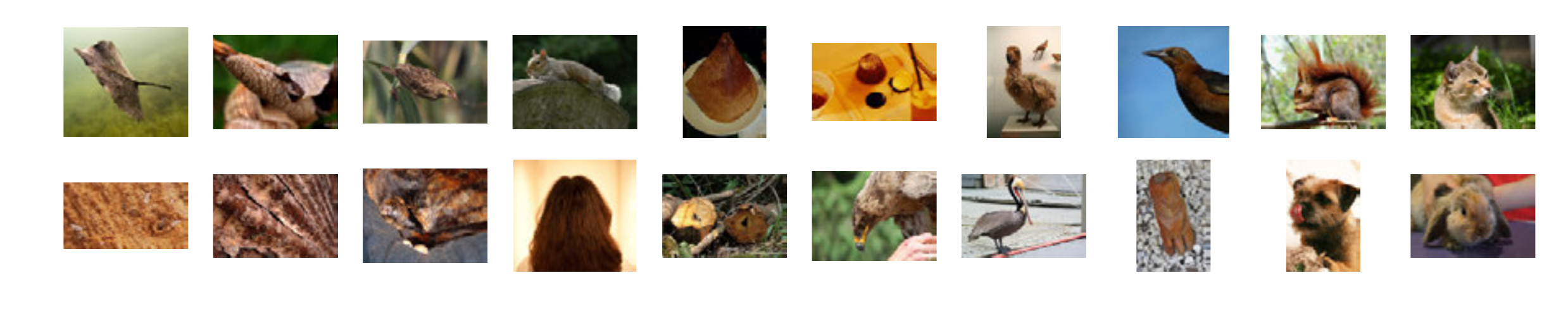}
		\vspace{-.4in}
                \caption{\mycaptionsize\tagname}
		\vspace{-.1in}
        \end{subfigure}
        \rule{500px}{1px}
	\vspace{-.1in}

        \renewcommand{\tagname}{blue}
        \begin{subfigure}[b]{\textwidth}
		\centering
                \includegraphics[width=.9\textwidth]{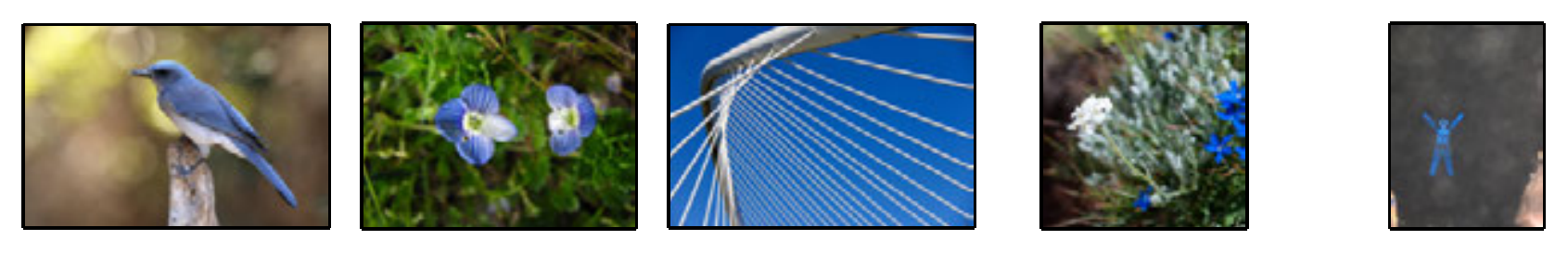}
        \end{subfigure} \\
	\vspace{-.1in}
        \begin{subfigure}[b]{\textwidth}
		\centering
                \includegraphics[width=.9\textwidth]{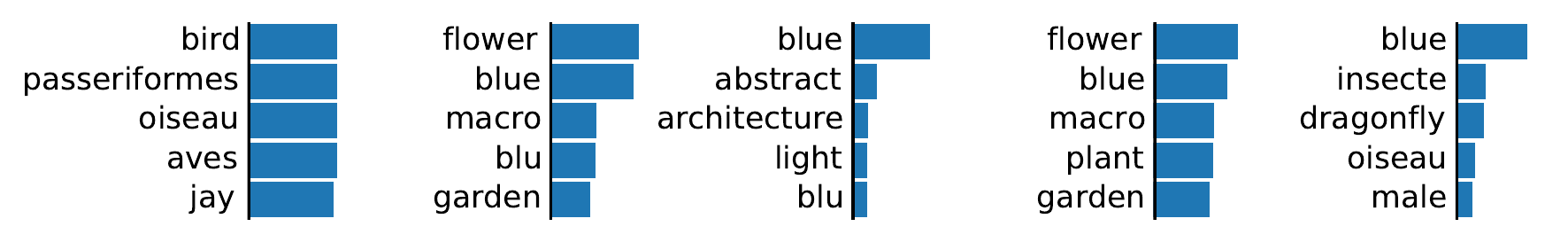}
        \end{subfigure} \\
	\vspace{-.1in}
        \begin{subfigure}[b]{\textwidth}
		\centering
                \includegraphics[width=1\textwidth]{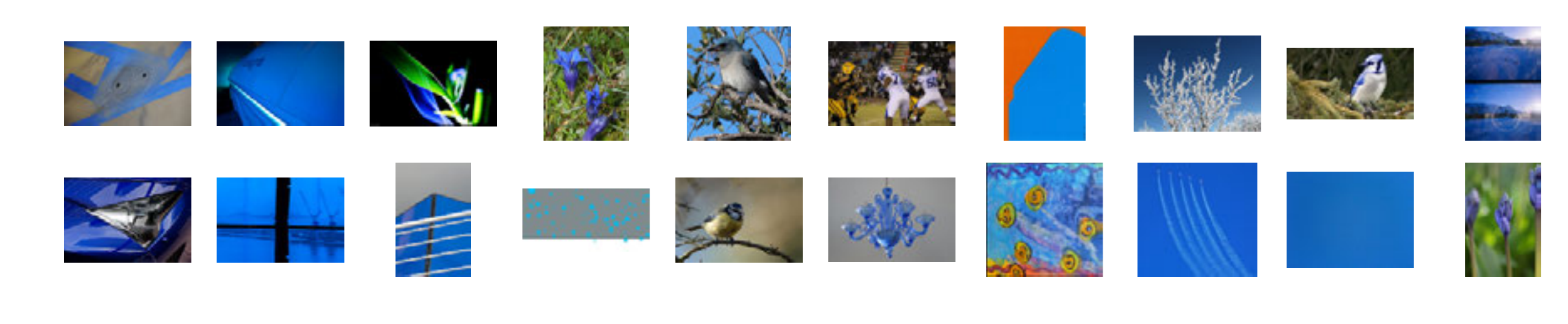}
		\vspace{-.4in}
                \caption{\mycaptionsize\tagname}
		\vspace{-.1in}
        \end{subfigure}
        \rule{500px}{1px}
	\vspace{-.1in}

        \renewcommand{\tagname}{green}
        \begin{subfigure}[b]{\textwidth}
		\centering
                \includegraphics[width=.9\textwidth]{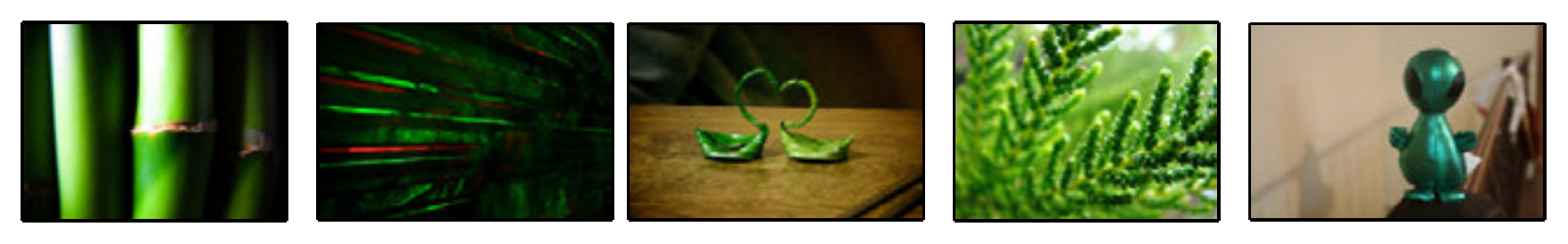}
        \end{subfigure} \\
	\vspace{-.1in}
        \begin{subfigure}[b]{\textwidth}
		\centering
                \includegraphics[width=.9\textwidth]{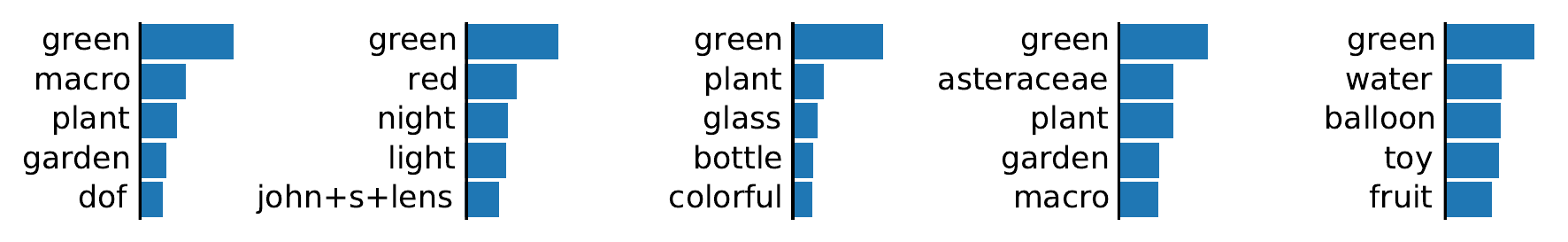}
        \end{subfigure} \\
	\vspace{-.01in}
        \begin{subfigure}[b]{\textwidth}
		\centering
                \includegraphics[width=1\textwidth]{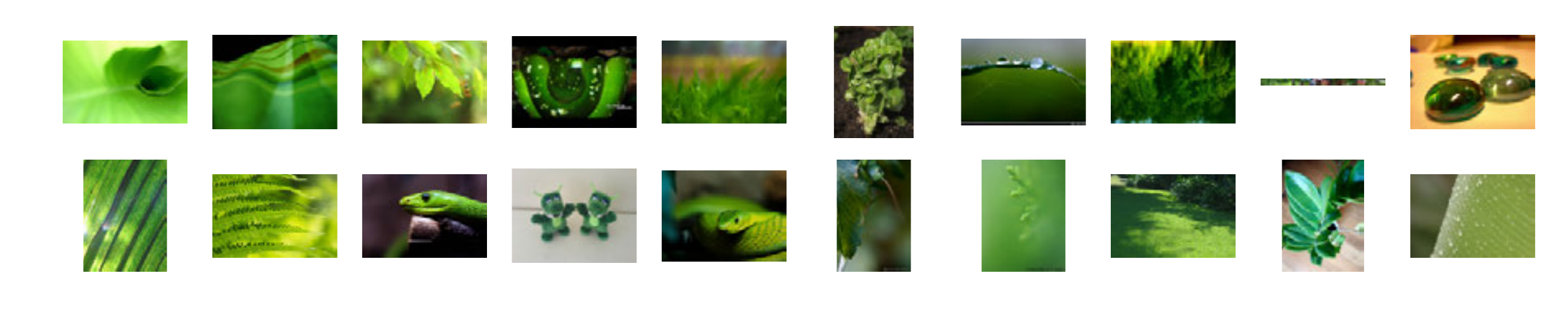}
		\vspace{-.4in}
                \caption{\mycaptionsize\tagname}
		\vspace{-.1in}
        \end{subfigure}
\end{figure*}

\addtocounter{subfigure}{-24}

\clearpage

\begin{figure*}
        \centering
	\vspace{-.3in}
        \newcommand{\tagname}{guitar}
        \begin{subfigure}[b]{\textwidth}
		\centering
                \includegraphics[width=.9\textwidth]{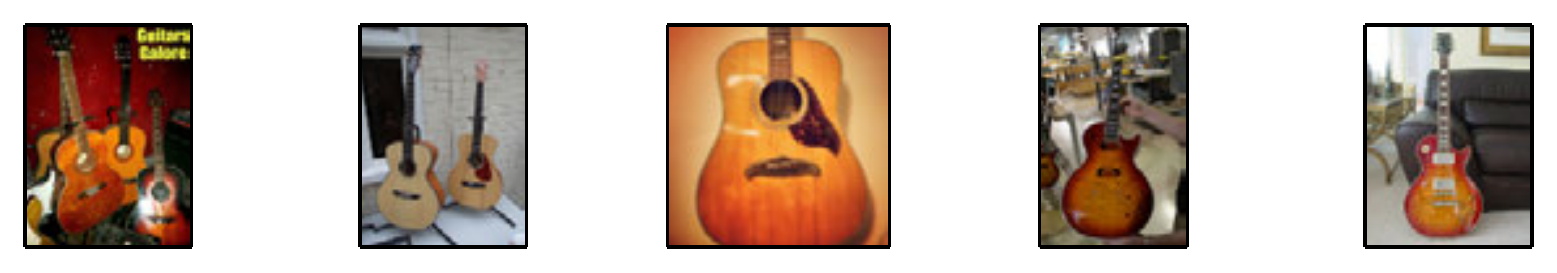}
        \end{subfigure} \\
	\vspace{-.1in}
        \begin{subfigure}[b]{\textwidth}
		\centering
                \includegraphics[width=.9\textwidth]{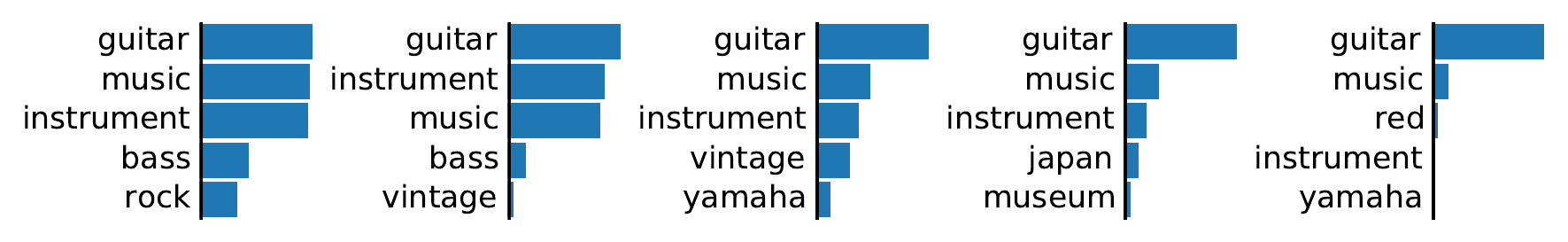}
        \end{subfigure} \\
	\vspace{-.1in}
        \begin{subfigure}[b]{\textwidth}
		\centering
                \includegraphics[width=1\textwidth]{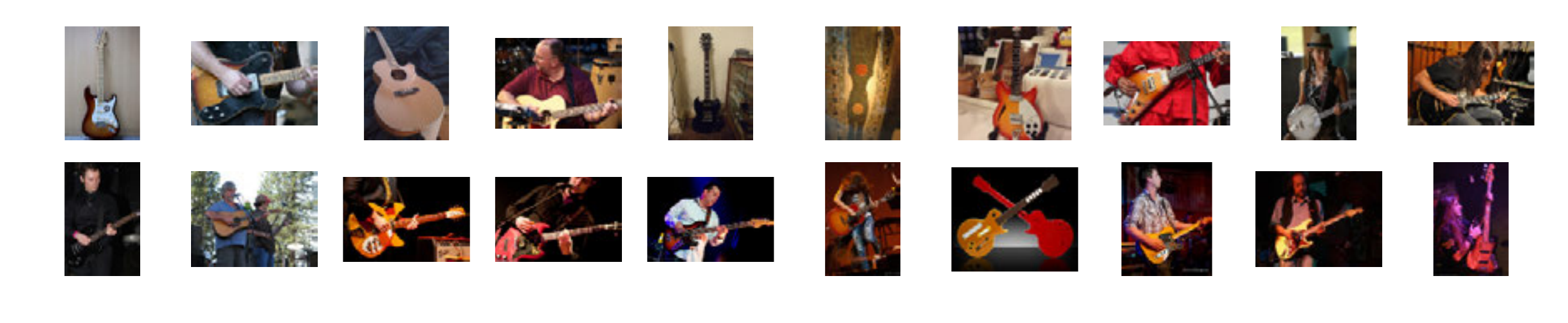}
		\vspace{-.4in}
                \caption{\mycaptionsize\tagname}
		\vspace{-.1in}
        \end{subfigure}
        \rule{500px}{1px}
	\vspace{-.1in}

        \renewcommand{\tagname}{drum}
        \begin{subfigure}[b]{\textwidth}
		\centering
                \includegraphics[width=.9\textwidth]{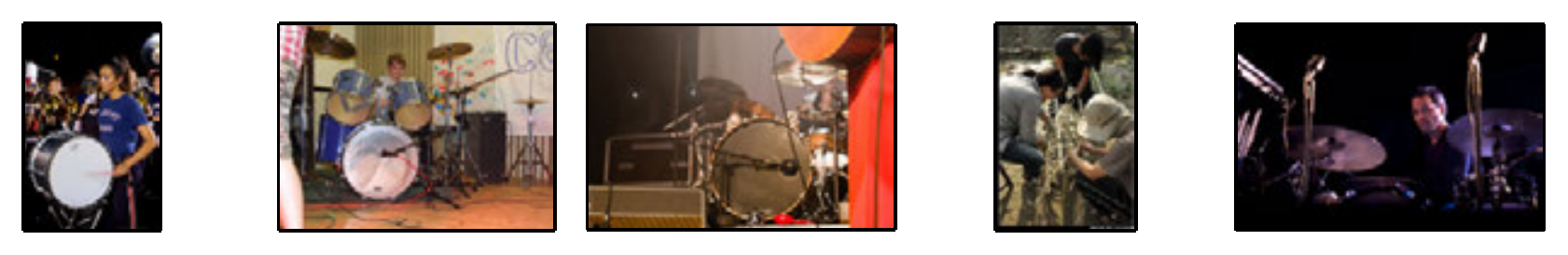}
        \end{subfigure} \\
	\vspace{-.1in}
        \begin{subfigure}[b]{\textwidth}
		\centering
                \includegraphics[width=.9\textwidth]{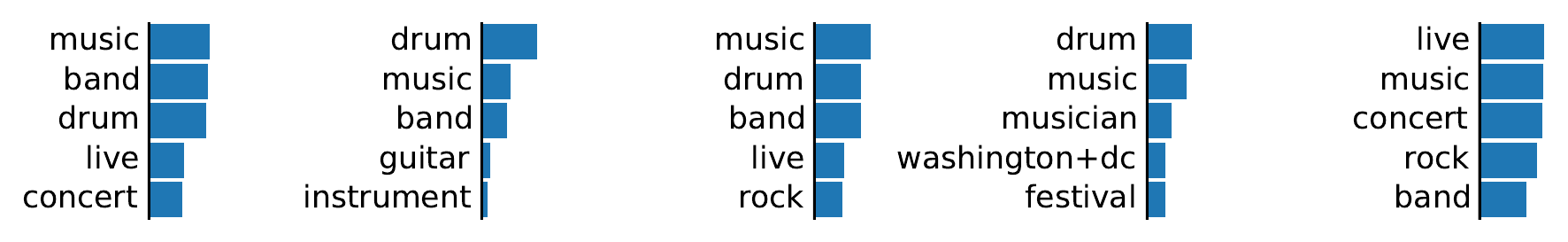}
        \end{subfigure} \\
	\vspace{-.1in}
        \begin{subfigure}[b]{\textwidth}
		\centering
                \includegraphics[width=1\textwidth]{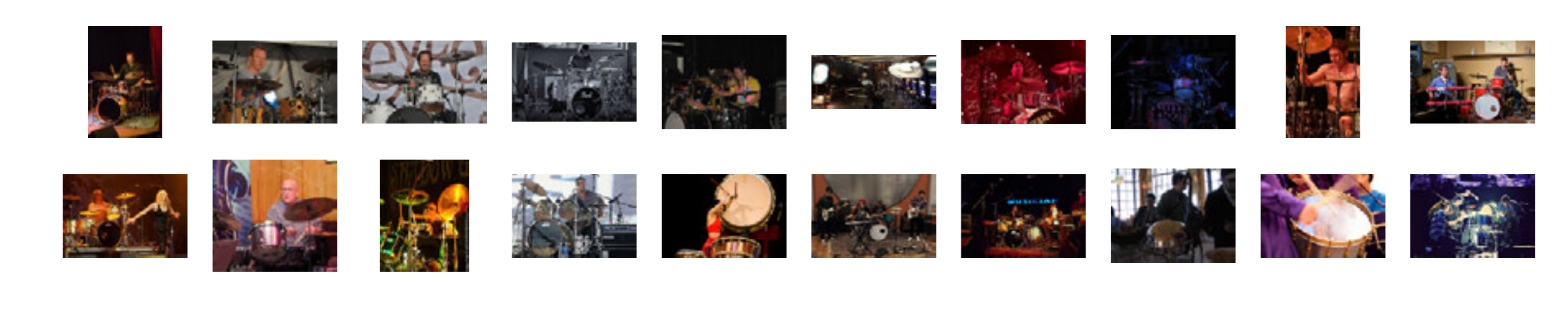}
		\vspace{-.4in}
                \caption{\mycaptionsize\tagname}
		\vspace{-.1in}
        \end{subfigure}
        \rule{500px}{1px}
	\vspace{-.1in}

        \renewcommand{\tagname}{bass}
        \begin{subfigure}[b]{\textwidth}
		\centering
                \includegraphics[width=.9\textwidth]{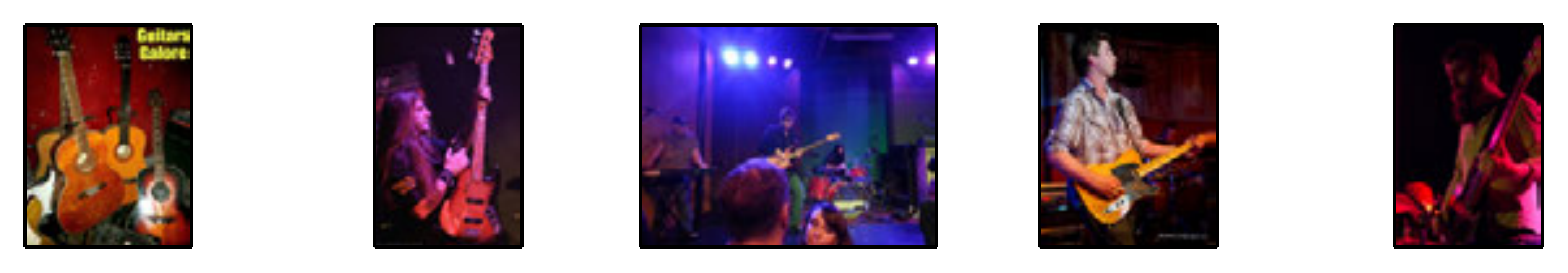}
        \end{subfigure} \\
	\vspace{-.1in}
        \begin{subfigure}[b]{\textwidth}
		\centering
                \includegraphics[width=.9\textwidth]{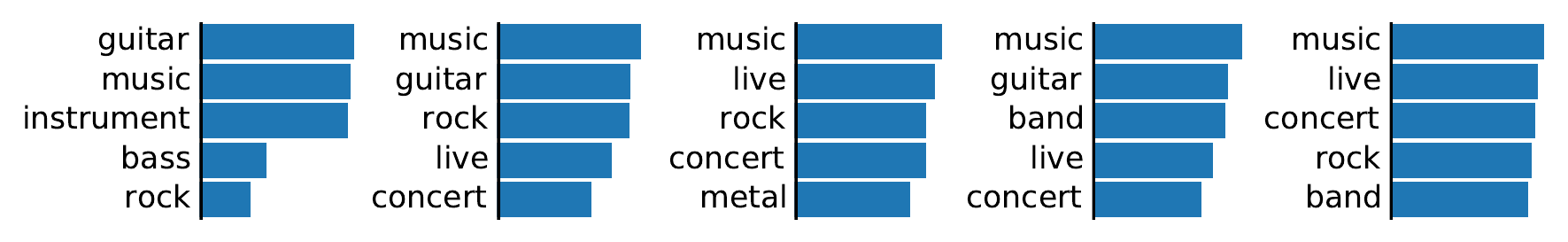}
        \end{subfigure} \\
	\vspace{-.01in}
        \begin{subfigure}[b]{\textwidth}
		\centering
                \includegraphics[width=1\textwidth]{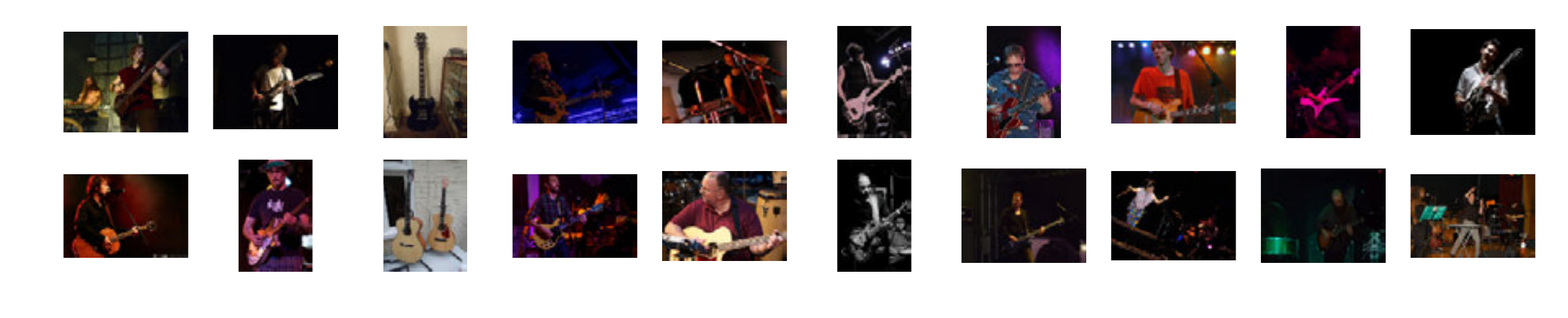}
		\vspace{-.4in}
                \caption{\mycaptionsize\tagname}
		\vspace{-.1in}
        \end{subfigure}
\end{figure*}
}

%\input{robust-logistic-notes}

%\end{document}

% 
% \documentclass[10pt,times]{article}
% 
% \textheight 8.5truein
% \textwidth  6.25truein
% \headheight 0.25truein
% \headsep 0.5truein
% \topmargin -0.25truein
% \evensidemargin -.25 truein
% \oddsidemargin  0 truein
% \parindent 0.25truein
% \baselineskip 16pt
% 
% \usepackage{graphicx}
% 
% \input{macros}
% 
% \title{Supplemental Material: Title TBD}
% \author{CVPR 2015 submission 337}
% 
% \begin{document}
% 
% \maketitle
% 
% 
% 
% \section{Details of correspondence calculation}
% \label{sec:appCalc}
% 
% We now explain how we estimated the percentage of Flickr tags absent from ImageNet concepts.
% We collected the top 1000 Flickr tags, and manually filtered out non-image and location tags, with 612 tags remaining. 
% We determined an automatic mapping from Flickr tags to WordNet synsets, by mapping each tag to its top WordNet noun synset, and manually corrected mismatches in the top 100 Flickr tags. We call an ImageNet synset \textit{large} if it has 1000 of more node in the subtree.
% Of the top 100 Flickr tags, we found that 54 of them had large ImageNet subtree before correcting mismatches, and 62 had large subtrees after manual corrections.
% Of the remaining 512 uncorrected tags, 189 (37\%) have large subtrees. Linear extrapolation suggests that 
% $1- ((189.0 * (62.0/54.0)) + 62)/612=54\%$ of tags are missing ImageNet subtrees. Of course, there are a number of questionable assumptions in this model, e.g., 1000 images may not be enough images for many classes, such as \tn{art}.
% 
% \end{document}

\maketitle

\end{comment}

\newcommand{\expect}[1]{<\!{#1}\!>}

\section{Basic Logistic Regression}

\begin{eqnarray}
\bw & \mbox{logistic weights} \\
\bx & \mbox{image features} \\
s= \bw^T\bx & \mbox{score given image data} \\
y \in \{0,1\} & \mbox{observed label for each image }
\end{eqnarray}

The model of label probabilities given image data is:
\begin{eqnarray}
s &=& \bw^T\bx \\
\sigmoid(s) &=& \frac{1}{1+e^{-s}} \\
P(y=1 | s) &=& \sigmoid(s)
\end{eqnarray}
The loss function for a dataset $\{(\bx_i, y_i)\}$ is
\begin{eqnarray}
L &=& -\ln P(y_{1:N} | \bx_{1:N}) \\
&=& -\ln  \left(\prod_{i : y_i = 1} P(y_i = 1 | \bx_i ) \right)
 \left(\prod_{i : y_i = 0} P(y_i = 0 | \bx_i ) \right) \\
&=&\sum_i \left( - y_i \ln P(y_i=1 | \bx_i) - (1-y_i) \ln P(y_i=0 | \bx_i\right) \\
&=&\sum_i \left( - y_i \ln \sigmoid(s_i) - (1-y_i) \ln (1-\sigmoid(s_i)) \right)
\end{eqnarray}
We can also rearrange terms:
\begin{eqnarray}
1-\sigmoid(s) &=& \frac{1+e^{-s}}{1+e^{-s}} - \frac{1}{1+e^{-s}}  = e^{-s} \sigmoid(s) \\
L &=& \sum_i \left( - y_i \ln \sigmoid(s_i) - (1-y_i) \ln e^{-s} \sigmoid(s_i) \right) \\
&=& \sum_i \left(-\ln \sigmoid(s_i) + (1-y_i) s_i \right) \\
&=& \sum_i \left(\ln (1+e^{-s}) + (1-y_i) s_i \right) 
%&=& \sum_i \left(\softmax(0,-s) + (1-y_i) s_i \right) 
\end{eqnarray}

\paragraph{Gradients.}
During optimization, we use the gradients with respect to $\bw$:
\begin{eqnarray}
\frac{d}{d\bw} \sigmoid(s) %&=& \sigmoid(s) \sigmoid(s) e^{-s} \frac{ds}{d\bw} \\
&=& \sigmoid(s) \sigmoid(s) e^{-s} \bx \\
&=& \sigmoid(s) (1-\sigmoid(s)) \bx \\
\frac{dL}{d\bw} 
%&=& \sum_i \left(- \sigmoid(s_i) e^{-s_i} \bx_i + y_i \bx_i \right) \\
%&=& \sum_i (-1+\sigmoid(s_i) + y_i )\bx_i 
&=& \sum_i \left( - \frac{y_i}{\sigmoid(s_i)} \frac{d}{d\bw} \sigmoid(s_i) 
-  \frac{1-y_i}{1-\sigmoid(s_i)} \frac{d}{d\bw} (1-\sigmoid(s_i)) \right)\\
&=& \sum_i \left( - y_i (1-\sigmoid(s)) \bx_i 
+  (1-y_i) \sigmoid(s_i) \bx_i\right) \\
&=& \sum_i \left(  \sigmoid(s_i) -y_i   \right) \bx_i
\end{eqnarray}
Note that this is zero when $y_i=\sigmoid(s_i)$, which indicates a perfect data fit.

Derivation using alternate form:
\begin{eqnarray}
\frac{dL}{d\bw} 
&=& \sum_i \left( (\sigmoid(s_i)-1)\bx_i + (1-y_i) \bx_i \right) \\
&=& \sum_i (\sigmoid(s_i)  - y_i )\bx_i 
\end{eqnarray}

\section{Robust Logistic Regression}

\begin{eqnarray}
\bw & \mbox{logistic weights} \\
\bx & \mbox{image features} \\
s= \bw^T\bx & \mbox{score given image data} \\
y \in \{0,1\} & \mbox{observed label for each image } \\
z \in \{0,1\} & \mbox{hidden true label for each image} 
\end{eqnarray}

The model of observations given scores is
\begin{eqnarray}
P(z=1 | s) &=& \sigmoid(s) \\
P(y=1 | z=1) &=& \pi \\
P(y=0 | z=1) &=& 1-\pi ~~\mbox{false negative probability}\\
P(y=0 | z=0) &=& \gamma \\
P(y=1 | z=0) &=& 1-\gamma ~~\mbox{false positive probability}
\end{eqnarray}
In the paper, we fix $\gamma=1$.

The marginal probability of a given observation is:
\begin{eqnarray}
P(y|s) &=& \sum_{z\in\{0,1\}} P(y,z|s) = P(y|z=1)P(z=1|s) + P(y|z=0)P(z=0|s)\\
P(y=1|s)&=& \pi \sigmoid(s) + (1-\gamma)(1-\sigmoid(s)) \\
&=& \pi \sigmoid(s) + (1-\gamma)e^{-s}\sigmoid(s)  \\
&=& \sigmoid(s)((1-\gamma)e^{-s} + \pi)  \\
 P(y=0|s)&=& (1-\pi) \sigmoid(s) + \gamma(1-\sigmoid(s)) \\
 &=& (1-\pi) \sigmoid(s) + \gamma e^{-s}\sigmoid(s) \\
 &=& \sigmoid(s) ( 1-\pi + \gamma e^{-s})
\end{eqnarray}

The Maximum Likelihood loss function can be written:
\begin{eqnarray}
L &=& \sum_i \left(-y_i \ln (\pi \sigmoid(s_i) + (1-\gamma)(1-\sigmoid(s_i)) )\right. \\
&& \quad \left.
- (1-y_i) \ln ((1-\pi) \sigmoid(s_i) + \gamma(1-\sigmoid(s_i))) \right) \\
&=& \sum_i \left( -y_i \ln \sigmoid(s)((1-\gamma)e^{-s_i} + \pi)
- (1-y_i) \ln \sigmoid(s_i) ( 1-\pi + \gamma e^{-s_i}) \right)\\
&=& \sum_i \left(-\ln \sigmoid(s_i) - y_i \ln ((1-\gamma)e^{-s_i} + \pi) - (1-y_i) \ln ( 1-\pi + \gamma e^{-s_i}) \right) 
%&=& \sum_i \softmax(0,-s_i) - y_i \softmax( \ln \pi, -s + \ln (1-\gamma))  \\
% \qquad -(1-y_i) \softmax(\ln (1-\pi), -s + \ln \gamma)
\end{eqnarray}
When $s_i>{}^{\sim}35$, and thus $P(z=1|s)\approx1$, the summand should be implemented as:
\begin{eqnarray}
-y_i \ln \pi - (1-y_i)\ln(1-\pi)
\end{eqnarray}

\paragraph{Gradients.}
During optimization, we could use the gradients with respect to $\bw$:
\begin{eqnarray}
\frac{dL}{d\bw} 
&=& \sum_i \left(\sigmoid(s_i)-1  - y_i \frac{ -(1-\gamma)e^{-s_i}} {(1-\gamma)e^{-s_i} + \pi} 
- (1-y_i) \frac{-\gamma e^{-s_i}}{1-\pi + \gamma e^{-s}}  
\right)  \bx_i  \\
&=& \sum_i \left(\sigmoid(s_i)-1  - y_i \frac{ -(1-\gamma)} {(1-\gamma) + \pi e^{s_i}} 
- (1-y_i) \frac{-\gamma }{ (1-\pi) e^{s_i}+ \gamma }  
\right)  \bx_i 
%&=& \sum_i \left( \sigmoid(s_i))-1
%- y_i (-\sigmoid(-s_i -\ln \pi +\ln(1-\gamma) ))
%- (1-y_i) (-\sigmoid(-s_i -\ln(1-\pi) + \ln\gamma )) \right) \bx_i \\
%&=& \sum_i \left( \sigmoid(s_i))-1
%+ y_i (1-\sigmoid(s_i +\ln \pi -\ln(1-\gamma) ))
%+ (1-y_i) (1-\sigmoid(s_i +\ln(1-\pi) - \ln\gamma )) \right) \bx_i
\end{eqnarray}
(Dividing by $e^s$ is done for stability. The case where $s$ is very large should also be handled by a separate condition.)

We also wish to optimize with respect to the parameters $\pi$ and $\gamma$:
\begin{eqnarray}
\frac{dL}{d\pi}&=& \sum_i \left ( -y_i \frac{-e^{-s_i}}{(1-\gamma)e^{-s_i} +\pi  }
-(1-y_i)\frac{e^{-s_i}}{1-\pi+\gamma e^{-s_i}} \right)\\
&=& \sum_i \left ( -y_i \frac{-1}{1-\gamma +\pi e^{s_i} }
-(1-y_i)\frac{1}{(1-\pi)e^{s_i}+\gamma } \right)\\
\frac{dL}{d\gamma}&=& \sum_i \left(
-y_i \frac{-e^{-s_i}}{(1-\gamma)e^{-s_i} + \pi}
-(1-y_i) \frac{e^{-s_i}}{1-\pi + \gamma e^{-s_i}} \right) \\
&=& \sum_i \left(
-y_i \frac{-1}{1-\gamma + \pi e^{s_i}}
-(1-y_i) \frac{1}{(1-\pi)e^{s_i} + \gamma } \right)
\end{eqnarray}

\subsection{Stochastic EM algorithm}

In the E-step, we compute the probabilities over the latent $z$'s given the data and the current model.
\begin{eqnarray}
\alpha_i \equiv P(z=1 | y_i, s_i) &= &\frac{P(y_i|z=1,s_i)P(z_i=1|s_i)}{P(y_i|s_i)}
\end{eqnarray}
which is computed with

%&=& \left\{ \begin{array}{ll}
%\frac{\pi \sigmoid(s_i)}{\pi \sigmoid(s_i) + (1-\gamma)(1-\sigmoid(s_i))} & y_i = 1\\
%\frac{(1-\pi) \sigmoid(s_i)}{(1-\pi) \sigmoid(s_i) + \gamma(1-\sigmoid(s_i))} & y_i =0 \end{array} \right.
\begin{eqnarray}
P(z=1 | y_i=1, s_i) &=& \frac{\pi \sigmoid(s_i)}{\pi \sigmoid(s_i) + (1-\gamma)(1-\sigmoid(s_i))} \\
P(z=1 | y_i=0, s_i) &=& \frac{(1-\pi) \sigmoid(s_i)}{(1-\pi) \sigmoid(s_i) + \gamma(1-\sigmoid(s_i))}
\end{eqnarray}

\paragraph{M-step derivation.}
In the M-step, we update the various model parameters. It can be derived by minimizing the negative expected complete log-likelihood:
\begin{eqnarray}
E &=& \expect{ -\sum_i \ln P(y_i, z_i | s_i)  }_{\alpha_i} \\
&=& \expect{ -\sum_i \ln P(y_i| z_i) P(z_i| s) }_{\alpha_i} \\
\expect{\ln P(y_i=1| z_i)} &=& \alpha_i \ln \pi + (1-\alpha_i) \ln(1-\gamma) \\
\expect{\ln P(y_i=0| z_i)} &=& \alpha_i \ln(1-\pi) + (1-\alpha_i) \ln\gamma \\
\expect{\ln P(z_i| s_i)} &=& \alpha_i\ln \sigmoid(s_i) + (1-\alpha_i) \ln(1-\sigmoid(s_i)) \\
\end{eqnarray}
The derivatives are then:
\begin{eqnarray}
\frac{dE}{d\pi} &=& 
-\sum_i \left(y_i \frac{\alpha_i}{\pi} +(1-y_i) \frac{-\alpha_i}{1-\pi}\right) \\
\frac{dE}{d\gamma}
&=& -\sum_i \left(-y_i \frac{1-\alpha_i}{1-\gamma}
+(1-y_i)\frac{1-\alpha_i}{\gamma}\right) \\
\frac{dE}{d\bw}
&=& 
-\sum_i \left(\alpha_i (1-\sigmoid(s_i)) - (1-\alpha_i) \sigmoid(s_i)\right) \bx_i \\
&=& \sum_i \left( \sigmoid(s_i) -\alpha_i  \right )\bx_i
\end{eqnarray}

Solving for $dE/d\pi=0$ and $dE/d\gamma=0$ gives:
\begin{eqnarray}
\pi &\leftarrow& \frac{\sum_i y_i \alpha_i}
{\sum_i  \alpha_i} \\
\gamma &\leftarrow&
\frac{\sum_i (1-y_i)(1-\alpha_i)}{\sum_i(1-\alpha_i)}
\end{eqnarray}
\paragraph{Stochastic EM algorithm.}
In the stochastic EM algorithm, we keep running tallies of 
\begin{eqnarray}
%S &=& \sum_i 1 \\
S_y &=& \sum_i y_i / N\\
S_{y\alpha} &=& \sum_i y_i \alpha_i /N\\
S_{\alpha} &=& \sum_i \alpha_i /N 
\end{eqnarray}
and then, in each M-step update, update the parameters as:
\begin{eqnarray}
\pi &\leftarrow& \frac{S_{y\alpha}}{S_{\alpha}} \\
\gamma &\leftarrow& \frac{1 -S_y-S_{\alpha} + S_{y\alpha}}{1-S_{\alpha}}
\end{eqnarray}

We can also use $dE/d\bw$ as a gradient estimate instead of $dL/d\bw$.

%\end{document}

\end{document}